\documentclass{article} 
\usepackage{iclr2027_conference,times}

\usepackage{amsmath,amsfonts,bm}

\def\eqref#1{equation~\ref{#1}}

\def\1{\bm{1}}

\DeclareMathAlphabet{\mathsfit}{\encodingdefault}{\sfdefault}{m}{sl}
\SetMathAlphabet{\mathsfit}{bold}{\encodingdefault}{\sfdefault}{bx}{n}

\usepackage[utf8]{inputenc} 
\usepackage[T1]{fontenc}    
\usepackage{booktabs}       
\usepackage{amsfonts}       
\usepackage{nicefrac}       
\usepackage{microtype}      
\usepackage{xcolor}         

\usepackage{amsmath, amssymb}
\usepackage{graphicx}
\usepackage{multirow}
\usepackage{mathtools}
\usepackage{amsthm}

\usepackage{lipsum,xspace,wrapfig,enumitem,titletoc}

\usepackage{url}
\usepackage[colorlinks=true,
            linkcolor=purple,
            citecolor=blue,
            urlcolor=purple]{hyperref}
            
\usepackage[capitalize,noabbrev]{cleveref}
\usepackage{pifont}

\usepackage{tikz}
\usetikzlibrary{shapes.geometric, arrows.meta, positioning, calc}

\usepackage{xcolor}

\newcommand{\cmark}{\checkmark}
\newcommand{\xmark}{\ding{55}}

\title{TopU-LBVS: A Realistic Multi Target Benchmark
for Ligand Based Virtual Screening}

\author{%
  Surbhi Kumar\thanks{Co-first authors.} \\
  UT Dallas, Math. Sci.\\
  Richardson, TX, USA\\
  \texttt{\small surbhi.kumar@utdallas.edu}
  \And
  Yuhe Zhou\footnotemark[1] \\
    National Inst. Bio. Sci.\\
  Beijing, China\\
  \texttt{\small zhouyuhe@nibs.ac.cn}
  \And
  Varun Shiralkar \\
  UT Dallas, Comp. Sci.\\
  Richardson, TX, USA\\
  \texttt{\small vjs230002@utdallas.edu}
  \AND
  Niu Huang\thanks{Co-senior authors.} \\
  National Institute of Biological Sciences\\
  Beijing, China\\
  \texttt{\small huangniu@nibs.ac.cn}
  \And
  Baris Coskunuzer\footnotemark[2] \\
  UT Dallas, Mathematical Sciences\\
  Richardson, TX, USA\\
  \texttt{\small coskunuz@utdallas.edu}
}

\iclrfinalcopy 
\begin{document}

\maketitle

\begin{abstract}
Ligand-based virtual screening (LBVS) is a practical first-pass tool in
early-stage drug discovery, but existing benchmarks can overestimate
performance through random negatives, easy decoys, limited target coverage,
and non-standardized evaluation protocols. We introduce
\textbf{TopU-LBVS}, a multi-target benchmark for LBVS under hard-negative
screening conditions. Starting from curated ChEMBL~35 bioactivity data,
TopU-LBVS covers 93 protein targets across 7 protein classes and constructs
target-specific screening libraries with property-matched, structurally
similar decoys at a fixed 1:40 active-to-decoy ratio. Libraries contain
roughly 400 to 10,000 compounds and are designed to reduce simple
physicochemical and nearest-neighbor fingerprint shortcuts.

TopU-LBVS provides three fixed protocols. \textbf{TopU-LBVS-full} evaluates
ChEMBL$^\ast \rightarrow$ TopU generalization across all 93 targets.
\textbf{TopU-LBVS-low} evaluates low-data TopU $\rightarrow$ TopU learning
within the hard-negative distribution. \textbf{TopU-LBVS-mini} provides a
compact seven-target protocol with a paired random-decoy control that changes
only the test decoys, enabling low-cost development and direct measurement of
the gap between random ChEMBL$^\ast$ and TopU decoys. Across ten reference
baselines spanning fingerprint methods, molecular GNNs, fingerprint hybrids,
and modern molecular models, performance under random-decoy evaluation
degrades sharply under hard-negative screening. We release data, fixed splits,
evaluation code, and baseline implementations for reproducible comparison of
future LBVS and molecular representation learning methods. Code and data are
available at~\footnote{Code: \url{https://github.com/topu-benchmark/topu-lbvs}\\
Data: \url{https://huggingface.co/datasets/topu-benchmark/topu-lbvs}}.

\end{abstract}

\vspace{-.15in}

\section{Introduction}
Ligand-based virtual screening (LBVS) ranks candidate compounds using
known active ligands and remains a practical first-pass tool in early-stage
drug discovery, especially when structural information is limited but
historical structure--activity relationship (SAR) data are available~\cite{Lagarde2015,walters2020new}.
Despite this importance, widely used LBVS benchmarks can substantially
simplify the screening problem~\cite{Sieg2019,Lagarde2015}. Many rely on
random negatives or easy decoys, have limited or uneven target coverage,
and lack fixed learning protocols for supervised model comparison~\cite{chen2019hidden,kramer2025need}.
Random or scaffold-agnostic splits can also place closely related analogues
across train and test sets, leading to overly optimistic estimates of
performance~\cite{Hu2020pretrain,yang2019analyzing}. Consequently, it remains unclear
which methods continue to work when candidate libraries contain
property-matched, structurally similar hard negatives rather than random
background compounds.

We introduce \textbf{TopU-LBVS}, a multi-target benchmark for ligand-based
virtual screening under hard-negative conditions. Starting from ChEMBL~35,
we curate a cleaned bioactivity resource, ChEMBL$^\ast$, covering
93 protein targets across 7 protein classes. For each target, we construct a
TopU screening library by pairing curated actives with hard decoys at a fixed
1:40 active-to-decoy ratio. TopU decoys are selected to match actives in
physicochemical properties while remaining structurally similar, reducing
simple property-based and trivial nearest-neighbour shortcuts. This design
targets a central weakness of standard LBVS evaluation: models that perform
well against randomly sampled negatives may fail on focused hard-negative
screening libraries~\cite{chen2019hidden,Sieg2019}.

TopU-LBVS is organized as a benchmark suite with three complementary
protocols. \textbf{TopU-LBVS-full} evaluates historical-to-hard-library
generalization: models are trained on target-specific ChEMBL$^\ast$ SAR data
and tested on the corresponding TopU library across all 93 targets.
\textbf{TopU-LBVS-low} evaluates low-data learning within the TopU
hard-negative distribution, again across all 93 targets, where training,
validation, and test compounds come from the same focused candidate library.
\textbf{TopU-LBVS-mini} provides a compact seven-target version, one target
per protein class, for rapid development, debugging, and low-cost comparison.
Its TopU arm follows the same curation, training, test-ratio, metric, and
evaluation-code design as the full protocol.

A key component of TopU-LBVS-mini is a paired random-decoy control for
hardness validation. For each mini target, we keep the same training data and
validation protocol, the same test actives, and the same 1:40 test ratio, but
replace TopU hard decoys with randomly sampled ChEMBL$^\ast$ decoys. The
paired comparison isolates the effect of test-decoy construction and directly
measures how much random-decoy evaluation can overestimate performance
relative to hard-negative screening.

We evaluate ten reference baselines spanning classical fingerprint methods,
molecular GNNs~\cite{Hu2020pretrain,xu2019powerful,velivckovic2018graph},
fingerprint-augmented hybrids, and modern molecular models~\cite{yang2019analyzing,ross2022large}.
All models are evaluated under fixed splits and identical target-specific
protocols. We use early enrichment as the primary metric: EF@1
TopU-LBVS-full and TopU-LBVS-mini, and EF@10\% for TopU-LBVS-low, where
EF@1\% and EF@5\% are unstable for the smallest low-data test libraries.
PR-AUC is used for validation and reported as a secondary metric, with
ROC-AUC, BEDROC, and additional enrichment metrics reported in the appendix.

\textbf{Contributions.}
We summarize the benchmark contributions as follows:
\begin{itemize}[left=0pt]
\item \textbf{A hard-negative LBVS benchmark.}
TopU-LBVS provides 93 target-specific tasks across 7 protein classes,
using property-matched, structurally similar decoys at a 1:40
active-to-decoy ratio.

\item \textbf{A curated ChEMBL$^\ast$ resource.}
We release a standardized ChEMBL~35 subset with assay filtering,
molecular standardization, activity binarization, artifact removal, and
duplicate resolution.

\item \textbf{Three fixed evaluation protocols.}
TopU-LBVS-full tests ChEMBL$^\ast \rightarrow$ TopU generalization;
TopU-LBVS-low tests low-data TopU $\rightarrow$ TopU learning; and
TopU-LBVS-mini provides a compact seven-target entry point for rapid
development.

\item \textbf{A paired random-decoy hardness control.}
The mini protocol includes matched random ChEMBL$^\ast$ control splits
that differ only in test-decoy construction, quantifying the gap between
random-decoy and TopU hard-negative evaluation.

\item \textbf{Reference baselines and open infrastructure.}
We provide ten baseline results and release the data, fixed splits,
implementations, and evaluation code for reproducible LBVS comparison.

\end{itemize}

\section{Background and Existing LBVS Benchmarks}
\label{sec:background}

\textbf{Desiderata for learning-based LBVS evaluation.} \quad
A useful learning-based LBVS benchmark should combine four properties:
\textit{challenging negatives}, so that activity cannot be inferred from
simple physicochemical or structural shortcuts; \textit{target diversity}, so
that results reflect performance across ligand chemistries and protein
families; \textit{screening-relevant imbalance}, rather than artificially
balanced splits; and a \textit{fixed reproducible learning protocol},
including released splits, seeds, and evaluation code for direct method
comparison.

\textbf{Classical decoy-based benchmarks.}\quad
DUD-E~\citep{mysinger2012directory} remains widely used for ligand-based and
structure-based virtual screening. Its decoys are matched to actives in coarse
physicochemical properties while avoiding close 2D structural similarity, but
they can often be separated by simple descriptors such as molecular weight or
logP~\citep{chen2019hidden}, and no canonical supervised-learning protocol or
fixed train/validation/test splits are provided.
DEKOIS~2.0~\citep{bauer2013dekois} improves aspects of decoy construction but
was likewise developed primarily for virtual-screening evaluation rather than
standardized supervised model comparison.

MUV~\citep{rohrer2009maximum} is an important precursor to hard-negative
evaluation. It uses experimentally screened PubChem compounds and
spatial-statistical optimization to reduce analogue bias and artificial
enrichment, producing deliberately challenging screening sets at a highly
imbalanced 1:500 active-to-inactive ratio. However, MUV contains only 17
targets and was not designed as a fixed supervised-learning benchmark with
historical-SAR training, low-data learning, and standardized
train/validation/test protocols. LIT-PCBA~\citep{tran2020lit} similarly
improves screening realism by using experimentally confirmed inactives from
PubChem dose-response assays, reducing false-negative concerns, but covers
only 15 targets and does not provide a canonical training-based learning
protocol.

\textbf{Learning-based evaluation beyond classical benchmarks.} \quad
Riniker and Landrum~\citep{riniker2013open} provide fixed compound lists and
evaluation scripts for fingerprint retrieval across 88 targets from MUV, DUD,
and ChEMBL, but the setup is similarity search rather than supervised
target-specific learning and uses random inactives. FS-Mol~\citep{stanley2021fsmol}
provides standardized few-shot ChEMBL-derived tasks and fixed evaluation
splits, but uses approximately balanced active-to-inactive ratios and random
inactives rather than explicitly constructed hard-negative screening
libraries. CACHE~\citep{ackloo2022cache} benchmarks prospective computational
hit-finding with experimental validation, but targets a structure-based
prospective setting rather than the retrospective ligand-based protocol
studied here. More broadly, ChEMBL-based activity-prediction
studies~\citep{yang2019analyzing,ross2022large} vary substantially in target
selection, activity thresholds, negative sampling, and splitting rules,
making direct cross-paper comparison difficult.

\begin{table}[t]
\centering
\caption{\footnotesize
\textbf{Comparison with representative LBVS benchmarks.}
\textit{Hard-negative design} denotes explicit construction intended to reduce
easy active--inactive separation or artificial enrichment.
\textit{Learning protocol} denotes a canonical training-based setup, and
\textit{fixed splits} denotes released benchmark-level train/validation/test
splits. TopU-LBVS additionally provides a 93-target low-data protocol, a
seven-target mini benchmark, and a paired random-decoy control.
}
\label{tab:benchmark_comparison}
\resizebox{\textwidth}{!}{%
\begin{tabular}{lcccccc}
\toprule
\textbf{Benchmark} &
\textbf{Targets} &
\textbf{Classes} &
\textbf{Hard-negative design} &
\textbf{Act:Dec ratio} &
\textbf{Learning protocol} &
\textbf{Fixed splits} \\
\midrule
DUD-E~\citep{mysinger2012directory}
& 102 & 8 & \xmark & 1:50 & \xmark & \xmark \\
MUV~\citep{rohrer2009maximum}
& 17 & 4 & \cmark & 1:500 & \xmark & \xmark \\
DEKOIS~2.0~\citep{bauer2013dekois}
& 81 & 5 & \xmark & 1:30 & \xmark & \xmark \\
LIT-PCBA~\citep{tran2020lit}
& 15 & 3 & $\dagger$ & varied & \xmark & \xmark \\
FS-Mol~\citep{stanley2021fsmol}
& varied & varied & \xmark & $\sim$balanced & \cmark & \cmark \\
\midrule
\textbf{TopU-LBVS (ours)}
& \textbf{93} & \textbf{7} & \cmark & \textbf{1:40} & \cmark & \cmark \\
\bottomrule
\end{tabular}%
}
\vspace{2pt}
\begin{minipage}{\textwidth}
\scriptsize
\raggedright
$\dagger$~LIT-PCBA uses experimentally confirmed inactives but does not
explicitly construct structurally challenging negatives around known actives.
MUV uses a distinct spatial-statistical design to reduce analogue bias and
artificial enrichment rather than TopU-style physicochemical and structural
matching.
\end{minipage}
\vspace{-.4in}
\end{table}

\textbf{Benchmark gap and TopU-LBVS.} \quad
The representative resources above address important components of LBVS
evaluation, but do not jointly provide a large multi-class target panel,
explicit hard-negative screening libraries, screening-relevant imbalance,
fixed supervised-learning protocols, and an accessible low-cost evaluation.
MUV is the closest classical precursor in its emphasis on avoiding easy
screening sets, while TopU-LBVS extends this hard-negative philosophy to a
larger, learning-oriented benchmark designed for contemporary molecular ML.
This gap matters increasingly as GNNs and molecular foundation models are
evaluated on benchmarks that may be substantially easier than the
hard-negative screening conditions they are meant to support.

TopU-LBVS covers 93 targets across 7 protein classes at a fixed 1:40
active-to-decoy ratio, with released data, fixed splits, and evaluation code.
\textbf{TopU-LBVS-full} evaluates generalization from historical
ChEMBL$^\ast$ SAR data to separate hard TopU screening libraries.
\textbf{TopU-LBVS-low} evaluates learning from limited labeled data within
the TopU hard-negative distribution. \textbf{TopU-LBVS-mini} provides a
seven-target entry point for rapid development and includes a paired
random-decoy control that holds the training data, validation protocol, test
actives, active count, active-to-decoy ratio, and metric fixed while replacing
only the TopU test decoys with randomly sampled ChEMBL$^\ast$ decoys.
Together, these protocols provide a controlled test of whether LBVS methods
remain effective under focused hard-negative screening conditions.

\vspace{-.15in}

\section{The TopU\textendash LBVS Benchmark}
\label{sec:topu}

\subsection{Target Panel and Protein Classes}
\label{sec:target_panel}

TopU-LBVS contains \textbf{93 protein targets} from ChEMBL~35 spanning
\textbf{7 protein classes}: cytochrome P450 enzymes, GPCRs, ion channels,
kinases, nuclear receptors, proteases, and other enzymes. Targets were selected
for biological diversity, ligand-discovery relevance, and sufficient assay
coverage.

The same panel supports the 93-target full and low-data protocols and the
seven-target mini benchmark. In TopU-LBVS-full, when too few target-specific
inactives remain after removing TopU compounds, we augment
\emph{training negatives only} with compounds from the same protein class.
Compounds with documented activity against the recipient target are excluded,
and augmentation never affects TopU test libraries. Full target counts and
the augmentation audit are given in Appendix~\ref{app:target-panel}.

\vspace{-.1in}

\subsection{ChEMBL$^\ast$ Curation}
\label{sec:chembl_star}

We curate ChEMBL~35 into a cleaned source dataset, denoted
ChEMBL$^\ast$. For each target, we retain reliable target and assay records,
standardize molecular structures, canonicalize compounds, resolve duplicates,
binarize activity, restrict molecules to common organic elements, and apply
drug-like physicochemical filters.

Because hard-negative construction can amplify label uncertainty, we audit
the inactive pool before decoy selection. Compounds with conflicting
active/inactive annotations are excluded from the decoy pool, and no final
TopU decoy is recorded as active for its own target under the benchmark
criterion. Detailed curation, label-consistency statistics, and leakage audits
are provided in Appendix~\ref{app:chembl_star}.

ChEMBL$^\ast$ supplies the historical SAR pool for TopU-LBVS-full after all
target-specific TopU compounds are removed. It also supplies the random-decoy
control in TopU-LBVS-mini, where TopU test decoys are replaced by randomly
sampled ChEMBL$^\ast$ decoys while preserving test actives, the
active-to-decoy ratio, and the training/validation protocol.

\vspace{-.1in}

\subsection{TopU Construction and Statistics}
\label{sec:topu_stats}

For each target, we construct a TopU library by pairing curated actives with
hard decoys from the corresponding ChEMBL$^\ast$ inactive pool. Candidate
decoy sets are optimized to reduce active--decoy distinguishability: a dynamic
search minimizes mutual information between activity labels and six
physicochemical descriptors plus MFP3 features while enforcing library size
and class ratio. A Random Forest is then used only as a
\emph{post-construction hardness audit}; easily separable candidate libraries
are rejected. Thus, the RF criterion works against simple Morgan/RF
separability rather than favoring the downstream Morgan-RF baseline, which is
trained independently from scratch on disjoint ChEMBL$^\ast$ data.

Each TopU library uses a fixed \textbf{1:40} active-to-decoy ratio. We treat
this as a focused hard-negative screening operating point rather than an
estimate of raw HTS prevalence. It is comparable to established benchmarks
such as DUD-E ($\sim$1:50) and DEKOIS~2.0 (1:30)~\citep{mysinger2012directory,bauer2013dekois},
while allowing stringent hard-negative constraints to be applied consistently
across all 93 targets. Increasing the nominal ratio substantially would often
require adding easier background negatives, changing the benchmark question.

Library sizes range from several hundred to more than ten thousand compounds.
The same TopU libraries serve as test sets in TopU-LBVS-full,
train/validation/test pools in TopU-LBVS-low, and the seven-target subset in
TopU-LBVS-mini. Detailed construction parameters, per-target statistics, and
post-construction robustness checks using alternative fingerprints and
molecular descriptors are provided in Appendix~\ref{app:topu-construction}.

\vspace{-.1in}

\section{Tasks and Evaluation Settings}
\label{sec:tasks-settings}

We formulate LBVS as target-specific ranking. For each target $t$ and
candidate compound $c$, a model outputs a score $s_t(c)$, which induces a
ranking over the candidate library. Metrics are computed per target and then
aggregated across targets.

\begin{figure*}[t]
\centering
\includegraphics[width=\textwidth]{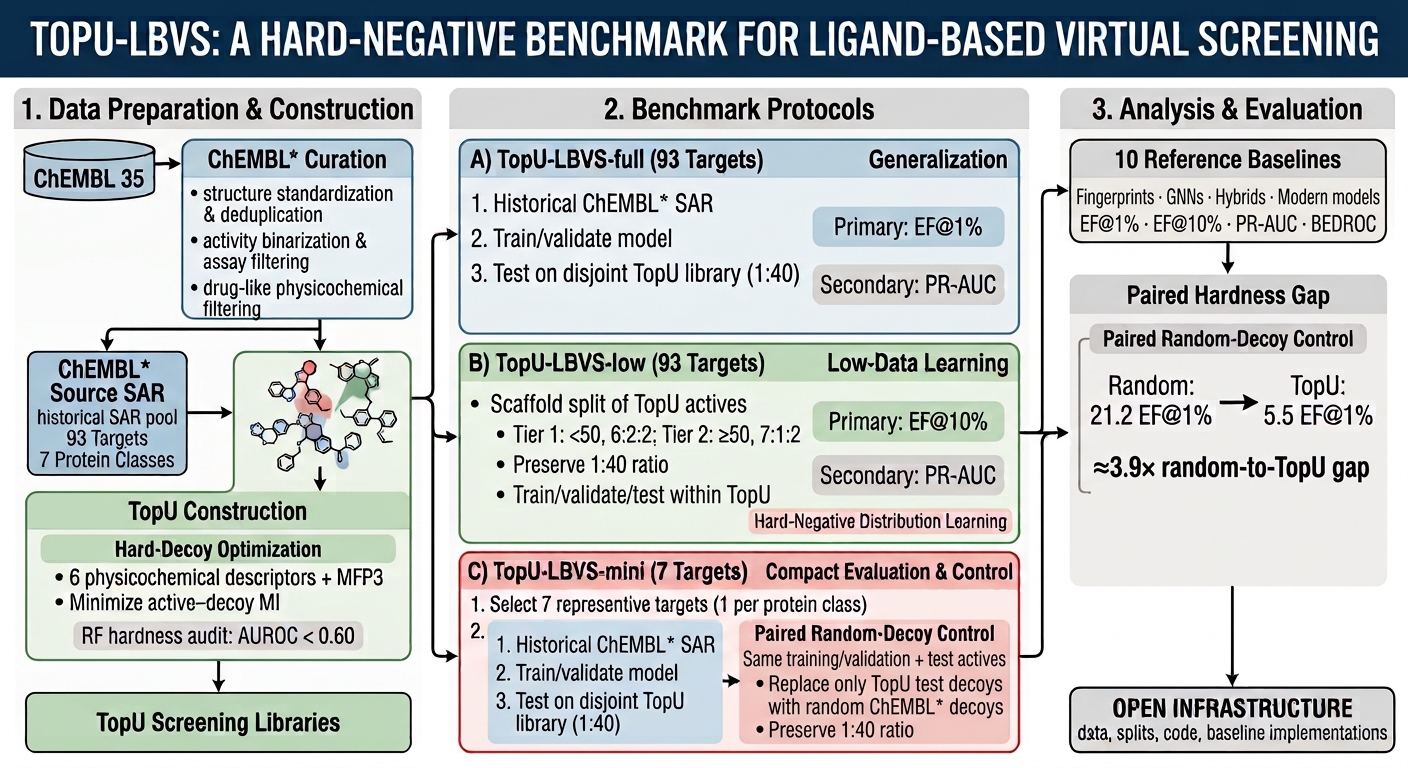}
\caption{\footnotesize \textbf{TopU-LBVS framework.}
ChEMBL~35 is curated into ChEMBL$^\ast$, which supplies historical SAR data
and the inactive pool for constructing TopU hard-negative libraries.
TopU-LBVS-full evaluates generalization from historical ChEMBL$^\ast$ SAR to
disjoint TopU screening libraries; TopU-LBVS-low evaluates learning from
limited labeled data within the TopU hard-negative distribution; and
TopU-LBVS-mini provides a seven-target protocol with a paired random-decoy
control that changes only the test decoys.}
\label{fig:framework}
\end{figure*}

TopU-LBVS contains 93 target-specific tasks across 7 protein classes. Each
target has a TopU library constructed at a fixed 1:40 active-to-decoy ratio
using property-matched, structurally challenging decoys
(Section~\ref{sec:topu}); library sizes range from roughly 400 to more than
10,000 compounds. We define three protocols: \textbf{TopU-LBVS-full} for
historical ChEMBL$^\ast \rightarrow$ TopU generalization,
\textbf{TopU-LBVS-low} for low-data learning within TopU hard-negative
libraries, and \textbf{TopU-LBVS-mini} for low-cost evaluation with a paired
random-decoy control. Figure~\ref{fig:framework} illustrates the benchmark
pipeline, and Table~\ref{tab:protocol_all_settings} summarizes the protocol
details.

\subsection{TopU-LBVS-full: ChEMBL$^\ast$ $\rightarrow$ TopU Evaluation}
\label{sec:setting1}

TopU-LBVS-full tests whether models trained on historical SAR data
generalize to hard-negative screening libraries. For each target, we remove
all target-specific TopU compounds from ChEMBL$^\ast$ to ensure zero
train-test overlap. The remaining actives and inactives form the training
pool, from which we construct a fixed 1:10 active-to-inactive dataset. When
too few target-specific inactives remain, we augment training negatives with
inactives from other targets in the same protein class using fixed seed 2026;
documented recipient-target actives are excluded and the exact selections
are released.

We reserve 15\% of this pool as a stratified random validation set,
preserving the 1:10 ratio, and use the remaining 85\% for training. The test
set is the full TopU library for the target, with a 1:40 active-to-decoy
ratio. The primary metric is EF@1\%; PR-AUC is used for validation and
reported as a secondary metric.

\subsection{TopU-LBVS-low: Low-Data TopU $\rightarrow$ TopU Evaluation}
\label{sec:setting2}

TopU-LBVS-low evaluates learning from limited labeled data when train,
validation, and test compounds all come from the same hard TopU distribution.
It is not intended to model a target-naive discovery campaign; rather, it
tests whether models can learn from limited labels within a focused
hard-negative candidate distribution. For each target, splits are drawn from
its TopU library while preserving the 1:40 active-to-decoy ratio.

Because active counts vary widely, targets with fewer than 50 TopU actives
(Tier~1) use a 6:2:2 active split, while targets with at least 50 actives
(Tier~2) use a 7:1:2 split. Active compounds are split by Bemis--Murcko
scaffold, while inactives are assigned to preserve the 1:40 ratio in each
split. Splits use seed 2026.

The primary metric is EF@10\%, which retains a direct screening-budget
interpretation while avoiding the instability of EF@1\% and EF@5\% for the
smallest low-data test libraries. PR-AUC is used for validation and reported
as a secondary metric; BEDROC provides a complementary continuous
early-recognition measure and is reported in the appendix. Full split rules
and per-target active counts are given in
Appendix~\ref{app:tasks-settings}.

\subsection{TopU-LBVS-mini: Compact Evaluation and Random-Decoy Control}
\label{sec:topu-mini}

TopU-LBVS-mini is a seven-target subset designed for rapid development,
debugging, and low-cost comparison. For each protein class, we select the
target with the largest number of TopU actives, yielding one representative
target per class. The selected targets and per-target counts are reported in
Appendix~\ref{app:tasks-settings}.

The TopU evaluation follows the same protocol as TopU-LBVS-full: models are
trained on ChEMBL$^\ast$ after removing all target-specific TopU compounds,
validated on the same fixed 15\% stratified split, and tested on the full
TopU library. EF@1\% is the primary metric, and PR-AUC is used for
validation.

TopU-LBVS-mini also includes a paired random-decoy control. For each mini
target, we keep the training data, validation protocol, test actives, and
1:40 test ratio fixed, but replace the TopU test decoys with randomly sampled
ChEMBL$^\ast$ decoys. The resulting gap isolates the effect of test-decoy
difficulty and quantifies how much random-decoy evaluation can overestimate
performance relative to hard-negative screening. Final benchmark claims
should be reported on TopU-LBVS-full; TopU-LBVS-mini is intended as an
accessible development and hardness-analysis protocol.

\begin{table*}[t]
\centering
\caption{\footnotesize
\textbf{Overview of TopU-LBVS evaluation protocols.}
TopU-LBVS-full and TopU-LBVS-low contain 93 target-specific tasks.
TopU-LBVS-mini follows the full protocol on seven representative targets,
one from each protein class. Ratios and splits are fixed for all baselines.}
\label{tab:protocol_all_settings}
\scriptsize
\resizebox{\textwidth}{!}{%
\begin{tabular}{lcccc}
\toprule
\textbf{Parameter} &
\textbf{TopU-LBVS-full} &
\textbf{TopU-LBVS-low, Tier 1} &
\textbf{TopU-LBVS-low, Tier 2} &
\textbf{TopU-LBVS-mini} \\
& ChEMBL$^\ast \rightarrow$ TopU &
TopU $\rightarrow$ TopU ($<50$ actives) &
TopU $\rightarrow$ TopU ($\geq 50$ actives) &
ChEMBL$^\ast \rightarrow$ TopU \\
\midrule
Training source & ChEMBL$^\ast$ & TopU & TopU & ChEMBL$^\ast$ \\
Test source & TopU & TopU & TopU & TopU \\
Number of targets & 93 & 46 & 47 & 7 \\
Train/val/test split & 85\% / 15\% / TopU test & 6:2:2 & 7:1:2 & 85\% / 15\% / paired test \\
Train ratio (act:dec) & 1:10 & 1:40 & 1:40 & 1:10 \\
Validation ratio (act:dec) & 1:10 & 1:40 & 1:40 & 1:10 \\
Test ratio (act:dec) & 1:40 & 1:40 & 1:40 & 1:40 \\
Random seed & 2026 & 2026 & 2026 & 2026 \\
Primary test metric & EF@1\% & EF@10\% & EF@10\% & EF@1\% \\
Secondary test metric & PR AUC & PR AUC & PR AUC & PR AUC \\
Validation metric & PR AUC (AP) & PR AUC (AP) & PR AUC (AP) & PR AUC (AP) \\
\bottomrule
\end{tabular}}
\vspace{2pt}
\begin{minipage}{\textwidth}
\scriptsize
In the random-control variant, TopU-LBVS-mini keeps the same training and
validation protocol but replaces TopU test decoys with randomly sampled
ChEMBL$^\ast$ decoys, preserving test active count and 1:40 ratio.
\end{minipage}
\vspace{-.2in}
\end{table*}

Across all protocols, enrichment factors are the primary test metrics:
EF@1\% for TopU-LBVS-full and TopU-LBVS-mini, and EF@10\% for
TopU-LBVS-low. PR-AUC is used for validation and reported as a secondary
metric. BEDROC, ROC-AUC, and additional enrichment factors are defined and
reported in Appendix~\ref{app:metrics}.

\section{Baselines}
\label{sec:baselines}

We evaluate ten reference baselines spanning three families:
(i) classical fingerprint-based LBVS methods, (ii) molecular GNNs and
fingerprint-augmented hybrids, and (iii) modern molecular models. Each method
produces a target-specific score used to rank the candidate library. Model
selection is performed by validation PR-AUC, and test performance is reported
using the primary metric of each protocol: EF@1\% for TopU-LBVS-full and
TopU-LBVS-mini, and EF@10\% for TopU-LBVS-low. Full architectures,
hyperparameter searches, optimization details, and training schedules are
provided in Appendix~\ref{app:baselines}.

\textbf{Classical fingerprint baselines.}
We include Morgan-RF and Tanimoto-NN, two standard LBVS references based on
2048-bit ECFP4-style Morgan fingerprints computed with RDKit
~\citep{rogers2010extended,landrum2006rdkit}. Morgan-RF trains a Random
Forest classifier, while Tanimoto-NN ranks compounds by their maximum
Tanimoto similarity to training actives~\citep{willett2006similarity}. These
methods provide strong, interpretable reference points for practical
ligand-based screening.

\textbf{GNN and hybrid baselines.}
We evaluate three atom-bond molecular graph architectures: GIN
~\citep{xu2019powerful}, GAT~\citep{velivckovic2018graph}, and GPS
~\citep{rampasek2022recipe}. Each is paired with a fingerprint-augmented
variant, GIN+FP, GAT+FP, and GPS+FP, which combines the learned graph
embedding with a learned projection of the Morgan fingerprint before
classification. These hybrids test whether learned molecular representations
add signal beyond classical substructure fingerprints under TopU
hard-negative evaluation.

\textbf{Modern molecular baselines.}
We include D-MPNN/Chemprop~\citep{yang2019analyzing,heid2024chemprop}, a
widely used directed message-passing model for molecular property prediction,
and MolFormer~\citep{ross2022large}, a pretrained SMILES Transformer. For
MolFormer, we fine-tune the publicly released 10
separately for each target.

All learned models are trained directly on the corresponding TopU-LBVS
training splits using class-weighted binary cross-entropy and model-specific
optimization schedules, with model selection based on validation PR-AUC.
Hyperparameters are selected on the benchmark training/validation splits, and
results are aggregated over three training seeds on the fixed data splits.
A separate model is trained for each target and protocol combination;
implementation and tuning details are given in Appendix~\ref{app:baselines}.

\section{Experimental Results}
\label{sec:results}

We evaluate ten baselines across the three TopU-LBVS protocols. Metrics are
computed per target from the induced ranking and then averaged by protein
class and overall. We report the primary metric for each setting in the main
text: EF@1\% for TopU-LBVS-full and TopU-LBVS-mini, and EF@10\% for
TopU-LBVS-low.

\textbf{Full per-target results and additional metrics} for TopU-LBVS-full and
TopU-LBVS-low, including EF@5\%, PR-AUC, ROC-AUC, BEDROC, and LogAUC, are
reported in the Appendix. \Cref{tab:appendix_index} provides a roadmap of all
appendix tables.

\subsection{TopU-LBVS-full}
\label{sec:results_full}

Table~\ref{tab:ef1pct_classwise_s1} reports class-averaged EF@1\% under the
ChEMBL$^*$$\to$TopU protocol across all 93 targets.

\textbf{Classical fingerprint methods are strongest in the full protocol.}\quad
Morgan-RF achieves the best overall EF@1\% (9.18), outperforming all GNN,
fingerprint-augmented hybrid, and modern learned baselines across the
93-target panel. Tanimoto-NN (7.58) also exceeds all pure GNN architectures.
The gap is substantial: the best pure GNN, GIN, reaches 4.47, while the
strongest modern learned baseline, D-MPNN, reaches 7.38. Thus, strong
performance under easier negative distributions does not necessarily transfer
to the TopU hard-negative regime. Although TopU construction reduces trivial
property and nearest-neighbour shortcuts, it does not remove all
substructure-level signal. Morgan-RF remains a strong reference baseline,
showing that fixed fingerprints retain useful discriminative information that
the evaluated learned representations do not consistently improve upon under
this target-specific training protocol.

\textbf{Fingerprint augmentation recovers part of the GNN deficit.}\quad
Adding Morgan fingerprints to learned graph embeddings consistently improves
GNN performance: GIN+FP, GAT+FP, and GPS+FP improve over their pure
counterparts by 2.17, 2.81, and 3.24 EF@1\% points, respectively. The three
hybrids cluster tightly around 6.6 overall, indicating that fingerprint
information contributes substantially across GNN backbones in this setting.
However, the hybrids remain about 2.5 EF@1\% points below Morgan-RF,
suggesting that the learned graph representations do not yet add reliable
signal beyond the fingerprint features under hard-negative evaluation.
Similarly, D-MPNN (7.38) and MolFormer (6.65), despite greater model
complexity and, for MolFormer, large-scale pretraining, do not surpass the
strongest fingerprint baseline. These results establish TopU-LBVS-full as a
stringent benchmark: under the evaluated training protocols, the learned
molecular models do not provide a consistent advantage over well-tuned fixed
fingerprints.

\begin{table}[t]
\centering
\caption{\textbf{TopU-LBVS-full: class-averaged EF@1\%.}
  Mean EF@1\% under the ChEMBL$^*$$\to$TopU protocol, averaged over targets
  within each protein class. The \textbf{Overall} column is the mean over all
  93 targets. Higher is better.
}
\label{tab:ef1pct_classwise_s1}
\resizebox{\textwidth}{!}{%
\begin{tabular}{ll ccccccc|c}
\toprule
\textbf{Group} & \textbf{Model}
  & \textbf{Cytochrome}
  & \textbf{GPCR}
  & \textbf{Ion Channel}
  & \textbf{Kinase}
  & \textbf{Nuc. Receptor}
  & \textbf{Misc.}
  & \textbf{Protease}
  & \textbf{Overall} \\
\midrule
\multirow{2}{*}{\textbf{Classical}}
  & Morgan-RF    & 14.18 & 5.21 & 12.41 & 9.15 & 11.12 & 10.01 & 6.66 & 9.18 \\
  & Tanimoto-NN  & 9.84  & 5.06 & 9.15  & 8.84 & 9.13  & 7.50  & 4.40 & 7.58 \\
\midrule
  & GIN          & 6.96  & 3.15 & 3.87  & 5.62 & 5.63  & 3.92  & 2.58 & 4.47 \\
  & GIN+FP       & 14.70 & 4.88 & 7.00  & 7.38 & 6.29  & 6.42  & 4.53 & 6.64 \\
\textbf{GNN \&}
  & GAT          & 6.03  & 2.82 & 3.68  & 5.13 & 3.98  & 3.42  & 1.10 & 3.76 \\
\textbf{Hybrid}
  & GAT+FP       & 13.87 & 4.78 & 6.41  & 7.89 & 7.33  & 6.05  & 3.04 & 6.57 \\
  & GPS          & 5.22  & 2.02 & 3.64  & 4.58 & 4.23  & 3.11  & 0.69 & 3.36 \\
  & GPS+FP       & 14.95 & 4.84 & 7.06  & 7.84 & 7.10  & 5.75 & 3.64 & 6.60 \\
\midrule
\multirow{2}{*}{\textbf{Modern}}
  & D-MPNN       & 13.54 & 4.81 & 5.04  & 8.62 & 8.80  & 7.52 & 4.34 & 7.38 \\
  & MolFormer    & 11.19 & 4.19 & 7.51  & 7.84 & 7.87  & 6.08 & 4.48 & 6.65 \\
\bottomrule
\end{tabular}%
}
\end{table}

\subsection{TopU-LBVS-low}
\label{sec:results_low}

Table~\ref{tab:ef10pct_classwise_s2} reports class-averaged EF@10\% under
the low-data TopU$\to$TopU protocol. EF@10\% is the primary metric for this
setting because EF@1\% and EF@5\% are unstable for the smallest Tier~1
libraries, where only a few actives appear within the top-ranked budget.

\begin{table}[ht]
\centering
\caption{\textbf{TopU-LBVS-low: class-averaged EF@10\%.}
  Mean EF@10\% under the TopU$\to$TopU few-shot protocol, averaged over
  targets within each protein class. Tier~1 contains targets with fewer than
  50 TopU actives; Tier~2 contains targets with at least 50 TopU actives.
  Higher is better.
}
\label{tab:ef10pct_classwise_s2}
\resizebox{\textwidth}{!}{%
\begin{tabular}{ll ccccccc|c}
\toprule
\textbf{Group} & \textbf{Model}
  & \textbf{Cytochrome}
  & \textbf{GPCR}
  & \textbf{Ion Channel}
  & \textbf{Kinase}
  & \textbf{Nuc. Receptor}
  & \textbf{Misc.}
  & \textbf{Protease}
  & \textbf{Overall} \\
\midrule
\multirow{2}{*}{\textbf{Classical}}
  & Morgan-RF    & 1.24 & 1.02 & 0.58 & 0.79 & 0.57 & 0.30 & 0.35 & 0.63 \\
  & Tanimoto-NN  & 1.13 & 0.53 & 0.58 & 0.67 & 0.51 & 0.38 & 0.05 & 0.50 \\
\midrule
  & GIN          & 1.15 & 1.46 & 0.75 & 0.96 & 0.98 & 0.90 & 1.09 & 1.03 \\
  & GIN+FP       & 0.91 & 0.85 & 0.63 & 1.00 & 0.83 & 0.86 & 0.42 & 0.84 \\
\textbf{GNN \&}
  & GAT          & 1.14 & 0.94 & 0.88 & 0.95 & 1.27 & 0.96 & 0.74 & 0.97 \\
\textbf{Hybrid}
  & GAT+FP       & 0.96 & 0.75 & 0.56 & 0.70 & 0.60 & 0.89 & 0.73 & 0.75 \\
  & GPS          & 1.39 & 0.95 & 1.36 & 0.99 & 1.18 & 1.11 & 0.76 & 1.05 \\
  & GPS+FP       & 0.66 & 0.72 & 1.00 & 0.84 & 1.01 & 0.90 & 0.74 & 0.85 \\
\midrule
\multirow{2}{*}{\textbf{Modern}}
  & D-MPNN       & 1.64 & 1.13 & 1.21 & 1.06 & 1.15 & 0.52 & 0.96 & 0.96 \\
  & MolFormer    & 1.72 & 1.04 & 1.06 & 1.09 & 1.24 & 1.22 & 0.75 & 1.12 \\
\midrule
\multicolumn{2}{l}{\textbf{Tier 1} ($<$50 actives, 46 targets)}
  & --   & 0.89 & 0.81 & 0.76 & 1.04 & 0.81 & 0.65 & 0.80 \\
\multicolumn{2}{l}{\textbf{Tier 2} ($\geq$50 actives, 47 targets)}
  & 1.19 & 0.95 & 1.07 & 0.98 & 0.84 & 0.78 & 0.70 & 0.94 \\
\bottomrule
\end{tabular}%
}
\end{table}

\textbf{Low-data hard-negative screening changes the model ranking.}\quad
The fingerprint baselines that dominate TopU-LBVS-full perform poorly in
this regime: Morgan-RF reaches 0.63 overall EF@10\%, and Tanimoto-NN reaches
0.50, both below random enrichment on most classes. With only a limited
number of labeled TopU actives for training, nearest-neighbour fingerprint
retrieval loses much of its advantage because the candidate library contains
structurally challenging decoys. Learned models become more competitive:
MolFormer obtains the best overall EF@10\% (1.12), followed by GPS (1.05)
and GIN (1.03). However, these gains remain modest. Since random ranking
corresponds to EF=1, the low-data hard-negative regime remains largely
unsolved, especially for Tier~1 targets, whose overall EF@10\% is only 0.80
across baselines.

\textbf{Fingerprint augmentation becomes protocol-dependent.}\quad
Unlike in TopU-LBVS-full, adding fingerprints does not consistently help in
the low-data setting. GIN+FP underperforms GIN (0.84 vs.\ 1.03), GAT+FP
underperforms GAT (0.75 vs.\ 0.97), and GPS+FP underperforms GPS
(0.85 vs.\ 1.05). This suggests that fingerprint similarity provides less
reliable additional signal when learning from a small labeled subset of an
already hard-negative candidate distribution. The contrast with
TopU-LBVS-full shows that robustness across data-rich and low-data
hard-negative regimes is non-trivial. The Tier~1/Tier~2 split further
highlights the data bottleneck: increasing from fewer than 50 to at least
50 TopU actives improves overall EF@10\% from 0.80 to 0.94, but even Tier~2
remains close to random enrichment.

To provide an accessible entry point for low-data molecular and graph
learning, we also introduce \textbf{TopU-LBVS-low-mini}, a compact 13-target
subset covering one Tier~1 and one Tier~2 target per protein class where
available; full details are given in Appendix~\ref{app:low_mini}.

\subsection{TopU-LBVS-mini: Random-Decoy Control}
\label{sec:results_mini}

TopU-LBVS-mini is a compact subset of TopU-LBVS-full, containing one
representative target from each protein class. It is designed as a
\textit{low-cost molecular benchmark for the broader machine learning
community} and also enables a paired random-decoy control.
Table~\ref{tab:mini_topu_vs_random} compares TopU hard-decoy (T) test sets
with randomly sampled ChEMBL$^*$ decoy (R) test sets across the seven mini
targets. The two settings use the same training data, validation protocol,
test actives, active count, and active-to-decoy ratio; only the test decoys
differ. Thus, the R$-$T gap isolates the effect of test-decoy difficulty.

\begin{table}[ht]
\centering
\caption{\textbf{TopU-LBVS-mini: TopU vs.\ random decoy test sets (EF@1\%).}
  Each target shows TopU hard decoys (T) and randomly sampled ChEMBL$^*$
  decoys (R). The gap between R and T measures how much random-decoy
  evaluation overestimates performance. The last two columns are averages
  across all seven targets. Higher is better.
}
\label{tab:mini_topu_vs_random}
\resizebox{\textwidth}{!}{%
\begin{tabular}{l cc| cc| cc |cc |cc |cc| cc| cc}
\toprule
& \multicolumn{2}{c}{\textbf{aa2ar}}
& \multicolumn{2}{c}{\textbf{aces}}
& \multicolumn{2}{c}{\textbf{cp3a4}}
& \multicolumn{2}{c}{\textbf{egfr}}
& \multicolumn{2}{c}{\textbf{esr1}}
& \multicolumn{2}{c}{\textbf{kcnh2}}
& \multicolumn{2}{c}{\textbf{thrb}}
& \multicolumn{2}{c}{\textbf{Average}} \\
\cmidrule(lr){2-3}\cmidrule(lr){4-5}\cmidrule(lr){6-7}
\cmidrule(lr){8-9}\cmidrule(lr){10-11}\cmidrule(lr){12-13}
\cmidrule(lr){14-15}\cmidrule(lr){16-17}
\textbf{Model} & T & R & T & R & T & R & T & R & T & R & T & R & T & R & T & R \\
\midrule
Morgan-RF   & 6.2 & 27.2 & 10.0 & 28.8 & 15.9 & 16.2 & 6.4 & 32.1 & 6.2 & 24.4 & 5.4 & 19.2 & 10.1 & 30.6 & \textbf{8.6} & \textbf{25.5} \\
Tanimoto-NN & 3.6 & 18.6 &  8.0 & 18.7 & 10.9 &  8.5 & 5.6 & 17.4 & 8.0 & 17.5 & 4.8 & 12.0 &  8.3 & 22.6 & \textbf{7.0} & \textbf{16.5} \\
\midrule
GIN         & 2.8 & 18.3 &  2.6 & 16.2 & 10.3 &  9.8 & 2.8 & 22.6 & 2.1 & 18.1 & 1.5 & 10.2 &  2.2 & 19.1 & \textbf{3.5} & \textbf{16.3} \\
GIN+FP      & 6.2 & 29.9 &  5.1 & 30.2 & 15.6 & 16.2 & 3.9 & 32.8 & 3.3 & 25.0 & 2.4 & 20.7 &  3.5 & 31.5 & \textbf{5.7} & \textbf{26.6} \\
GAT         & 4.1 & 13.4 &  2.8 & 16.5 &  9.0 &  9.4 & 2.0 & 16.9 & 2.4 & 15.4 & 2.0 & 10.2 &  2.2 & 18.1 & \textbf{3.5} & \textbf{14.3} \\
GAT+FP      & 5.2 & 29.7 &  4.8 & 32.2 & 14.3 & 18.2 & 2.0 & 31.7 & 4.2 & 23.5 & 2.5 & 20.5 &  3.5 & 31.8 & \textbf{5.2} & \textbf{26.8} \\
GPS         & 1.0 &  9.3 &  3.4 & 13.1 &  5.8 &  6.6 & 2.6 & 15.8 & 2.4 & 10.1 & 2.5 &  9.4 &  2.5 & 15.9 & \textbf{2.9} & \textbf{11.5} \\
GPS+FP      & 5.7 & 30.2 &  6.8 & 30.5 & 16.4 & 16.9 & 3.9 & 30.8 & 3.3 & 22.3 & 1.9 & 21.1 &  1.9 & 31.8 & \textbf{5.7} & \textbf{26.2} \\
\midrule
D-MPNN      & 3.1 & 28.1 &  6.5 & 30.5 & 14.9 & 15.8 & 7.4 & 28.0 & 5.9 & 24.7 & 9.1 & 20.1 &  4.5 & 29.6 & \textbf{7.3} & \textbf{25.3} \\
MolFormer   & 2.8 & 25.0 &  3.4 & 27.3 & 14.5 & 13.5 & 3.5 & 28.4 & 2.7 & 19.6 & 3.8 & 19.1 &  5.7 & 27.7 & \textbf{5.2} & \textbf{22.9} \\
\midrule
\textbf{Avg.} & \textbf{4.1} & \textbf{23.0} & \textbf{5.3} & \textbf{24.4} & \textbf{12.8} & \textbf{13.1} & \textbf{4.0} & \textbf{25.6} & \textbf{4.0} & \textbf{20.1} & \textbf{3.6} & \textbf{16.2} & \textbf{4.4} & \textbf{25.9} & \textbf{5.5} & \textbf{21.2} \\
\bottomrule
\end{tabular}%
}
\vspace{-.1in}
\end{table}

\textbf{Random-decoy evaluation substantially overestimates performance.}\quad
Across all ten models and seven targets, mean EF@1\% drops from 21.2 on
random decoys to 5.5 on TopU decoys, a nearly fourfold reduction caused
solely by changing the test negatives. The gap is highly consistent:
random-decoy EF@1\% exceeds TopU EF@1\% in 67 of 70 paired model-target
comparisons and appears across all model families. This controlled comparison
shows that strong performance against random negatives can largely reflect
separation from chemically dissimilar background compounds rather than
robust enrichment under hard-negative screening.

\textbf{The hardness gap is pervasive but target-dependent.}\quad
The largest average R$-$T gaps occur on egfr (21.6) and thrb (21.4), whereas
cp3a4 shows almost no average gap between the two decoy regimes. On cp3a4,
several models achieve similar performance under TopU and random decoys:
GAT+FP changes from 18.2 to 14.3, while GPS+FP achieves the best TopU
EF@1\% (16.4) with a nearly identical random-decoy result (16.9). Thus,
hard-negative difficulty varies substantially across targets even under a
fixed construction protocol. Overall, no baseline achieves consistently
strong enrichment across all seven TopU mini targets, making
TopU-LBVS-mini a low-cost diagnostic for the gap between apparent
random-decoy performance and hard-negative screening performance.

\subsection{Discussion, Limitations, and Release}
\label{sec:analysis}
\vspace{-.05in}

Taken together, the three protocols reveal a consistent picture: the evaluated
learned molecular models do not reliably outperform strong fingerprint
baselines under hard-negative screening, while the nearly 4$\times$ gap
exposed by the paired mini control shows how random-decoy evaluation can
substantially overestimate enrichment. In TopU-LBVS-low, learned models become
more competitive, but performance remains close to random enrichment on the
most data-limited tasks. TopU-LBVS is designed to make these regime-dependent
failures visible and measurable.

TopU-LBVS also has important limitations. It is retrospective and
ligand-based: no 3D binding-site information, docking scores, or prospective
experimental validation are included, and ChEMBL-derived inactive labels may
still contain assay noise or unobserved activity despite our consistency and
leakage audits. TopU operationalizes hard-negative similarity using
physicochemical descriptors and MFP3 features; alternative construction spaces
could yield different libraries, although post-construction robustness checks
using other molecular representations are reported in the Appendix. The 93
targets are not class-balanced and favor families with richer public assay
coverage. Finally, Tier~1 TopU-LBVS-low tasks contain very limited labeled
data, so individual-target enrichment estimates can have high variance and
class-level or overall results are more reliable.

All data, fixed splits, evaluation code, and baseline implementations are
released with the benchmark, including the ChEMBL$^*$ curation pipeline,
TopU construction scripts, train/validation/test splits, and implementations
of all ten baselines. A common evaluation script accepts ranked compound
lists and returns all reported metrics. Future benchmark versions will be
separately versioned so that results against the current release remain
directly comparable.

\section{Conclusion} \label{sec:conclusion}

We introduced TopU-LBVS, a multi-target benchmark for ligand-based virtual
screening under hard-negative conditions. Built from curated ChEMBL~35 data,
TopU-LBVS provides 93 target-specific tasks, fixed 1:40 active-to-decoy
libraries, and three protocols: ChEMBL$^\ast \rightarrow$ TopU evaluation,
low-data TopU $\rightarrow$ TopU evaluation, and a compact mini benchmark with
a paired random-decoy control. Across ten reference baselines, the benchmark
shows that performance under random-decoy evaluation can degrade sharply when
the same models are tested against TopU hard negatives; in the mini control,
mean EF@1\% is nearly 4$\times$ lower with TopU decoys than with random
ChEMBL$^\ast$ decoys. The evaluated learned molecular models also do not
consistently outperform strong fingerprint baselines under these conditions.
By releasing curated data, fixed splits, evaluation code, and reference
baselines, TopU-LBVS provides a reproducible testbed for assessing future LBVS
and molecular representation learning methods under focused hard-negative
screening conditions.

\section*{AI Use Statement}
Generative AI tools were used to improve the clarity and English of the
manuscript, assist in drafting and revising portions of the text, and 
design the pipeline figure. All AI-assisted text and
visual content were reviewed, edited, and approved by the authors. Generative
AI was not used to generate experimental data, conduct experiments, or
produce the reported numerical results. The authors take full responsibility
for the content of the paper.

\section*{Ethics Statement}
This work is entirely computational and uses publicly available molecular
bioactivity data from ChEMBL; it does not involve human participants,
personally identifiable information, animal experiments, or new wet-lab
experiments. TopU-LBVS is intended as a research benchmark for evaluating
ligand-based virtual screening methods and should not be interpreted as
providing experimentally validated binding predictions or guidance for
clinical or therapeutic use. As with other ChEMBL-derived resources, activity
and inactivity labels may contain assay noise or incomplete annotations, and
we document these limitations and apply consistency and leakage checks during
curation. The released data and code are intended to support reproducible
methodological research and comparison under hard-negative screening
conditions.

\section*{Reproducibility Statement}
To support reproducibility, we release the curated benchmark data, fixed
train/validation/test splits, target metadata, TopU construction code,
baseline implementations, and evaluation scripts. All reported experiments
use the released benchmark splits and specified evaluation protocols, with
model selection performed on validation data using PR-AUC. Results are
reported over multiple training seeds on fixed data splits, and the Appendix
provides implementation details, hyperparameters, per-target results, and
additional evaluation metrics. The released evaluation code accepts ranked
compound lists and computes the metrics reported in the paper, enabling direct
comparison with future methods. Code and data are publicly available through
the project repository and dataset release.

\clearpage
\bibliography{references}
\bibliographystyle{iclr2027_conference}


\clearpage
\appendix
\startcontents[appendices]
\section*{Appendix}
\printcontents[appendices]{}{1}{}

\section{TopU-LBVS Benchmark Details}

\subsection{ChEMBL$^*$ Curation}
\label{app:chembl_star}


We construct ChEMBL$^*$ from ChEMBL~35 to serve as the foundation for the TopU-LBVS. The goal of this curation pipeline is to remove noisy or assay artifacts, standardize molecular representations, and generate a high-confidence, consistent dataset of actives and inactives for subsequent benchmark construction.

\paragraph{Target-specific record collection.}
For each target of interest, we retrieve bioactivity data from ChEMBL~35 with reliable target assignments and assay annotations. We retain only measurements corresponding to four canonical potency endpoints ($\text{IC}_{50}$, $K_{\text{i}}$, $K_{\text{d}}$, $\text{EC}_{50}$) and exclude records with ambiguous or inconsistent target assignments. Where multiple assay types exist for a single target, we harmonize them into a unified target-level activity table.

\paragraph{Molecular standardization.}
All compounds are standardized prior to filtering and deduplication. This includes canonicalization of SMILES, removal of salts and solvent fragments when possible, and normalization of charges according to the RDKit-based preprocessing pipeline used in our benchmark code. Molecules that fail sanitization or cannot be converted to a valid canonical representation are discarded.

\paragraph{Activity binarization.}
For each target, we convert continuous activity measurements into binary labels (active versus inactive). These thresholds are chosen to be consistent within target and to preserve a reasonable separation between active and inactive compounds. To establish a definitive supervision signal for downstream benchmark construction, the retained compounds are strictly binarized based on their bioactivity: molecules with a pChEMBL value > 4.9 are designated as actives, while all remaining entries are categorized as inactives. The threshold of 4.9 is chosen to maximize target coverage while maintaining assay reliability. 

\paragraph{Deduplication and conflict resolution.}
After standardization, molecules are deduplicated by canonical SMILES. If multiple measurements map to the same standardized molecule for a given target, they are merged into a single entry. To account for inherent experimental variability, cases with conflicting activity labels are resolved by designating the molecule as active. This step ensures that the final benchmark does not contain duplicate structures with inconsistent supervision.

\paragraph{Removal of problematic compounds.}
To prevent the distortion of screening benchmarks and eliminate confounding factors, we remove problematic entries—such as those with unrealistically small or large structures, or other obvious cheminformatics anomalies—by applying rigorous physicochemical filtering rules. Specifically, compounds are retained only if they meet the following criteria: (i) atomic composition restricted to C, N, O, F, S, Cl, Br, I, P and H atoms; (ii) molecular weight between 150 and 800 Da; (iii) lipophilicity (LogP) between -3.0 and 6.0; (iv) fewer than 20 rotatable bonds; (v) fewer than 10 hydrogen bond acceptors and donors; (vi) a total formal charge constrained between -2.0 and +2.0.

\paragraph{Resulting curated dataset.}
The output of this pipeline is the cleaned target-specific dataset we denote by
\emph{ChEMBL$^*$}. Throughout the paper, “ChEMBL$^*$” refers exclusively to
this curated subset of ChEMBL~35. ChEMBL$^*$ serves as the source pool for
constructing TopU and for defining the historical SAR training data used in
Settings~1 and~3.

Tables~\ref{tab:per-target-a} and~\ref{tab:per-target-b} report, for each
target, the number of raw ChEMBL compounds, the number of compounds remaining
after ChEMBL$^*$ curation, and the final TopU counts. The gap between raw and
cleaned counts reflects the cumulative effect of assay filtering, removal of
problematic compounds, and canonical SMILES deduplication.

\subsection{TopU Construction and Statistics}
\label{app:topu-construction}

For each target, we construct a TopU library by pairing a selected set of
actives with hard decoys drawn from the corresponding ChEMBL$^*$ inactive
pool. The main design goal of TopU is to remove the trivial shortcuts that
often make public LBVS benchmarks overly easy. In particular, we aim to
ensure that decoys cannot be separated from actives by simple
physicochemical filters, while also remaining structurally similar enough to
known actives that nearest-neighbour fingerprint retrieval is no longer
sufficient. We intentionally restrict baselines to 2D and sequence-based methods to isolate the effect of representation learning under hard-negative regimes and avoid confounding performance with target-specific structural availability. 

To construct the TopU library for each target, we extract carefully curated actives and hard decoys from the corresponding ChEMBL$^*$ pool using a modified dynamic genetic algorithm. Unlike standard heuristic matching, this approach dynamically optimizes the size of the maximum matched subset while strictly enforcing a predefined active-to-decoy ratio. To ensure rigorous similarity in both physicochemical and topological spaces, molecules are encoded using a combination of six key physicochemical descriptors (Molecular Weight, LogP, Rotatable Bonds, Hydrogen Bond Acceptors, Hydrogen Bond Donors and Net Charge) and Morgan fingerprints 3 (MFP3). The algorithm's fitness function is designed to minimize the mutual information between these molecular features and bioactivity labels, thereby forcing the distributions of actives and decoys to become virtually indistinguishable. Furthermore, a Random Forest classifier is integrated as an additional validator, dynamically scaling the subset to guarantee that the cross-validation AUROC remains below 0.6. The Random Forest is configured with 100 estimators and a minimum of 10 samples per leaf, and is trained on the concatenated physicochemical and MFP3 features. The AUROC threshold of 0.6 is chosen to ensure that actives and decoys remain nearly indistinguishable while still allowing feasible library construction across diverse targets. The genetic algorithm is run with target‑specific parameter tuning: starting active sizes (5–10), maximum generations (100–700), patience (8–18), and step sizes for active increments (5–10). These ranges are chosen to balance search granularity against convergence speed for each target. Targets with highly overlapping active–decoy distributions require smaller starting sizes and step sizes for finer chemical space exploration, while larger values accelerate convergence for easier targets. Higher patience and generation limits are used for targets where the active pool is large, ensuring sufficient iterations to search feasible subsets. Collectively, these results demonstrate that TopU decoys are not random background molecules; rather, they are purposefully curated hard negatives.

Each TopU library is constructed at a fixed \textbf{1:40 active-to-decoy ratio}. This ratio is large enough to create a realistic early-enrichment problem, while still keeping library sizes manageable for systematic benchmarking across many targets. By construction, the TopU test libraries used in Setting 1 and Setting 2 therefore represent focused hard-negative screening collections rather than arbitrary random splits of ChEMBL.

The resulting TopU libraries vary considerably in size across targets,
reflecting natural variation in public assay coverage and the number of
surviving actives after curation. Across the 93 targets, TopU libraries range
from a few hundred compounds to well over ten thousand compounds per target.
This variation is an intentional feature of the benchmark: some tasks are
small and data-limited, while others support larger-scale screening, making
the benchmark more representative of real discovery settings.

The statistics in \Cref{tab:per-target-a,tab:per-target-b}  show the relationship between the original
ChEMBL counts, the cleaned ChEMBL$^*$ counts, and the final TopU counts for
each target. In most cases, the number of TopU actives is substantially
smaller than the number of cleaned actives, reflecting the stringency of the
TopU selection process. Similarly, although ChEMBL$^*$ may contain many
thousands of inactives for some targets, only a carefully chosen subset is
used in TopU, again to maintain the intended hard-negative regime.

Overall, TopU should be viewed not simply as another decoy set, but as a
target-specific benchmark library designed to stress-test LBVS methods under
conditions that are much closer to realistic prospective screening than
standard random-negative evaluation.

\subsection{Label, Leakage, and Representation Robustness Audits}
\label{app:topu-audits}

\paragraph{Inactive-label consistency.}
Because TopU intentionally selects chemically challenging inactives, we
performed an additional annotation-consistency audit before decoy selection.
Approximately 24.5\% of compounds in the initial inactive pool had multiple
retained ChEMBL measurements. Among these compounds, only 0.56\% showed
conflicting active/inactive annotations across measurements; all such
conflict-bearing compounds were removed from the inactive pool before TopU
construction. After molecular standardization and filtering, none of the
224,440 final TopU decoys is recorded as active against its own target under
the benchmark activity criterion. These checks remove known label conflicts,
although, as with any retrospective ChEMBL-derived benchmark, unmeasured
activity cannot be ruled out completely.

\paragraph{Training-negative augmentation audit.}
Same-class negative borrowing is used only for TopU-LBVS-full training pools
when insufficient target-specific inactives remain after removing TopU
compounds; it is never used in the TopU test libraries. Borrowing affects 15
of the 93 targets. After molecular standardization, only 0.039\% of borrowed
training-negative candidates overlapped compounds with documented activity
against the recipient target. These compounds were removed from the released
training splits.

\paragraph{Robustness to molecular representation.}
TopU construction uses physicochemical descriptors and MFP3 fingerprints to
operationalize hard-negative similarity. To test whether the resulting
difficulty is narrowly specific to this representation, we additionally
evaluated active--decoy separability on the fixed TopU libraries using
alternative representations, including ECFP4, ECFP6, RDKit fingerprints,
MACCS keys, and physicochemical descriptors. The great majority of targets
remain difficult under these alternative representations according to the
same separability criterion. This is a post-construction robustness audit,
rather than independent re-mining of the 93 libraries, and therefore does not
claim invariance to every possible hard-negative construction. It nevertheless
shows that TopU difficulty is not confined to the MFP3 representation used
during library construction.

\subsection{TopU-LBVS-low-mini: A Compact Low-Data Benchmark}
\label{app:low_mini}

\paragraph{Motivation.}
Evaluating new methods across all 93 targets of TopU-LBVS-low can be
computationally demanding, particularly during model development, ablation,
and hyperparameter analysis. TopU-LBVS-low-mini provides a compact,
standardised entry point to the same low-data hard-negative regime, following
the design philosophy of TopU-LBVS-mini. It preserves the fixed splits,
metrics, active-to-decoy ratio, and evaluation code of the full
TopU-LBVS-low protocol while substantially reducing evaluation cost.

\paragraph{Target selection.}
For each protein class, we select one Tier~1 target (fewer than 50 TopU
actives) and one Tier~2 target ($\geq 50$ TopU actives). Tier~2 targets are
chosen to match the TopU-LBVS-mini collection, creating a direct connection
between the two compact benchmarks. For Tier~1, we select the target with the
largest number of TopU actives below the 50-active threshold, improving the
stability of enrichment estimates while retaining the low-data setting.
Cytochrome P450 has no Tier~1 target because all four CYP targets contain at
least 50 TopU actives; it therefore contributes only its Tier~2
representative. This yields 13 targets across the 7 protein classes. The full
target list is given in Table~\ref{tab:few_mini_targets}.

\begin{table}[ht]
\centering
\caption{\textbf{TopU-LBVS-low-mini target collection.}
  One Tier~1 and one Tier~2 target per protein class.
  Tier~2 targets match the TopU-LBVS-mini collection.
  TopU active counts determine tier assignment ($<$50: Tier~1;
  $\geq$50: Tier~2). Cytochrome P450 contributes one target only
  (no Tier~1 CYP exists). Splits, metrics, and evaluation code are
  identical to TopU-LBVS-low.
}
\label{tab:few_mini_targets}
\resizebox{\textwidth}{!}{%
\begin{tabular}{l ll r ll r}
\toprule
& \multicolumn{3}{c}{\textbf{Tier 1} ($<$50 actives, 6:2:2 split)}
& \multicolumn{3}{c}{\textbf{Tier 2} ($\geq$50 actives, 7:1:2 split)} \\
\cmidrule(lr){2-4}\cmidrule(lr){5-7}
\textbf{Class}
  & \textbf{Target} & \textbf{Full name} & \textbf{Actives}
  & \textbf{Target} & \textbf{Full name} & \textbf{Actives} \\
\midrule
Cytochrome P450  & --     & (no Tier~1 target)               & --
                 & cp3a4 & Cytochrome P450 3A4 & 154\\
GPCR             & 5ht6r  & 5-hydroxytryptamine receptor 6   & 47
                 & aa2ar  & Adenosine A2a receptor           & 127 \\
Ion Channel      & 5ht3a  & Serotonin 3a (5-HT3a) receptor            & 49
                 & kcnh2  & hERG                             & 356 \\
Kinase           & akt2   & RAC-beta serine/threonine kinase & 48
                 & egfr   & Epidermal growth factor receptor & 153 \\
Nuclear Receptor & ppara  & Peroxisome proliferator-activated receptor $\alpha$ & 40
                 & esr1   & Estrogen receptor $\alpha$       & 113 \\
Misc.\ Enzymes   & hivint & HIV integrase                    & 48
                 & aces   & Acetylcholinesterase             & 118 \\
Protease         & fa10   & Coagulation factor Xa            & 43
                 & thrb   & Thrombin                         & 104 \\
\bottomrule
\end{tabular}%
}
\end{table}

\paragraph{Protocol.}
TopU-LBVS-low-mini is a strict subset of TopU-LBVS-low and uses the same
fixed TopU libraries and random seed 2026. Tier~1 targets use a 6:2:2
train/validation/test split and Tier~2 targets use a 7:1:2 split, with the
1:40 active-to-decoy ratio preserved throughout. The primary metric is
EF@10\%, with PR-AUC as the secondary metric; BEDROC and the remaining
ranking metrics are reported in Appendix~\ref{sec:additional_metrics}.
Because the low-mini targets use the corresponding fixed splits from
TopU-LBVS-low, their results are directly comparable to the same targets in
the full 93-target benchmark. TopU-LBVS-low-mini is intended for rapid
development and ablation; broad benchmark claims should be supported by
evaluation on the full TopU-LBVS-low panel.

\paragraph{Relationship to TopU-LBVS-mini.}
The Tier~2 targets of TopU-LBVS-low-mini coincide with the seven targets in
TopU-LBVS-mini and use the same underlying TopU libraries, but the protocols
differ in training regime. TopU-LBVS-mini trains on historical
ChEMBL$^\ast$ SAR data at a 1:10 training ratio, whereas TopU-LBVS-low trains
on a limited labeled subset of the TopU hard-negative distribution at 1:40.
The paired target collection therefore provides a controlled way to examine
whether a method remains effective when moving from a relatively data-rich
historical-SAR setting to a low-data hard-negative setting.

\begin{table*}[h!]
\centering
\scriptsize
\caption{Per target statistics for the TopU LBVS benchmark (part a). For each target we report counts of ChEMBL compounds before and after cleaning, as well as the number of actives and decoys in the TopU library.}
\label{tab:per-target-a}
\resizebox{\textwidth}{!}{%
\begin{tabular}{ccccc|cc|cc}
\toprule
 & & & \multicolumn{2}{c}{\textbf{ChEMBL (raw)}} & \multicolumn{2}{c}{\textbf{ChEMBL$^*$ (cleaned)}} & \multicolumn{2}{c}{\textbf{TopU}} \\
 \cmidrule(lr){4-5} \cmidrule(lr){6-7} \cmidrule(lr){8-9}
\textbf{Target} & \textbf{PDB} & \textbf{Class} & \textbf{active} & \textbf{decoy} & \textbf{active} & \textbf{decoy} & \textbf{active} & \textbf{decoy} \\
\midrule
cp2c9	&	1r9o	&	Cytochrome P450	&	1893	&	24225	&	1882	&	24052	&	119	&	4760	\\
cp1a2	&	2hi4	&	Cytochrome P450	&	1029	&	22581	&	1024	&	22415	&	136	&	5440	\\
cp3a4	&	3nxu	&	Cytochrome P450	&	3702	&	30526	&	3686	&	30235	&	154	&	6160	\\
cp2d6	&	4wnt	&	Cytochrome P450	&	1891	&	25216	&	1884	&	25028	&	139	&	5560	\\
\midrule
cxcr4	&	3odu	&	GPCR	&	919	&	855	&	907	&	853	&	15	&	600	\\
5ht6r	&	8jlz	&	GPCR	&	4168	&	2212	&	4131	&	2189	&	47	&	1880	\\
cnr2	&	5zty	&	GPCR	&	6645	&	3098	&	6604	&	3085	&	55	&	2200	\\
adrb1	&	7bu7	&	GPCR	&	1408	&	2964	&	1384	&	2920	&	62	&	2480	\\
hrh3	&	7f61	&	GPCR	&	4286	&	3051	&	4240	&	3023	&	65	&	2600	\\
oprk	&	9mql	&	GPCR	&	5064	&	3992	&	5010	&	3948	&	84	&	3360	\\
drd3	&	3pbl	&	GPCR	&	5351	&	4042	&	5304	&	3960	&	85	&	3400	\\
oprm	&	9mqi	&	GPCR	&	5546	&	5403	&	5489	&	5329	&	113	&	4520	\\
adrb2	&	3ny8	&	GPCR	&	1961	&	4648	&	1923	&	4575	&	102	&	4080	\\
oprd	&	6pt3	&	GPCR	&	4321	&	4824	&	4265	&	4771	&	103	&	4120	\\
drd2	&	6cm4	&	GPCR	&	7860	&	5197	&	7795	&	5119	&	113	&	4520	\\
aa3r	&	8x16	&	GPCR	&	4190	&	5798	&	4142	&	5756	&	120	&	4800	\\
aa2ar	&	3eml	&	GPCR	&	5194	&	6237	&	5159	&	6212	&	127	&	5080	\\
\midrule
trpa1	&	6x2j	&	Ion Channel	&	632	&	463	&	631	&	463	&	10	&	400	\\
cac1h	&	9ayh	&	Ion Channel	&	399	&	1421	&	398	&	1417	&	27	&	1080	\\
scn5a	&	6lqa	&	Ion Channel	&	766	&	2065	&	764	&	2061	&	43	&	1720	\\
5ht3a	&	8bla	&	Ion Channel	&	778	&	2307	&	772	&	2289	&	49	&	1960	\\
kcnh2	&	8zyq	&	Ion Channel	&	7110	&	16267	&	7067	&	16086	&	352	&	14080	\\
\midrule
wee1	&	3biz	&	Kinase	&	541	&	627	&	541	&	624	&	12	&	480	\\
tgfr1	&	3hmm	&	Kinase	&	1129	&	1030	&	1127	&	1028	&	22	&	880	\\
kpcb	&	2i0e	&	Kinase	&	620	&	1252	&	616	&	1235	&	26	&	1040	\\
mk10	&	2zdt	&	Kinase	&	1262	&	1653	&	1258	&	1648	&	37	&	1480	\\
fak1	&	3bz3	&	Kinase	&	1649	&	2123	&	1642	&	2099	&	44	&	1760	\\
mp2k1	&	3eqh	&	Kinase	&	889	&	2043	&	887	&	2021	&	44	&	1760	\\
rock1	&	2etr	&	Kinase	&	1667	&	2022	&	1649	&	2010	&	44	&	1760	\\
csf1r	&	3krj	&	Kinase	&	1884	&	2059	&	1867	&	2048	&	46	&	1840	\\
akt2	&	3d0e	&	Kinase	&	1021	&	2258	&	1016	&	2242	&	48	&	1920	\\
kit	&	3g0e	&	Kinase	&	1536	&	2526	&	1530	&	2517	&	53	&	2120	\\
braf	&	3d4q	&	Kinase	&	4066	&	2726	&	4060	&	2715	&	58	&	2320	\\
mapk2	&	3m2w	&	Kinase	&	869	&	2718	&	868	&	2700	&	59	&	2360	\\
plk1	&	2owb	&	Kinase	&	1253	&	25981	&	1252	&	25947	&	66	&	2640	\\
igf1r	&	2oj9	&	Kinase	&	2428	&	3289	&	2427	&	3273	&	66	&	2640	\\
mk01	&	2ojg	&	Kinase	&	3128	&	18279	&	3127	&	18105	&	68	&	2720	\\
met	&	3lq8	&	Kinase	&	3784	&	3622	&	3777	&	3597	&	76	&	3040	\\
akt1	&	3cqw	&	Kinase	&	3126	&	3744	&	3120	&	3706	&	77	&	3080	\\
fgfr1	&	3c4f	&	Kinase	&	2829	&	3534	&	2807	&	3509	&	77	&	3080	\\
jak2	&	3lpb	&	Kinase	&	6568	&	3777	&	6553	&	3757	&	84	&	3360	\\
mk14	&	2qd9	&	Kinase	&	4362	&	4056	&	4356	&	4032	&	86	&	3440	\\
lck	&	2of2	&	Kinase	&	1943	&	4375	&	1937	&	4346	&	93	&	3720	\\
cdk2	&	1h00	&	Kinase	&	2133	&	4403	&	2125	&	4358	&	97	&	3880	\\
abl1	&	2hzi	&	Kinase	&	2433	&	4548	&	2420	&	4496	&	97	&	3880	\\
src	&	3el8	&	Kinase	&	3011	&	4570	&	2992	&	4529	&	98	&	3920	\\
vgfr2	&	2p2i	&	Kinase	&	8825	&	6280	&	8751	&	6214	&	133	&	5320	\\
egfr	&	2rgp	&	Kinase	&	7575	&	7645	&	7509	&	7582	&	153	&	6120	\\
\bottomrule
\end{tabular}}
\end{table*}

\begin{table*}[h!]
\centering
\scriptsize
\caption{Per target statistics for the TopU LBVS benchmark (part b).}
\label{tab:per-target-b}
\resizebox{\textwidth}{!}{%
\begin{tabular}{ccccc|cc|cc}
\toprule
 & & & \multicolumn{2}{c}{\textbf{ChEMBL (raw)}} & \multicolumn{2}{c}{\textbf{ChEMBL$^*$ (cleaned)}} & \multicolumn{2}{c}{\textbf{TopU}} \\
 \cmidrule(lr){4-5} \cmidrule(lr){6-7} \cmidrule(lr){8-9}
\textbf{Target} & \textbf{PDB} & \textbf{Class} & \textbf{active} & \textbf{decoy} & \textbf{active} & \textbf{decoy} & \textbf{active} & \textbf{decoy} \\
\midrule
mcr	&	2aa2	&	Nuclear Receptor	&	944	&	490	&	943	&	489	&	10	&	400	\\
rxra	&	1mv9	&	Nuclear Receptor	&	610	&	1066	&	609	&	1063	&	20	&	800	\\
thb	&	1q4x	&	Nuclear Receptor	&	574	&	5734	&	574	&	5709	&	20	&	800	\\
ppard	&	2znp	&	Nuclear Receptor	&	1698	&	2055	&	1689	&	2038	&	35	&	1400	\\
ppara	&	2p54	&	Nuclear Receptor	&	2566	&	2365	&	2559	&	2341	&	40	&	1600	\\
gcr	&	3bqd	&	Nuclear Receptor	&	2488	&	2992	&	2484	&	2952	&	56	&	2240	\\
prgr	&	3kba	&	Nuclear Receptor	&	1877	&	2743	&	1874	&	2737	&	55	&	2200	\\
esr2	&	2fsz	&	Nuclear Receptor	&	2135	&	2936	&	2108	&	2796	&	56	&	2240	\\
andr	&	2am9	&	Nuclear Receptor	&	2502	&	3433	&	2492	&	3402	&	68	&	2720	\\
pparg	&	2gtk	&	Nuclear Receptor	&	3990	&	3996	&	3979	&	3971	&	73	&	2920	\\
esr1	&	1sj0	&	Nuclear Receptor	&	3268	&	5813	&	3226	&	5614	&	112	&	4480	\\
\midrule
hxk4	&	3f9m	&	Misc.\ Enzymes	&	1041	&	463	&	1041	&	463	&	10	&	400	\\
pygm	&	1c8k	&	Misc.\ Enzymes	&	149	&	687	&	148	&	685	&	10	&	400	\\
tysy	&	1syn	&	Misc.\ Enzymes	&	507	&	505	&	505	&	503	&	10	&	400	\\
hmdh	&	3ccw	&	Misc.\ Enzymes	&	269	&	1064	&	261	&	1058	&	11	&	440	\\
fnta	&	3e37	&	Misc.\ Enzymes	&	1745	&	652	&	1745	&	651	&	12	&	480	\\
ampc	&	1l2s	&	Misc.\ Enzymes	&	170	&	61861	&	121	&	61814	&	12	&	480	\\
dhi1	&	3frj	&	Misc.\ Enzymes	&	3295	&	675	&	3294	&	670	&	14	&	560	\\
inha	&	4trj	&	Misc.\ Enzymes	&	267	&	671	&	261	&	671	&	14	&	560	\\
pgh2	&	3ln1	&	Misc.\ Enzymes	&	462	&	746	&	462	&	743	&	15	&	600	\\
dyr	&	3nxo	&	Misc.\ Enzymes	&	978	&	921	&	975	&	918	&	19	&	760	\\
glcm	&	2v3f	&	Misc.\ Enzymes	&	415	&	12008	&	410	&	11966	&	21	&	840	\\
parp1	&	3l3m	&	Misc.\ Enzymes	&	3596	&	952	&	3589	&	947	&	20	&	800	\\
nos1	&	1qw6	&	Misc.\ Enzymes	&	567	&	1261	&	544	&	1246	&	19	&	760	\\
pyrd	&	1d3g	&	Misc.\ Enzymes	&	1093	&	1084	&	1089	&	1084	&	25	&	1000	\\
hdac2	&	3max	&	Misc.\ Enzymes	&	1995	&	1767	&	1978	&	1758	&	28	&	1120	\\
hdac8	&	3f07	&	Misc.\ Enzymes	&	2129	&	1465	&	2123	&	1451	&	29	&	1160	\\
pde5a	&	1udt	&	Misc.\ Enzymes	&	1872	&	1649	&	1866	&	1641	&	34	&	1360	\\
hivint	&	3nf7	&	Misc.\ Enzymes	&	2275	&	2462	&	2254	&	2447	&	48	&	1920	\\
cah2	&	1bcd	&	Misc.\ Enzymes	&	7822	&	3078	&	7540	&	2942	&	62	&	2480	\\
pgh1	&	2oyu	&	Misc.\ Enzymes	&	900	&	2984	&	898	&	2974	&	63	&	2520	\\
ptn1	&	2azr	&	Misc.\ Enzymes	&	2377	&	3377	&	2375	&	3352	&	64	&	2560	\\
aofb	&	7p4h	&	Misc.\ Enzymes	&	3417	&	3014	&	3403	&	2996	&	68	&	2720	\\
hivrt	&	3lan	&	Misc.\ Enzymes	&	4797	&	3676	&	4702	&	3576	&	75	&	3000	\\
aces	&	1e66	&	Misc.\ Enzymes	&	4701	&	5907	&	4575	&	5797	&	117	&	4680	\\
\midrule
ace	&	3bkl	&	Protease	&	581	&	529	&	578	&	529	&	10	&	400	\\
mmp13	&	830c	&	Protease	&	2588	&	762	&	2585	&	763	&	17	&	680	\\
try1	&	2ayw	&	Protease	&	1223	&	817	&	1165	&	797	&	16	&	640	\\
urok	&	1sqt	&	Protease	&	856	&	868	&	840	&	848	&	17	&	680	\\
ada17	&	2oi0	&	Protease	&	1728	&	877	&	1728	&	876	&	18	&	720	\\
hivpr	&	1xl2	&	Protease	&	4511	&	1556	&	4491	&	1550	&	25	&	1000	\\
dpp4	&	2i78	&	Protease	&	4009	&	1483	&	3875	&	1437	&	31	&	1240	\\
fa10	&	3kl6	&	Protease	&	5395	&	2188	&	5315	&	2126	&	43	&	1720	\\
bace1	&	3l5d	&	Protease	&	7195	&	2809	&	7158	&	2786	&	54	&	2160	\\
thrb	&	1ype	&	Protease	&	4661	&	5339	&	4434	&	5256	&	104	&	4160	\\
\bottomrule
\end{tabular}}
\end{table*}

\subsection{Target Panel and Protein Classes}
\label{app:target-panel}

TopU-LBVS is built on a panel of \textbf{93 protein targets} spanning
\textbf{7 major protein classes}: Cytochrome P450 enzymes, GPCRs, ion
channels, kinases, nuclear receptors, proteases, and other enzymes.
Targets labelled as ``Miscellaneous'' in ChEMBL were excluded because they
did not form a sufficiently coherent class for meaningful grouped analysis.

We begin from ChEMBL~35 and retain targets with substantial public
bioactivity data, reliable target annotations, and enough active and
inactive compounds to support both curation and benchmark construction.
From this initial pool, we keep only targets for which a challenging TopU
library can be constructed after cleaning and filtering, yielding the final
set of 93 benchmark tasks used throughout the paper.

The target panel is intentionally diverse rather than class balanced.
Kinases, GPCRs, and other enzymes contribute the largest number of tasks,
while ion channels, proteases, nuclear receptors, and cytochrome P450
enzymes add distinct activity landscapes and screening regimes. This
breadth is important because benchmark difficulty, decoy quality, and model
behaviour can vary substantially across target families. A benchmark
concentrated in only one or two families can therefore give a misleading
picture of model performance.

The class structure also supports downstream protocol design. In
Setting~1, after removing TopU compounds from ChEMBL$^*$, some targets
contain too few inactives to support the desired training ratio. In those
cases, we allow limited augmentation of the negative pool using inactives
drawn from other targets within the same protein class. Restricting this
augmentation to the same class preserves a biologically plausible negative
background while avoiding unrealistically small training sets for sparse
targets.

Tables~\ref{tab:per-target-a} and~\ref{tab:per-target-b} report the full
list of targets, their PDB identifiers, protein classes, and the numbers of
raw ChEMBL, ChEMBL$^*$, and TopU compounds per target. These statistics
highlight the wide range of task sizes represented in the benchmark, from
targets with only a few dozen TopU actives to targets with several hundred,
and from focused libraries of a few hundred compounds to much larger
hard-negative collections.

Across all 93 targets, the cleaned ChEMBL* data contain a median of 1 978 actives (min 121, max 8 751) and 2 737 inactives (min 463, max 61 814) per target. The TopU libraries contain a median of 53 actives (min 10, max 352) and a median total size of 2 173 compounds (min 410, max 14 432) per target. These per target statistics are provided
to document the benchmark and support downstream analysis; all train,
validation, and test splits are defined in the main text.

\section{Baseline Models}
\label{app:baselines}

We provide full details of all baseline models, including architectures,
training procedures, and hyperparameters. Table~\ref{tab:baselines_summary_app} summarises all baseline models, their
paradigm, molecular representation, and key implementation details.

\subsection{Category I: Classical Similarity-Based Baselines}
\label{app:baselines_classical}

These baselines match long-standing practice in LBVS and provide an
interpretable reference point: any complex learned model should at least
match or exceed their performance.

\paragraph{Morgan Fingerprint + Random Forest (Morgan-RF).}
We compute 2048-bit Morgan (ECFP4) fingerprints~\citep{rogers2010extended}
with radius $r=2$ for each compound using RDKit~\citep{landrum2006rdkit}, and
train a Random Forest classifier~\citep{breiman2001random} with balanced
class weights to account for the active-to-decoy imbalance. The predicted
probability of the active class is used as the ranking score. Morgan-RF is
included as a canonical fingerprint-based LBVS baseline, reflecting decades
of established practice in the field~\citep{willett2006similarity}.


\paragraph{Tanimoto Nearest-Neighbour Search (Tanimoto-NN).}
For each test compound, we compute the maximum Tanimoto coefficient against all
training actives using 2048-bit Morgan fingerprints, and use this maximum
similarity as the ranking score. No learning is involved; the method ranks
compounds purely by structural proximity to known actives. Tanimoto-NN is one
of the most widely deployed LBVS methods in practice~\citep{willett2006similarity}
and serves as a crucial reference point: a learned model that cannot outperform
Tanimoto-NN provides no practical benefit over off-the-shelf similarity search.
This baseline is particularly informative under our three-setting design, as its
performance gap relative to learned models is expected to change noticeably, and
often widen, on the TopU test sets (Settings 1 and 2), where hard negatives are
selected by a closely related topological similarity criterion.

\subsection{Category II: Graph Neural Networks and Hybrid Models}
\label{app:baselines_gnn}

Graph neural networks (GNNs) operate directly on the molecular graph, learning
hierarchical representations of atoms and bonds without relying on fixed
hand-crafted features. We evaluate three message-passing architectures and one
Graph Transformer, together with their fingerprint-augmented hybrid variants.

\paragraph{Molecular Graph Representation.}
Each molecule is represented as an undirected graph $\mathcal{G} = (V, E)$,
where nodes $v \in V$ correspond to heavy atoms and edges $e \in E$ correspond
to covalent bonds. Node features (9 dimensional) encode atomic number, degree,
formal charge, number of hydrogens, number of radical electrons,
hybridisation state, aromaticity, ring membership, and chirality. Edge features
(3 dimensional) encode bond type, conjugation, and ring membership.

\paragraph{Graph Isomorphism Network (GIN).}
GIN~\citep{xu2019powerful} uses sum aggregation with a learned
$\epsilon$-weighted self-loop, achieving the maximum expressive power of
1-dimensional Weisfeiler-Leman graph isomorphism tests among message-passing
GNNs. We use 5 message-passing layers, hidden dimension 256, global sum
pooling, and a 2-layer MLP classification head. All
hyperparameters follow common practice in molecular GNN benchmarks, with exact
values listed in Appendix~\ref{app:hyperparams}.

\paragraph{Graph Attention Network (GAT).}
GAT~\citep{velivckovic2018graph} replaces uniform neighbourhood aggregation
with multi-head attention, allowing the model to assign differential importance
to neighbours. We use 4 attention heads, 5 layers, hidden dimension 256, and
the same pooling and classification head as GIN (full details in
Appendix~\ref{app:hyperparams}).

\paragraph{Graph Transformer (GPS).}
GPS~\citep{rampasek2022recipe} combines local message-passing with global
Transformer-style self-attention, enabling the model to capture both local
chemical environments and long-range atomic dependencies within a single
unified architecture. We use 5 GPS layers with hidden dimension 256,
4 attention heads, and sum pooling.

\begin{table}[t]
\centering
\caption{
    \textbf{Summary of all baseline models.} ``Group'' indicates the model family
    (Similarity based, GNN / Hybrid, Modern pretrained).
    All neural models use BCE loss with positive class weighting and
    Adam optimisation. Validation metric is PR-AUC for all models.
}
\label{tab:baselines_summary_app}
\resizebox{\textwidth}{!}{%
\begin{tabular}{lllll}
\toprule
\textbf{Group} & \textbf{Model} & \textbf{Paradigm} &
\textbf{Representation} & \textbf{Parameters}  \\
\midrule
\multirow{2}{*}{\textbf{Naive}}
    & Morgan-RF \citep{rogers2010extended}
    & Fingerprint + Ensemble
    & Morgan FP (2048-bit, $r=2$)
    & up to 500 trees
    \\
    & Tanimoto-NN \citep{willett2006similarity}
    & Similarity search
    & Morgan FP (2048-bit, $r=2$)
    & --
    \\
\midrule
    & GIN \citep{xu2019powerful}
    & Message passing (sum)
    & Molecular graph
    & 5L, $d=256$
    \\
    & GIN+FP
    & Message passing + FP
    & Graph $\|$ Morgan FP
    & 5L, $d=256$
    \\
\textbf{GNN}  \&  & GAT \citep{velivckovic2018graph}
    & Attention weighted MP
    & Molecular graph
    & 5L, $d=256$, 4 heads
    \\
\textbf{Hybrid}    & GAT+FP
    & Attention weighted MP + FP
    & Graph $\|$ Morgan FP
    & 5L, $d=256$, 4 heads
    \\
    & GPS \citep{rampasek2022recipe}
    & Graph Transformer
    & Molecular graph
    & 5L, $d=256$, 4 heads
    \\
    & GPS+FP
    & Graph Transformer + FP
    & Graph $\|$ Morgan FP
    & 5L, $d=256$, 4 heads
    \\
\midrule
\textbf{Modern}
    & D-MPNN \citep{heid2024chemprop}
    & Directed message passing
    & Graph + RDKit (200-dim)
    & 3 steps, $d=300$
    \\
\textbf{pre-trained }   
    & MolFormer \citep{ross2022large}
    & Pretrained SMILES Transformer
    & SMILES (1.1B pretraining)
    & Fine tuned
    \\
    
    \bottomrule
\end{tabular}%
}
\end{table}

\paragraph{Hybrid Variants (GNN + Morgan FP).}

For each of the three GNN architectures above, we additionally evaluate a hybrid model combining the learned graph embedding with a 2048-bit Morgan fingerprint via a late-fusion MLP. The GNN encoder produces a graph-level embedding $\mathbf{h}_{\text{graph}}$, while the Morgan fingerprint $\mathbf{fp}_{\text{Morgan}} \in \{0,1\}^{2048}$ is passed
through a separate two-layer MLP with ReLU activation and dropout to
obtain a fingerprint embedding $\mathbf{h}_{\text{fp}}$ of the same
dimensionality as $\mathbf{h}_{\text{graph}}$. The two embeddings are
concatenated and fed to a two-layer classification head:
\[
    \mathbf{h}_{\text{hybrid}} = \left[\mathbf{h}_{\text{graph}}\;\|\;
    \mathbf{h}_{\text{fp}}\right],
    \qquad
    \hat{y} \;=\; \mathrm{MLP}_{\text{cls}}\!\left(\mathbf{h}_{\text{hybrid}}\right).
\]

Each hybrid variant introduces an
additional fingerprint-projection MLP and a wider classification head
whose input is the concatenated graph and fingerprint embeddings. This late-fusion design allows the model to leverage both
the expressive learned graph representations and the well-validated
substructural information encoded by Morgan fingerprints, with the list of hyperparameters reported in Appendix~\ref{app:hyperparams}. We denote the
hybrid variants GIN+FP, GAT+FP, and GPS+FP.

\paragraph{Training Details (All GNN Models).}
All GNN models are trained end-to-end using binary cross-entropy loss with
positive class weighting:
\quad $  w_{+} = \dfrac{N_{\text{decoy}}}{N_{\text{active}}}$ \quad 
to compensate for class imbalance. We use the Adam
optimiser~\citep{kingma2014adam} with learning rate $10^{-3}$, weight decay
$10^{-4}$, batch size 64, and a maximum of 100 epochs with early stopping based
on validation PR-AUC (patience $= 20$ epochs). The learning rate is reduced by
a factor of 0.5 when validation PR-AUC plateaus for 5 consecutive epochs. A
complete list of hyperparameters and training schedules is given in
Appendix~\ref{app:hyperparams}.

\subsection{Category III: Modern Pretrained Models}
\label{app:baselines_modern}

Modern pretrained models leverage large-scale self-supervised pretraining on
unlabelled molecular data, followed by task-specific fine-tuning on the target
dataset. We select two models representing distinct and complementary
pretraining paradigms that are among the most widely adopted in contemporary
molecular property prediction.

\paragraph{D-MPNN (Chemprop).}
The Directed Message Passing Neural Network (D-MPNN), implemented in
Chemprop~\citep{yang2019analyzing, heid2024chemprop}, propagates messages
along directed edges rather than nodes, mitigating oversmoothing and enabling
more precise encoding of local chemical environments. The molecular embedding
is obtained by aggregating learned directed bond representations, and passed to
a feed-forward network (FFN) for classification. Unlike the GNN models in
Category II, the D-MPNN additionally accepts an optional vector of global
molecular features; we augment the molecular embedding with 200-dimensional
RDKit physicochemical descriptors, following the standard recommended
configuration~\citep{yang2019analyzing}. We use the Chemprop v2 default
architecture: 3 message-passing steps, hidden dimension 300, depth-2 FFN,
dropout $p=0.0$, and ReLU activations. Models are trained with binary
cross-entropy loss and the Adam optimiser for up to 50 epochs with early
stopping on validation PR-AUC (patience $= 10$ epochs). D-MPNN is included as a
widely adopted and empirically validated modern LBVS baseline, with successful
applications spanning antibiotic discovery~\citep{stokes2020deep} and
large-scale bioactivity prediction.

\paragraph{MolFormer (Fine-tuned).}
MolFormer~\citep{ross2022large} is a large-scale chemical language model
pretrained on over 1.1 billion SMILES strings from PubChem and ZINC using a
masked language modelling objective, with linear attention and rotary position
embeddings for computational efficiency. We use the publicly released
\texttt{ibm/MoLFormer-XL-both-10pct} checkpoint. For each target, the
pretrained encoder produces token-level representations, which are pooled
into a single molecular embedding by attention-mask-weighted mean pooling over
the last hidden state. following the IBM
MolFormer fine-tuning reference. Models are trained with binary cross-entropy
loss using positive-class weighting ($\text{pos\_weight} = n_{\text{neg}} / n_{\text{pos}}$)
to compensate for class imbalance, and optimized end-to-end with the LAMB
optimizer at a single learning rate of $3 \times 10^{-5}$ (no weight decay).
SMILES strings are canonicalized with \texttt{isomericSmiles=False} and
truncated to $202$ tokens (MolFormer's maximum position embedding). We use a
batch size of $32$, a maximum of $30$ fine-tuning epochs, and early stopping
on validation PR-AUC (patience $= 5$ epochs); the checkpoint with the highest
validation PR-AUC is retained. MolFormer serves as the representative
large-scale pretrained transformer baseline, allowing us to test whether
chemical foundation model pretraining improves LBVS performance in our
hard-negative and data-limited regimes.


\section{Hyperparameters and Implementation Details}
\label{app:hyperparams}

This section provides complete hyperparameter configurations and
implementation details for all ten baseline models. All models are
implemented in PyTorch~\citep{paszke2019pytorch}, with GNN models using
PyTorch Geometric~\citep{fey2019fast} and fingerprint computation using
RDKit~\citep{landrum2006rdkit}. Chemprop v2 is used for D-MPNN, and the
pretrained MolFormer weights are loaded from the official HuggingFace
repository.\footnote{\url{https://huggingface.co/ibm/MoLFormer-XL-both-10pct}}
All split files, model configurations, training scripts, and random seeds
are released alongside the benchmark to ensure full reproducibility.

\subsection{Classical Baselines}

\paragraph{Morgan-RF.}
Each compound is represented by a 2048-bit Morgan (ECFP4) fingerprint with
radius $r=2$. We train a scikit-learn \texttt{RandomForestClassifier} with
\texttt{max\_features="sqrt"} and balanced class weights
$w_j = n / (2 \cdot n_j)$ to account for the active-to-decoy imbalance. The
random seed (2026, 2027, 2028) controls both bootstrap sampling and
feature selection. Trees are grown incrementally with early stopping on
validation PR-AUC: starting from 10 trees, we add 10 trees per step and
stop after 5 consecutive steps without PR-AUC improvement, with a hard cap
of 500 trees. The optimal tree count $n^{\star}$ is then refit from scratch
to obtain the final model. The predicted probability of the active class is
used as the ranking score.

\paragraph{Tanimoto-NN.}
Each compound is represented by a 2048-bit Morgan fingerprint identical to
the Morgan-RF representation. The model stores fingerprints of training
actives only; the validation set and decoys are not used. At inference, we
compute the full $(n_{\text{test}} \times n_{\text{train-actives}})$
Tanimoto similarity matrix and use the row-wise maximum as the ranking
score. Tanimoto similarity is defined as
$T(a,b) = (a \cdot b) / (\lVert a \rVert_1 + \lVert b \rVert_1 - a \cdot b)$.
No model selection or hyperparameter tuning is performed. Tanimoto-NN has
no random state, but each seed produces a different training split, so
the three-seed variance reported in the main tables reflects split-induced
variance only.

\subsection{GNN and Hybrid Models}

All GNN models share the training infrastructure described below unless
otherwise noted.

\paragraph{Shared training setup.}
All GNN models are trained end-to-end as binary classifiers using
\texttt{BCEWithLogitsLoss} with positive-class weighting
\[
w_{+}=\frac{N_{\text{decoy}}}{N_{\text{active}}},
\]
to compensate for class imbalance in each training split. We use the Adam
optimizer with learning rate $10^{-3}$, weight decay $10^{-4}$, batch size 64,
and a maximum of 100 epochs. Hyperparameter selection is performed with seed
2026 on each target. Within each run, model
selection uses validation PR-AUC: after every epoch, PR-AUC is computed on the
held-out validation split, the checkpoint with the best validation PR-AUC is
saved, and training stops early if validation PR-AUC does not improve for 20
consecutive epochs (patience $=20$). We additionally apply
\texttt{ReduceLROnPlateau} on validation PR-AUC (\texttt{mode=max}), reducing
the learning rate by a factor of 0.5 after 5 consecutive non-improving epochs..

\paragraph{Backbone-specific architectures.}
Under the shared training protocol above, all three graph backbones (GIN, GAT, and GPS) use five message-passing layers with hidden width 256, sum pooling, projection to a 32-dimensional graph embedding, and a two-layer prediction MLP with ReLU and dropout between the two linear layers.

These models differ only in the message-passing operator. GIN uses Graph Isomorphism Network layers with a two-layer update MLP and a learnable $\epsilon$ term initialized at zero (i.e., sum-based neighborhood aggregation with a scaled self-contribution), comprising a total of 672,358 trainable parameters. GAT uses four-headed graph attention in each layer, with head outputs averaged rather than concatenated so the hidden dimensionality remains 256 throughout, and it has 1,336,673 trainable parameters in total. GPS uses GPS layers that combine a local GIN-style message-passing module (matching the GIN operator above) with a global four-head self-attention module, followed by the standard feed-forward and normalization sublayers of GPS blocks, yielding 3,310,438 trainable parameters. Dropout for all backbones is tuned over $p\in\{0.3,\,0.5\}$.

\paragraph{Hybrid variants (GIN+FP, GAT+FP, GPS+FP).} Each GNN encoder produces a 32-dimensional graph embedding which, for hybrid variants, is concatenated  with a 32-dimensional fingerprint embedding. This embedding is obtained by passing
the 2048-bit Morgan fingerprint through a two-layer projection MLP with ReLU
and dropout, yielding a 64-dimensional joint representation. The classification
head is widened accordingly for an input dimension of 64 instead of 32. Its hidden dimension $d_{\text{fuse}}$ is tuned over
$\{64, 128, 256\}$. The fingerprint projector hidden width
$d_{\text{fp}}$ is tuned over $\{128, 256, 512\}$. All other settings are inherited
unchanged from the corresponding pure-GNN variant from the shared training setup, and the tuning values for dropout, like for the backbones, remains over $p\in\{0.3,\,0.5\}$.

\subsection{Modern Pretrained Models}

\paragraph{D-MPNN (Chemprop).}
We use the Chemprop v2 reference architecture ~\citep{heid2024chemprop}. Each
molecule is represented as a directed atom-bond graph; messages are propagated
along directed edges for $T = 3$ message-passing steps with hidden dimension
$300$ and ReLU activations. The molecular embedding is concatenated with
$200$-dimensional RDKit physicochemical descriptors (computed via the standard
\texttt{normalized\_features} generator and standardized with a per-fold
\texttt{StandardScaler}) before being passed to a $2$-layer feed-forward
network of hidden size $300$ with dropout $p = 0.0$. Models are trained with
binary cross-entropy loss using positive-class weighting and optimized with
Adam~\citep{kingma2014adam} under the Chemprop v2 default Noam learning-rate
schedule (initial LR $10^{-4}$, peak LR $10^{-3}$ after $2$ warmup epochs,
final LR $10^{-4}$). We use batch size $64$, a maximum of $50$ epochs, and
early stopping on validation PR-AUC. Following Chemprop v2, edge updates use
preactivation initial hidden states. The checkpoint with the highest
validation PR-AUC is retained.

\paragraph{MolFormer.}
We fine-tune the publicly released \texttt{ibm/MoLFormer-XL-both-10pct}
checkpoint, loaded via the HuggingFace \texttt{transformers} library
(\texttt{trust\_remote\_code=True} required by the IBM custom modeling code).
SMILES are canonicalized with \texttt{isomericSmiles=False} and tokenized with
the official MolFormer tokenizer; sequences are truncated to the maximum
position embedding length of $202$ tokens. The encoder produces a
$768$-dimensional pooled molecular embedding via attention-mask-weighted mean
pooling over the last hidden state. A three-layer MLP head with two residual
skip connections (hidden dimensions $768/768/768$, output dimension $1$) is
appended, matching the IBM reference fine-tuning architecture. All parameters
are optimized end-to-end with the LAMB optimizer at a
single learning rate of $3 \times 10^{-5}$ and weight decay $0.0$, with
parameters grouped into a ``decay'' group (Linear weights) and a ``no\_decay''
group (LayerNorm/Embedding weights and all biases) following the IBM
reference. We use batch size $32$, a maximum of $30$ fine-tuning epochs, and
early stopping on validation PR-AUC with patience $= 5$ epochs. The checkpoint
with the highest validation PR-AUC is retained.
\subsection{Computational Resources}

All experiments were run on a shared university HPC cluster using single-GPU
training runs throughout. The cluster provides a heterogeneous mix of NVIDIA
GPUs, including A100 (40\,GB and 80\,GB), A100 MIG slices, H100 NVL, H200 NVL,
A30, V100 (16\,GB), P100 (16\,GB), RTX~3090 (24\,GB), and RTX Pro 6000.
MolFormer fine-tuning was preferentially scheduled on A100 (80\,GB) or H100
nodes due to its higher memory footprint; smaller GNN and fingerprint
baselines ran on whatever GPU was available.

Per-target wall-clock time varies substantially with model and target size.
Classical fingerprint baselines (Morgan-RF, Tanimoto-NN) complete in
under a minute per target. Pure GNN backbones (GIN, GAT, GPS) and their
fingerprint-augmented variants typically train in $5$--$30$~minutes per
target. D-MPNN takes roughly $10$--$60$~minutes per target depending on
dataset size. MolFormer fine-tuning is the most expensive baseline,
requiring approximately $15$--$90$~minutes per target on an A100 (80\,GB).
The full benchmark (all $93$ targets, $10$ models, $3$ seeds, both
Settings~1 and~2) is on the order of $2{,}000$--$3{,}000$ GPU-hours total,
dominated by D-MPNN and MolFormer fine-tuning.

\subsection*{Implementation Notes}

Neural baselines are implemented in PyTorch~\citep{paszke2019pytorch} with
GNN models using PyTorch Geometric~\citep{fey2019fast}. Classical baselines
(Morgan-RF, Tanimoto-NN) use scikit-learn and NumPy; fingerprint computation
uses RDKit~\citep{landrum2006rdkit}. Chemprop v2 is used for D-MPNN, and the
pretrained MolFormer weights are loaded from the official HuggingFace
repository.\footnote{\url{https://huggingface.co/ibm/MoLFormer-XL-both-10pct}}
All split files, model configurations, training scripts, and random seeds are
released alongside the benchmark to ensure full reproducibility.
A complete description of all hyperparameters is provided in
Appendix~\ref{app:hyperparams}.

\section{Metric Definitions}
\label{app:metrics}

In all settings, models produce a real valued score for each test compound of a given target.
Sorting compounds in descending order of this score induces a ranking.
All metrics below are computed per target from this ranking, then aggregated across targets as described at the end of this section.

Let
\begin{itemize}
  \item $N$ be the number of test compounds for a given target,
  \item $N^{+}$ be the number of actives among these $N$ compounds,
  \item $N^{-} = N - N^{+}$ be the number of decoys,
  \item $\pi = N^{+} / N$ be the fraction of actives in the test set.
\end{itemize}

Let the ranked list be $c_1, c_2, \dots, c_N$, where $c_1$ is the top ranked compound.
Let $y_i \in \{0,1\}$ denote the label of $c_i$ ($1$ for active, $0$ for decoy).

\subsection{Enrichment Factor at $x$ percent (EF@x\%)}

For a given fraction $x \in (0,1]$, define $k = \lceil x N \rceil$ as the number of top ranked compounds considered.
Let
\quad $H(x) = \sum_{i=1}^{k} y_i$ \quad  be the number of actives found among the top $k$ compounds.

The enrichment factor at $x$ percent, EF@x\%, is defined as
\quad $
\textbf{EF@x\%}\;=\; 
\dfrac{\dfrac{H(x)}{k}}{\dfrac{N^{+}}{N}}
\;=\;
\dfrac{H(x) \, N}{k \, N^{+}}.$

Under random ranking, $\mathbb{E}[H(x)] = k \, N^{+} / N$, so $\text{EF}@x\% = 1$.
Values greater than $1$ indicate enrichment of actives relative to random.

In this work we use EF@1\% and EF@10\%.
For smaller libraries, EF@1\% can be coarse because $k$ is small, but it still reflects the very top of the ranking.
EF@10\% provides a more stable summary of early enrichment.

\subsection{Precision Recall AUC (PR AUC)}

For a threshold $\tau$ on the model score, let
\[
\text{TP}(\tau),\quad \text{FP}(\tau),\quad \text{FN}(\tau)
\]
denote the numbers of true positives, false positives, and false negatives respectively.
Precision and recall at threshold $\tau$ are
\[
\text{Precision}(\tau) = 
\frac{\text{TP}(\tau)}{\text{TP}(\tau) + \text{FP}(\tau)},\qquad
\text{Recall}(\tau) =
\frac{\text{TP}(\tau)}{\text{TP}(\tau) + \text{FN}(\tau)}.
\]

Sweeping $\tau$ from $+\infty$ to $-\infty$ traces out the precision recall curve.
The precision recall area under the curve (PR AUC), also known as average precision, is defined as the area under this curve
\[
\text{PR AUC} = \int_{0}^{1} \text{Precision}(r) \, \mathrm{d}r,
\]
where $r$ denotes recall.
In practice we approximate this integral using the standard stepwise interpolation over the discrete set of thresholds induced by the ranking.
PR AUC is more informative than ROC AUC in the strongly imbalanced regime relevant for LBVS.

\subsection{Receiver Operating Characteristic AUC (ROC AUC)}

For the same threshold $\tau$, define the true positive rate and false positive rate as
\[
\text{TPR}(\tau) = \frac{\text{TP}(\tau)}{N^{+}},\qquad
\text{FPR}(\tau) = \frac{\text{FP}(\tau)}{N^{-}}.
\]

Sweeping $\tau$ produces the receiver operating characteristic (ROC) curve, which plots TPR versus FPR.
The ROC area under the curve (ROC AUC) is
\[
\text{ROC AUC} = \int_{0}^{1} \text{TPR}(f) \, \mathrm{d}f,
\]
where $f$ denotes the false positive rate.
Equivalently, ROC AUC is the probability that a randomly chosen active is scored higher than a randomly chosen decoy.
ROC AUC can remain high even when early enrichment is modest, so we treat it as a secondary, legacy metric and report it in the appendix.

\subsection{BEDROC}

Boltzmann enhanced discrimination of ROC (BEDROC) is a rank based metric that places an exponential weight on early retrieved actives.
Let $N^{+}$ be the number of actives, and let $r_j \in \{1, \dots, N\}$ denote the rank of the $j$th active in the sorted list (lower rank indicates higher score).
For a given parameter $\alpha > 0$, define
\[
R = \frac{1}{N^{+}} \sum_{j=1}^{N^{+}} \exp\!\left(-\alpha \, \frac{r_j}{N}\right).
\]
BEDROC rescales $R$ so that a random ranking yields a value of $0$ and a perfect ranking yields $1$:
\[
\text{BEDROC}(\alpha) =
\frac{R - R_{\text{rand}}}{R_{\text{max}} - R_{\text{rand}}}.
\]
Here $R_{\text{rand}}$ and $R_{\text{max}}$ are constants that depend on $\alpha$, $N$, and $N^{+}$ and correspond to the expected value under random ranking and the value under a perfect ranking respectively.
Larger $\alpha$ places more weight on the very top of the ranked list.
In our experiments we fix a single $\alpha$ value and use BEDROC only as an additional early enrichment summary in the appendix.

\subsection{Aggregation Across Targets and Seeds}

All metrics above are computed separately for each target.
For each method and setting we then aggregate as follows:
\begin{itemize}
  \item For each target we average the metric over three independent training seeds on the fixed data split.
  \item Across the 93 targets we report the mean and standard deviation of the per target values in the main tables.
  \item In the appendix we also report medians and interquartile ranges across targets, and class wise averages grouped by protein class.
\end{itemize}
This per target then across targets aggregation avoids biasing metrics toward targets with larger test sets.

\section{Additional Details for Tasks and Evaluation Settings}
\label{app:tasks-settings}

This section expands the main protocol description in
Section~\ref{sec:tasks-settings} and Table~\ref{tab:protocol_all_settings}
with full per setting details, including split rules and per target counts.

\subsection{Per Target LBVS Task}
\label{sec:per-target-task}

Each of the 93 protein targets defines a separate LBVS task. Given a ligand
(SMILES or molecular graph) and a fixed protein target, the goal is to rank
compounds by the probability of being active. For target $t$, the model
outputs a real valued score $s_t(c)$ for each test compound $c$, which
induces a ranking. Evaluation metrics are computed per target from this
ranking and then aggregated across targets.

The targets span 7 protein classes (GPCRs, kinases, nuclear receptors,
ion channels, proteases, cytochrome P450s, and miscellaneous enzymes),
with per target test libraries ranging from a few hundred to over ten thousand
compounds. Targets classified as ``Miscellaneous'' in ChEMBL were excluded
due to insufficient class coherence for meaningful grouping. TopU libraries
are constructed at a fixed 1 to 40 active to decoy ratio using property
matched, structurally challenging decoys (Section~\ref{sec:topu}).

\subsection{Setting 1: ChEMBL$^*$ to TopU (Main Benchmark)}
\label{sec:setting1-main}

In Setting~1, models are trained on ChEMBL$^*$ and evaluated on the TopU
libraries. This setting is intended to mimic a realistic prospective workflow:
train on historical SAR for a target and then screen a new, challenging
library.

For each target, we start from ChEMBL~35, apply our cleaning pipeline
(Section~\ref{app:chembl_star}), and then remove all TopU compounds for that
target to ensure zero overlap between training and test. The remaining
ChEMBL$^*$ actives and decoys form the training pool. We construct per
target training sets at a fixed 1 to 10 active to decoy ratio. When a
target has too few decoys after removing TopU, we augment its negative
pool with decoys sampled from other targets within the same protein class,
using a fixed random seed and releasing the exact selections with the
benchmark.

From this pool, we reserve 15 \% as a stratified random validation set, preserving the 1:10 ratio, and
use the remaining 85\% for training. The test set for Setting~1 is the full
TopU library for each target, constructed at a 1 to 40 active to decoy ratio.
All models use the same fixed train, validation, and test splits.

For Setting~1, the primary test metric is EF@1\%, and PR AUC is reported as
a secondary ranking metric. Model selection is performed by maximizing PR AUC
on the validation set. We aggregate results across the 93 targets by
reporting mean and median per target EF@1\%, with analogous aggregation for
PR AUC.

\subsection{Setting 2: TopU Only (Few Shot)}
\label{sec:setting2-main}

Setting~2 evaluates models when both training and test compounds come from
the same hard, property matched TopU distribution. This setting captures a
few shot regime where only a limited number of high quality TopU actives are
available for training.

For each target, all splits (train, validation, test) are drawn from its
TopU library at the fixed 1 to 40 active to decoy ratio. Because the number
of TopU actives varies widely across targets, we use a tiered split strategy
based on the number of TopU actives. Targets with fewer than 50 TopU actives
(Tier~1) use a 6:2:2 split of actives into train, validation, and test, while
targets with at least 50 TopU actives (Tier~2) use a 7:1:2 split. In both
tiers, decoys are assigned to splits to preserve the 1 to 40 ratio per
target. All splits are scaffold based and use the same fixed seed as in
Setting~1.

Because the smallest TopU libraries make EF@1\% numerically unstable (only a
few molecules in the top 1\%), we use EF@10\% as the primary test metric for
Setting~2 and PR AUC as the secondary metric. We report mean and median
EF@10\% separately for Tier~1 and Tier~2, as well as overall. Detailed split
rules and per target ranges of train, validation, and test actives are
summarized in Table~\ref{tab:setting2_tiers}.

\begin{table}[h]
\centering
\caption{Tier summary for Setting~2. Active counts are shown as 
min to max ranges across targets within each tier.}
\label{tab:setting2_tiers}
\begin{tabular}{lcccccc}
\toprule
\textbf{Tier} & \textbf{Cutoff} & \textbf{Targets} & \textbf{Split} & 
\textbf{Train actives} & \textbf{Val actives} & \textbf{Test actives} \\
\midrule
Tier~1 & $< 50$    & 46 & 6:2:2 & 6 to 29   & 2 to 10  & 2 to 10  \\
Tier~2 & $\geq 50$ & 47 & 7:1:2 & 37 to 246 & 5 to 35  & 11 to 71 \\
\bottomrule
\end{tabular}
\end{table}

\subsection{Setting 3: ChEMBL$^*$ Only (Ablation)}
\label{sec:setting3-main}

Setting~3 is an ablation that uses only ChEMBL$^*$ for both training and
test, with randomly sampled negatives and no TopU decoy curation. The goal
is to contrast performance on standard ChEMBL style splits with the harder
TopU test libraries in Setting~1.

To keep the comparison focused, Setting~3 is defined on seven representative
targets, one from each protein class, chosen as the targets with the largest
number of TopU actives in that class. The selected targets are listed in
Table~\ref{tab:setting3_targets}.

\begin{table}[h]
\centering
\caption{Selected targets for Setting~3, one per protein class.}
\label{tab:setting3_targets}
\resizebox{\textwidth}{!}{%
\begin{tabular}{llccc}
\toprule
\textbf{Class} & \textbf{Target} & \textbf{Full Name} & 
\textbf{TopU actives} & \textbf{ChEMBL$^*$ actives} \\
\midrule
Cytochrome P450  & cp3a4 & Cytochrome P450 3A4             & 154 & 3,686 \\
GPCR             & aa2ar & Adenosine A2a receptor           & 127 & 5,159 \\
Ion Channel      & kcnh2 & hERG                             & 352 & 7,067 \\
Kinase           & egfr  & Epidermal growth factor receptor & 153 & 7,509 \\
Nuclear Receptor & esr1  & Estrogen receptor $\alpha$       & 113 & 3,231 \\
Misc.\ Enzymes   & aces  & Acetylcholinesterase             & 118 & 4,665 \\
Protease         & thrb  & Thrombin                         & 104 & 4,487 \\
\bottomrule
\end{tabular}}
\end{table}

To enable a direct, controlled comparison with Setting~1, the training and
validation sets in Setting~3 are constructed to be identical in size to those
used in Setting~1 for the same targets. Specifically, for each selected
target, the TopU compounds are first excluded from ChEMBL$^*$ to prevent any
overlap between the training pool and the Setting~1 test set. The remaining
ChEMBL$^*$ compounds are then split into training and validation sets using the same 85\%/15\% stratified random split (seed 2026) and 1 to 10 active to decoy ratio as Setting~1.

For each target, we pool all ChEMBL$^*$ inactives (including same-class augmentation when used in Setting~1) together with all TopU inactives into a single inactive pool, and shuffle this pool deterministically with the per-target seed. The shuffled pool is then sliced sequentially into train (1:10 active-to-decoy), validation (1:10), and test (1:40) inactive sets, preserving the active counts from the corresponding Setting~1 splits. This design uses identical actives across Settings~1 and~3, so the only variable across the two test sets is the construction of the test inactives: a focused TopU hard-decoy library in Setting~1, versus a shuffled mixture of ChEMBL$^*$ background inactives and TopU compounds in Setting~3. Across the seven mini targets, TopU compounds make up a small minority of the Setting~3 test inactives (typically under 10\%), so the Setting~1 vs.\ Setting~3 comparison reported in Table~\ref{tab:mini_topu_vs_random}  is conservative. The true gap between hard-decoy and random-decoy evaluation is therefore at least as large as Table~\ref{tab:mini_topu_vs_random} shows.

Table~\ref{tab:setting3_counts} reports the resulting dataset sizes for each
target.

\begin{table}[h]
\centering
\caption{\textbf{Dataset sizes for Setting~3.} All counts are number of active 
compounds. Decoys follow a 1:10 active-to-decoy ratio in the train and validation 
splits and a 1:40 ratio in the test split.}
\label{tab:setting3_counts}
\begin{tabular}{lcccc}
\toprule
\textbf{Target} & \textbf{Class} & \textbf{Train actives} & 
\textbf{Val actives} & \textbf{Test actives} \\
\midrule
 cp3a4 & Cytochrome P450   & 3,002 & 530   & 154 \\
aa2ar & GPCR             & 4,277 & 755   & 127 \\
kcnh2 & Ion Channel      & 5,707 & 1,008 & 352 \\
egfr  & Kinase           & 6,252 & 1,104 & 153 \\
esr1  & Nuclear Receptor & 2,646 & 468   & 112 \\
aces  & Misc.\ Enzymes   & 3,789 & 669   & 117 \\
thrb  & Protease         & 3,680 & 650   & 104 \\
\bottomrule
\end{tabular}
\end{table}

As in Settings~1 and~2, the validation set uses PR AUC (Average Precision)
as the model selection metric. The primary test metric for Setting~3 is
EF@1\%, identical to Setting~1, enabling direct comparison between the two
settings. We also report PR AUC on the ChEMBL$^*$ test sets as a secondary
metric, using the same aggregation protocol as in Setting~1. Performance is
reported per target as well as mean and median EF@1\% across the seven
selected targets.

\section{Limitations and Scope}
\label{app:limitations}

TopU-LBVS is designed to stress-test ligand-based virtual screening methods under hard-negative conditions, and its scope reflects this focus. First, the benchmark is restricted to 2D ligand-based representations and does not incorporate protein structural information, docking, or 3D conformational effects. As a result, it does not evaluate structure-based screening methods or hybrid ligand–protein models. Second, TopU decoys are intentionally\textbf{\textbf{ }}adversarial, emphasizing worst-case scenarios in which inactive compounds are highly similar to known actives. While this design exposes failure modes that standard benchmarks obscure, it may overestimate difficulty relative to early exploratory screening tasks with more diverse libraries. Third, bioactivity labels are derived from curated public data and binarized using fixed thresholds. Although this reflects common practice and improves consistency, it does not capture experimental uncertainty, assay context, or graded potency information. Finally, the benchmark evaluates a fixed panel of targets drawn from ChEMBL 35. While diverse across major protein classes, conclusions may not directly generalize to underrepresented target families or proprietary chemical spaces. These limitations are intentional trade-offs that prioritize controlled evaluation of robustness and generalization over broad task coverage. TopU-LBVS is intended to complement, rather than replace, existing LBVS benchmarks. 

\section{Broader impacts} \label{app:broader}
TopU-LBVS is intended to improve the reliability and reproducibility of ligand-based virtual screening research by providing fixed hard-negative protocols, standardized metrics, reference baselines, and released evaluation code. By exposing the gap between random-decoy performance and hard-negative screening performance, the benchmark may help reduce overoptimistic claims about molecular machine learning models and encourage methods that are more robust under realistic early-stage discovery conditions. The benchmark could support more efficient prioritization of candidate compounds, potentially reducing wasted computational and experimental effort. At the same time, TopU-LBVS is a retrospective evaluation resource and should not be interpreted as evidence that any model is ready for clinical or experimental deployment. As with other molecular screening tools, improved virtual screening methods could be misused to prioritize harmful bioactive compounds; however, TopU-LBVS does not provide generative design capabilities, synthesis routes, or prospective validation, and its release is focused on transparent evaluation rather than molecule generation or deployment. We therefore view the primary impact as improving scientific rigor and reproducibility in molecular machine learning benchmarks.

\section{Additional Results}
\label{sec:additional_metrics}

This appendix provides complete per-target numerical results for all three
TopU-LBVS protocols. It is organised in three parts. Part~I contains the
primary-metric per-target tables --- one table per protocol --- which are the
canonical reference for comparing new methods against our baselines.
Part~II contains full per-model, per-metric tables for TopU-LBVS-full,
reporting all seven metrics (EF@1\%, EF@5\%, EF@10\%, PR-AUC, ROC-AUC,
BEDROC, LogAUC) for each of the ten baselines individually.
Part~III contains the analogous full per-model, per-metric tables for
TopU-LBVS-low.

A quick-reference index is given in Table~\ref{tab:appendix_index}.

\begin{table}[ht]
\centering
\caption{\textbf{Appendix table index.} All per-target result tables and
  their locations.}
\label{tab:appendix_index}
\begin{tabular}{p{7.5cm} l l}
\toprule
\textbf{Content} & \textbf{Table(s)} & \textbf{Section} \\
\midrule
\multicolumn{3}{l}{\textit{Part I — Primary metrics, all models, all targets}} \\
\midrule
TopU-LBVS-full: EF@1\%, all 10 models $\times$ 93 targets
  & \ref{tab:full_morganrf_1}--\ref{tab:full_molformer_2} & \ref{sec:partI} \\
TopU-LBVS-low: EF@10\%, all 10 models $\times$ 93 targets
  & \ref{tab:few_morganrf_1}--\ref{tab:few_molformer_2} & \ref{sec:partI} \\
TopU-LBVS-mini: EF@1\%, TopU hard decoys (T), all 10 models $\times$ 7 targets
  & \ref{tab:mini_hard_full} & \ref{sec:partI} \\
TopU-LBVS-mini: EF@1\%, random decoys (R), all 10 models $\times$ 7 targets
  & \ref{tab:mini_random_full} & \ref{sec:partI} \\
\midrule
\multicolumn{3}{l}{\textit{Part II — Full metrics per model, TopU-LBVS-full}} \\
\midrule
Morgan-RF  & \ref{tab:full_morganrf_1}--\ref{tab:full_morganrf_2}   & \ref{sec:partII} \\
Tanimoto-NN & \ref{tab:full_tanimoto_1}--\ref{tab:full_tanimoto_2} & \ref{sec:partII} \\
GAT        & \ref{tab:full_gat_1}--\ref{tab:full_gat_2}             & \ref{sec:partII} \\
GAT+FP     & \ref{tab:full_gatfp_1}--\ref{tab:full_gatfp_2}         & \ref{sec:partII} \\
GIN        & \ref{tab:full_gin_1}--\ref{tab:full_gin_2}             & \ref{sec:partII} \\
GIN+FP     & \ref{tab:full_ginfp_1}--\ref{tab:full_ginfp_2}         & \ref{sec:partII} \\
GPS        & \ref{tab:full_gps_1}--\ref{tab:full_gps_2}             & \ref{sec:partII} \\
GPS+FP     & \ref{tab:full_gpsfp_1}--\ref{tab:full_gpsfp_2}         & \ref{sec:partII} \\
D-MPNN     & \ref{tab:full_dmpnn_1}--\ref{tab:full_dmpnn_2}         & \ref{sec:partII} \\
MolFormer  & \ref{tab:full_molformer_1}--\ref{tab:full_molformer_2} & \ref{sec:partII} \\
\midrule
\multicolumn{3}{l}{\textit{Part III — Full metrics per model, TopU-LBVS-low}} \\
\midrule
Morgan-RF  & \ref{tab:few_morganrf_1}--\ref{tab:few_morganrf_2}     & \ref{sec:partIII} \\
Tanimoto-NN & \ref{tab:few_tanimoto_1}--\ref{tab:few_tanimoto_2}   & \ref{sec:partIII} \\
GAT        & \ref{tab:few_gat_1}--\ref{tab:few_gat_2}               & \ref{sec:partIII} \\
GAT+FP     & \ref{tab:few_gatfp_1}--\ref{tab:few_gatfp_2}           & \ref{sec:partIII} \\
GIN        & \ref{tab:few_gin_1}--\ref{tab:few_gin_2}               & \ref{sec:partIII} \\
GIN+FP     & \ref{tab:few_ginfp_1}--\ref{tab:few_ginfp_2}           & \ref{sec:partIII} \\
GPS        & \ref{tab:few_gps_1}--\ref{tab:few_gps_2}               & \ref{sec:partIII} \\
GPS+FP     & \ref{tab:few_gpsfp_1}--\ref{tab:few_gpsfp_2}           & \ref{sec:partIII} \\
D-MPNN     & \ref{tab:few_dmpnn_1}--\ref{tab:few_dmpnn_2}           & \ref{sec:partIII} \\
MolFormer  & \ref{tab:few_molformer_1}--\ref{tab:few_molformer_2}   & \ref{sec:partIII} \\
\bottomrule
\end{tabular}
\end{table}

\subsection{Part I: Primary-Metric Per-Target Results}
\label{sec:partI}

Tables~\ref{tab:full_ef1_all1} and~\ref{tab:full_ef1_all2} report per-target EF@1\% for
TopU-LBVS-full across all 93 targets and all 10 models.
Tables~\ref{tab:few_ef10_all1} and~\ref{tab:few_ef10_all2} report per-target EF@10\% for
TopU-LBVS-few, grouped by protein class with tier indicated.
Tables~\ref{tab:mini_hard_full} and~\ref{tab:mini_random_full} report
per-target EF@1\% for TopU-LBVS-mini under TopU hard decoys (T) and
randomly sampled ChEMBL$^*$ decoys (R), respectively; the gap between
the two tables quantifies decoy difficulty per target and per model.

\begin{table}
\centering
\caption{\textbf{TopU-LBVS-Full EF1\% results - part 1.}  Per-target EF@1\% comparison across baseline models. }
\label{tab:full_ef1_all1}
\setlength{\tabcolsep}{3pt}
\footnotesize
\resizebox{.9\textwidth}{!}{%
\begin{tabular}{l cccccccccc}
\toprule
\textbf{Target} & \textbf{Mor-RF} & \textbf{Tan-NN} & \textbf{GAT} & \textbf{GAT+FP} & \textbf{GIN} & \textbf{GIN-FP} & \textbf{GPS} & \textbf{GPS-FP} & \textbf{DMPNN} & \textbf{MolF} \\
\midrule
\multicolumn{11}{c}{\textbf{Cytochrome P450}} \\
\midrule
cp1a2 & 12.4 & 12.2 & 5.1 & 12.2 & 4.9 & 13.2 & 3.2 & 13.2 & 11.7 & 7.8 \\
cp2c9 & 12.6 & 6.4 & 1.4 & 13.4 & 5.0 & 14.2 & 3.1 & 15.1 & 11.4 & 11.2 \\
cp2d6 & 15.8 & 9.8 & 8.6 & 15.6 & 7.7 & 15.8 & 8.9 & 15.1 & 16.1 & 11.3 \\
cp3a4 & 15.9 & 10.9 & 9.0 & 14.3 & 10.2 & 15.6 & 5.8 & 16.4 & 14.9 & 14.5 \\
\midrule
\multicolumn{11}{c}{\textbf{GPCR}} \\
\midrule
5ht6r & 2.8 & 6.2 & 4.1 & 6.2 & 3.4 & 3.4 & 0.7 & 5.5 & 4.8 & 4.1 \\
\bottomrule
\end{tabular}%
}
\end{table}

\begin{table}
\centering
\caption{\textbf{TopU-LBVS-Full EF1\% results - part 2.} Per-target EF@1\% comparison across baseline models.}
\label{tab:full_ef1_all2}
\setlength{\tabcolsep}{3pt}
\footnotesize
\resizebox{.9\textwidth}{!}{%
\begin{tabular}{l cccccccccc}
\toprule
\textbf{Target} & \textbf{Mor-RF} & \textbf{Tan-NN} & \textbf{GAT} & \textbf{GAT+FP} & \textbf{GIN} & \textbf{GIN-FP} & \textbf{GPS} & \textbf{GPS-FP} & \textbf{DMPNN} & \textbf{MolF} \\
\midrule
\multicolumn{11}{c}{\textbf{Nuclear Receptor}} \\
\midrule
andr & 12.2 & 12.7 & 1.5 & 4.9 & 0.5 & 4.4 & 2.4 & 3.9 & 4.4 & 4.9 \\
esr1 & 6.2 & 8.0 & 2.4 & 4.2 & 2.1 & 3.3 & 2.4 & 3.3 & 5.9 & 2.7 \\
esr2 & 8.0 & 8.3 & 2.4 & 4.8 & 2.4 & 3.6 & 2.4 & 5.9 & 4.2 & 2.4 \\
gcr & 10.1 & 7.1 & 5.9 & 8.9 & 5.3 & 10.7 & 3.0 & 5.9 & 8.3 & 8.3 \\
mcr & 7.0 & 6.2 & 0.0 & 8.2 & 0.0 & 0.0 & 2.7 & 2.7 & 2.7 & 8.2 \\
ppara & 6.5 & 4.9 & 0.0 & 0.0 & 3.2 & 0.0 & 0.8 & 1.6 & 0.8 & 4.8 \\
\bottomrule
\end{tabular}%
}
\end{table}

\begin{table}[ht]
\centering
\caption{\textbf{TopU-LBVS-few: EF@10\% results - part 1.}
  Rows are individual targets grouped by protein class; the Tier column indicates
  the split strategy (1: fewer than 50 TopU actives, 6:2:2 split;
  2: at least 50 TopU actives, 7:1:2 split).
  Higher is better.
}
\label{tab:few_ef10_all1}
\resizebox{\textwidth}{!}{%
\begin{tabular}{l c rrrrrrrrrr}
\toprule
\textbf{Target} & \textbf{Tier}
  & \textbf{Morgan-RF} & \textbf{Tanimoto-NN}
  & \textbf{GIN} & \textbf{GIN+FP}
  & \textbf{GAT} & \textbf{GAT+FP}
  & \textbf{GPS} & \textbf{GPS+FP}
  & \textbf{DMPNN} & \textbf{MolFormer} \\
\midrule
\multicolumn{12}{l}{\textit{Cytochrome P450}} \\
\midrule
cp1a2 & 2 & 1.5 & 1.2 & 0.9 & 1.1 & 0.7 & 1.1 & 1.6 & 0.5 & 1.6 & 2.2 \\
cp2c9 & 2 & 0.7 & 0.6 & 1.5 & 0.8 & 1.7 & 0.7 & 1.1 & 0.7 & 1.1 & 1.2 \\
cp2d6 & 2 & 1.5 & 1.7 & 1.0 & 1.0 & 1.2 & 1.1 & 1.7 & 0.8 & 2.3 & 2.0 \\
cp3a4 & 2 & 1.2 & 1.1 & 1.3 & 0.7 & 1.0 & 1.0 & 1.2 & 0.6 & 1.6 & 1.4 \\
\midrule
\multicolumn{12}{l}{\textit{GPCR}} \\
\midrule
5ht6r  & 1 & 0.3 & 1.3 & 2.7 & 1.7 & 1.3 & 0.7 & 1.0 & 1.3 & 2.7 & 0.7 \\
\bottomrule
\end{tabular}%
}
\end{table}

\begin{table}[ht]
\centering
\caption{\textbf{TopU-LBVS-few: EF@10\% results - part 2.}
  Rows are individual targets grouped by protein class; the Tier column indicates
  the split strategy (1: fewer than 50 TopU actives, 6:2:2 split;
  2: at least 50 TopU actives, 7:1:2 split).
  Higher is better.
}
\label{tab:few_ef10_all2}
\resizebox{\textwidth}{!}{%
\begin{tabular}{l c rrrrrrrrrr}
\toprule
\textbf{Target} & \textbf{Tier}
  & \textbf{Morgan-RF} & \textbf{Tanimoto-NN}
  & \textbf{GIN} & \textbf{GIN+FP}
  & \textbf{GAT} & \textbf{GAT+FP}
  & \textbf{GPS} & \textbf{GPS+FP}
  & \textbf{DMPNN} & \textbf{MolFormer} \\
\midrule
\multicolumn{12}{l}{\textit{Nuclear Receptor}} \\
\midrule
andr  & 2 & 0.7 & 0.0 & 0.8 & 0.8 & 0.8 & 0.8 & 0.5 & 0.8 & 0.5 & 1.0 \\
esr1  & 2 & 0.8 & 0.1 & 1.6 & 1.2 & 1.0 & 0.3 & 0.9 & 0.0 & 1.7 & 1.2 \\
esr2  & 2 & 0.7 & 0.3 & 1.2 & 0.6 & 1.5 & 0.0 & 1.2 & 0.0 & 0.6 & 0.3 \\
gcr   & 2 & 0.1 & 0.3 & 2.7 & 0.9 & 0.9 & 0.9 & 1.2 & 0.6 & 3.3 & 1.5 \\
mcr   & 1 & 0.0 & 0.0 & 0.0 & 0.0 & 1.5 & 0.0 & 0.0 & 3.0 & 0.0 & 1.5 \\
ppara & 1 & 0.2 & 0.0 & 0.4 & 0.4 & 2.1 & 0.4 & 1.7 & 1.7 & 0.0 & 2.1 \\
\bottomrule
\end{tabular}%
}
\end{table}
\begin{table}[ht]
\centering
\caption{\textbf{TopU-LBVS-mini: full metric results on TopU hard-decoy test sets.}
  Each block reports one metric; rows are the 7 mini targets, columns are models.
  PR-AUC, ROC-AUC, BEDROC, and LogAUC are on a 0--100 scale.
  Higher is better for all metrics.
}
\label{tab:mini_hard_full}
\resizebox{\textwidth}{!}{%
\begin{tabular}{l cccccccccc}
\toprule
\textbf{Target}
  & \textbf{Morgan-RF} & \textbf{Tanimoto-NN}
  & \textbf{GAT} & \textbf{GAT+FP}
  & \textbf{GIN} & \textbf{GIN+FP}
  & \textbf{GPS} & \textbf{GPS+FP}
  & \textbf{DMPNN} & \textbf{MolFormer} \\
\midrule
\multicolumn{11}{l}{\textbf{EF@1\%}} \\
\midrule
aa2ar         &  6.2 &  3.6 &  4.1 &  5.2 &  2.8 &  6.2 &  1.0 &  5.7 &  3.1 &  2.8 \\
aces          & 10.0 &  8.0 &  2.8 &  4.8 &  2.6 &  5.1 &  3.4 &  6.8 &  6.5 &  3.4 \\
cp2d6         & 12.6 &  9.8 &  8.6 & 15.6 &  7.7 & 15.8 &  8.9 & 15.1 & 16.1 & 11.3 \\
egfr          &  6.4 &  5.6 &  2.0 &  2.0 &  2.8 &  3.9 &  2.6 &  3.9 &  7.4 &  3.5 \\
esr1          &  6.2 &  8.0 &  2.4 &  4.2 &  2.1 &  3.3 &  2.4 &  3.3 &  5.9 &  2.7 \\
kcnh2         &  5.4 &  4.8 &  2.0 &  2.5 &  1.5 &  2.4 &  2.5 &  1.9 &  9.1 &  3.8 \\
\bottomrule
\end{tabular}%
}
\end{table}
\begin{table}[ht]
\centering
\caption{\textbf{TopU-LBVS-mini: full metric results on random-decoy test sets.}
  Each block reports one metric; rows are the 7 mini targets, columns are models.
  PR-AUC, ROC-AUC, BEDROC, and LogAUC are on a 0--100 scale.
  Higher is better for all metrics.}
\label{tab:mini_random_full}
\resizebox{\textwidth}{!}{%
\begin{tabular}{l cccccccccc}
\toprule
\textbf{Target}
  & \textbf{Morgan-RF} & \textbf{Tanimoto-NN}
  & \textbf{GAT} & \textbf{GAT+FP}
  & \textbf{GIN} & \textbf{GIN+FP}
  & \textbf{GPS} & \textbf{GPS+FP}
  & \textbf{DMPNN} & \textbf{MolFormer} \\
\midrule
\multicolumn{11}{c}{\textbf{EF@1\%}} \\
\midrule
aa2ar         & 27.2 & 18.6 & 13.4 & 29.7 & 18.3 & 29.9 &  9.3 & 30.2 & 28.1 & 25.0 \\
aces          & 28.8 & 18.7 & 16.5 & 32.2 & 16.2 & 30.2 & 13.1 & 30.5 & 30.5 & 27.3 \\
cp2d6         & 15.5 & 10.1 &  8.2 & 14.6 &  7.2 & 15.1 &  5.5 & 15.1 & 14.1 & 10.3 \\
egfr          & 32.1 & 17.4 & 16.9 & 31.7 & 22.6 & 32.8 & 15.8 & 30.8 & 28.0 & 28.4 \\
esr1          & 24.4 & 17.5 & 15.4 & 23.5 & 18.1 & 25.0 & 10.1 & 22.3 & 24.7 & 19.6 \\
kcnh2         & 19.2 & 12.0 & 10.2 & 20.5 & 10.2 & 20.7 &  9.4 & 21.1 & 20.1 & 19.1 \\
\bottomrule
\end{tabular}%
}
\end{table}

\subsection{Part II: Full Per-Model Metrics, TopU-LBVS-full}
\label{sec:partII}

Each pair of tables below reports all seven evaluation metrics
(EF@1\%, EF@5\%, EF@10\%, PR-AUC, ROC-AUC, BEDROC, LogAUC)
for a single baseline model across all 93 targets under the
ChEMBL$^*$$\to$TopU protocol. Results are split across two tables
per model due to the number of targets; the split follows the same
class ordering as the main paper (Cytochrome P450 through Protease).

\begin{table}[h!]
\centering
\caption{\textbf{TopU-LBVS-Full Morgan-RF results.}
All values are reported as mean\,$\pm$\,std over three random seeds.
EF\,=\,Enrichment Factor. PR-AUC, ROC-AUC, BEDROC ($\alpha$\,=\,20), and LogAUC are reported on a 0--100 scale.
}
\label{tab:full_morganrf_1}
\setlength{\tabcolsep}{4pt}
\renewcommand{\arraystretch}{1.08}
\footnotesize
\begin{tabular}{l ccc||cccc}
\toprule
\textbf{Target} 
  & \textbf{EF@1\%} & \textbf{EF@5\%} & \textbf{EF@10\%}
  & \textbf{PR-AUC} & \textbf{ROC-AUC} & \textbf{BEDROC} & \textbf{LogAUC} \\
\midrule

\multicolumn{8}{c}{\textbf{Cytochrome P450}} \\[1pt]
\midrule
cp1a2   & 12.7{\scriptsize$\pm$1.2} & 8.0{\scriptsize$\pm$0.3} & 5.3{\scriptsize$\pm$0.1} & 17.5{\scriptsize$\pm$0.3} & 81.4{\scriptsize$\pm$0.4} & 36.7{\scriptsize$\pm$1.0} & 40.2{\scriptsize$\pm$0.3} \\
cp2c9   & 12.0{\scriptsize$\pm$0.4} & 7.6{\scriptsize$\pm$0.3} & 5.4{\scriptsize$\pm$0.1} & 16.6{\scriptsize$\pm$0.7} & 84.3{\scriptsize$\pm$0.2} & 36.4{\scriptsize$\pm$0.9} & 41.0{\scriptsize$\pm$0.4} \\
cp2d6   & 16.3{\scriptsize$\pm$0.3} & 9.5{\scriptsize$\pm$0.2} & 6.2{\scriptsize$\pm$0.2} & 23.4{\scriptsize$\pm$0.9} & 84.6{\scriptsize$\pm$0.2} & 44.3{\scriptsize$\pm$0.4} & 45.2{\scriptsize$\pm$0.4} \\
cp3a4   & 15.8{\scriptsize$\pm$0.8} & 9.5{\scriptsize$\pm$0.1} & 6.3{\scriptsize$\pm$0.2} & 24.5{\scriptsize$\pm$0.3} & 85.6{\scriptsize$\pm$0.4} & 45.4{\scriptsize$\pm$0.4} & 46.1{\scriptsize$\pm$0.4} \\
\midrule

\multicolumn{8}{c}{\textbf{GPCR}} \\[1pt]
\midrule
aa2ar  & 5.7{\scriptsize$\pm$0.4}  & 4.0{\scriptsize$\pm$0.1} & 3.5{\scriptsize$\pm$0.1} & 8.0{\scriptsize$\pm$0.1} & 75.2{\scriptsize$\pm$0.3} & 19.0{\scriptsize$\pm$0.2} & 30.3{\scriptsize$\pm$0.4} \\
\bottomrule
\end{tabular}
\end{table}
\begin{table}[htbp]
\centering
\caption{\textbf{TopU-LBVS-Full Morgan-RF results.}
All values are reported as mean\,$\pm$\,std over three random seeds.
EF\,=\,Enrichment Factor. PR-AUC, ROC-AUC, BEDROC ($\alpha$\,=\,20), and LogAUC are reported on a 0--100 scale.
}
\label{tab:full_morganrf_2}
\setlength{\tabcolsep}{4pt}
\renewcommand{\arraystretch}{1.08}
\footnotesize
\begin{tabular}{l ccc||cccc}
\toprule
\textbf{Target} 
  & \textbf{EF@1\%} & \textbf{EF@5\%} & \textbf{EF@10\%}
  & \textbf{PR-AUC} & \textbf{ROC-AUC} & \textbf{BEDROC} & \textbf{LogAUC} \\
\midrule

\multicolumn{8}{c}{\textbf{Nuclear Receptor}} \\[1pt]
\midrule
andr    & 11.7{\scriptsize$\pm$0.0} & 5.6{\scriptsize$\pm$0.4} & 4.4{\scriptsize$\pm$0.1} & 13.7{\scriptsize$\pm$1.0} & 82.7{\scriptsize$\pm$0.8} & 28.0{\scriptsize$\pm$1.9} & 37.4{\scriptsize$\pm$1.1} \\
esr1    & 5.4{\scriptsize$\pm$0.0}  & 5.4{\scriptsize$\pm$0.2} & 4.0{\scriptsize$\pm$0.1} & 9.3{\scriptsize$\pm$0.3} & 72.4{\scriptsize$\pm$0.3} & 22.5{\scriptsize$\pm$0.6} & 30.7{\scriptsize$\pm$0.3} \\
esr2    & 7.1{\scriptsize$\pm$0.0}  & 5.0{\scriptsize$\pm$0.6} & 4.2{\scriptsize$\pm$0.2} & 13.3{\scriptsize$\pm$1.0} & 75.4{\scriptsize$\pm$0.5} & 23.7{\scriptsize$\pm$2.1} & 32.8{\scriptsize$\pm$1.0} \\
gcr     & 8.3{\scriptsize$\pm$0.8}  & 8.2{\scriptsize$\pm$0.5} & 6.1{\scriptsize$\pm$0.2} & 15.0{\scriptsize$\pm$0.2} & 84.8{\scriptsize$\pm$0.2} & 36.1{\scriptsize$\pm$0.7} & 40.1{\scriptsize$\pm$0.1} \\
mcr     & 8.2{\scriptsize$\pm$0.0}  & 3.2{\scriptsize$\pm$1.8} & 3.0{\scriptsize$\pm$0.8} & 12.6{\scriptsize$\pm$4.2} & 76.7{\scriptsize$\pm$1.5} & 17.4{\scriptsize$\pm$5.6} & 29.8{\scriptsize$\pm$2.4} \\
ppara   & 7.2{\scriptsize$\pm$2.0}  & 5.8{\scriptsize$\pm$0.2} & 3.2{\scriptsize$\pm$0.2} & 9.5{\scriptsize$\pm$0.9} & 78.4{\scriptsize$\pm$0.7} & 21.9{\scriptsize$\pm$1.9} & 32.1{\scriptsize$\pm$1.1} \\
\bottomrule
\end{tabular}
\end{table}


\begin{table}[htbp]
\centering
\caption{\textbf{TopU-LBVS-Full Tanimoto-NN results.}
  All values are reported as mean\,$\pm$\,std over three random seeds.
  EF\,=\,Enrichment Factor. PR-AUC, ROC-AUC, BEDROC ($\alpha$\,=\,20), and LogAUC are reported on a 0--100 scale.
}
\label{tab:full_tanimoto_1}
\setlength{\tabcolsep}{4pt}
\renewcommand{\arraystretch}{1.08}
\footnotesize
\begin{tabular}{l ccc||cccc}
\toprule
\textbf{Target}
  & \textbf{EF@1\%} & \textbf{EF@5\%} & \textbf{EF@10\%}
  & \textbf{PR-AUC} & \textbf{ROC-AUC} & \textbf{BEDROC} & \textbf{LogAUC} \\
\midrule

\multicolumn{8}{c}{\textbf{Cytochrome P450}} \\[1pt]
\midrule
cp1a2  & 12.2{\scriptsize$\pm$1.5} & 8.5{\scriptsize$\pm$0.3} & 5.1{\scriptsize$\pm$0.1} & 15.6{\scriptsize$\pm$1.1} & 75.4{\scriptsize$\pm$1.0} & 37.0{\scriptsize$\pm$1.4} & 37.8{\scriptsize$\pm$1.1} \\
cp2c9  & 6.4{\scriptsize$\pm$1.0}  & 7.7{\scriptsize$\pm$0.3} & 6.1{\scriptsize$\pm$0.1} & 14.1{\scriptsize$\pm$0.5} & 82.8{\scriptsize$\pm$0.6} & 35.4{\scriptsize$\pm$1.0} & 39.0{\scriptsize$\pm$0.5} \\
cp2d6  & 9.8{\scriptsize$\pm$0.3}  & 8.9{\scriptsize$\pm$0.1} & 6.4{\scriptsize$\pm$0.1} & 16.2{\scriptsize$\pm$0.2} & 81.9{\scriptsize$\pm$0.2} & 39.3{\scriptsize$\pm$0.3} & 40.5{\scriptsize$\pm$0.2} \\
cp3a4  & 10.9{\scriptsize$\pm$0.0} & 7.8{\scriptsize$\pm$0.2} & 5.8{\scriptsize$\pm$0.1} & 15.0{\scriptsize$\pm$0.5} & 82.3{\scriptsize$\pm$0.3} & 36.4{\scriptsize$\pm$0.9} & 39.7{\scriptsize$\pm$0.5} \\
\midrule

\multicolumn{8}{c}{\textbf{GPCR}} \\[1pt]
\midrule
5ht6r  & 6.2{\scriptsize$\pm$0.0}  & 5.6{\scriptsize$\pm$0.4} & 5.0{\scriptsize$\pm$0.3} & 11.8{\scriptsize$\pm$0.5} & 78.0{\scriptsize$\pm$0.4} & 27.5{\scriptsize$\pm$1.1} & 34.8{\scriptsize$\pm$0.6} \\
\bottomrule
\end{tabular}
\end{table}

\begin{table}[htbp]
\centering
\caption{\textbf{TopU-LBVS-Full Tanimoto-NN results.}
  All values are reported as mean\,$\pm$\,std over three random seeds.
  EF\,=\,Enrichment Factor. PR-AUC, ROC-AUC, BEDROC ($\alpha$\,=\,20), and LogAUC are reported on a 0--100 scale.
}
\label{tab:full_tanimoto_2}
\setlength{\tabcolsep}{4pt}
\renewcommand{\arraystretch}{1.08}
\footnotesize
\begin{tabular}{l ccc||cccc}
\toprule
\textbf{Target}
  & \textbf{EF@1\%} & \textbf{EF@5\%} & \textbf{EF@10\%}
  & \textbf{PR-AUC} & \textbf{ROC-AUC} & \textbf{BEDROC} & \textbf{LogAUC} \\
\midrule

\multicolumn{8}{c}{\textbf{Nuclear Receptor}} \\[1pt]
\midrule
andr   & 12.7{\scriptsize$\pm$0.7} & 6.8{\scriptsize$\pm$0.3} & 5.0{\scriptsize$\pm$0.3} & 16.3{\scriptsize$\pm$0.4} & 86.0{\scriptsize$\pm$0.9} & 34.3{\scriptsize$\pm$1.5} & 39.9{\scriptsize$\pm$0.7} \\
esr1   & 8.0{\scriptsize$\pm$0.0}  & 4.8{\scriptsize$\pm$0.2} & 4.7{\scriptsize$\pm$0.1} & 10.2{\scriptsize$\pm$0.1} & 77.3{\scriptsize$\pm$0.5} & 25.1{\scriptsize$\pm$0.2} & 33.0{\scriptsize$\pm$0.3} \\
esr2   & 8.3{\scriptsize$\pm$1.7}  & 4.8{\scriptsize$\pm$0.3} & 4.3{\scriptsize$\pm$0.2} & 10.0{\scriptsize$\pm$1.0} & 76.1{\scriptsize$\pm$0.3} & 23.7{\scriptsize$\pm$2.3} & 31.9{\scriptsize$\pm$0.9} \\
gcr    & 7.1{\scriptsize$\pm$0.0}  & 7.8{\scriptsize$\pm$0.3} & 5.8{\scriptsize$\pm$0.1} & 14.2{\scriptsize$\pm$0.2} & 86.6{\scriptsize$\pm$0.2} & 33.9{\scriptsize$\pm$0.7} & 39.2{\scriptsize$\pm$0.1} \\
mcr    & 5.5{\scriptsize$\pm$3.9}  & 4.6{\scriptsize$\pm$1.8} & 2.3{\scriptsize$\pm$0.9} & 8.2{\scriptsize$\pm$1.8}  & 75.4{\scriptsize$\pm$0.8} & 18.0{\scriptsize$\pm$8.1} & 26.8{\scriptsize$\pm$2.1} \\
ppara  & 3.2{\scriptsize$\pm$3.0}  & 4.3{\scriptsize$\pm$0.2} & 3.8{\scriptsize$\pm$0.2} & 8.2{\scriptsize$\pm$0.5}  & 79.5{\scriptsize$\pm$0.6} & 18.5{\scriptsize$\pm$2.2} & 29.8{\scriptsize$\pm$0.9} \\
\bottomrule
\end{tabular}
\end{table}

\begin{table}[htbp]
\centering
\caption{\textbf{TopU-LBVS-Full GAT results - part 1.}
All values are reported as mean\,$\pm$\,std over three random seeds.
EF\,=\,Enrichment Factor. PR-AUC, ROC-AUC, BEDROC ($\alpha$\,=\,20), and LogAUC are reported on a 0--100 scale.
}
\label{tab:full_gat_1}
\setlength{\tabcolsep}{4pt}
\renewcommand{\arraystretch}{1.08}
\footnotesize
\begin{tabular}{l ccc||cccc}
\toprule
\textbf{Target} 
  & \textbf{EF@1\%} & \textbf{EF@5\%} & \textbf{EF@10\%}
  & \textbf{PR-AUC} & \textbf{ROC-AUC} & \textbf{BEDROC} & \textbf{LogAUC} \\
\midrule
\multicolumn{8}{c}{\textbf{Cytochrome P450}} \\[1pt]
\midrule
cp1a2   & 5.1{\scriptsize$\pm$1.0} & 4.4{\scriptsize$\pm$0.2} & 3.5{\scriptsize$\pm$0.2} & 7.2{\scriptsize$\pm$0.4} & 72.7{\scriptsize$\pm$0.4} & 18.4{\scriptsize$\pm$1.1} & 28.7{\scriptsize$\pm$0.6} \\
cp2c9   & 1.4{\scriptsize$\pm$0.8} & 2.9{\scriptsize$\pm$0.1} & 2.6{\scriptsize$\pm$0.2} & 5.6{\scriptsize$\pm$0.2} & 75.2{\scriptsize$\pm$1.4} & 10.7{\scriptsize$\pm$0.4} & 26.3{\scriptsize$\pm$0.7} \\
cp2d6   & 8.6{\scriptsize$\pm$0.6} & 6.3{\scriptsize$\pm$0.4} & 4.8{\scriptsize$\pm$0.3} & 13.3{\scriptsize$\pm$1.1} & 81.6{\scriptsize$\pm$0.6} & 30.2{\scriptsize$\pm$1.9} & 37.2{\scriptsize$\pm$1.1} \\
cp3a4   & 9.0{\scriptsize$\pm$1.0} & 4.8{\scriptsize$\pm$0.1} & 4.2{\scriptsize$\pm$0.2} & 11.2{\scriptsize$\pm$0.5} & 78.9{\scriptsize$\pm$0.9} & 23.4{\scriptsize$\pm$1.0} & 33.7{\scriptsize$\pm$0.7} \\
\midrule
\multicolumn{8}{c}{\textbf{GPCR}} \\[1pt]
\midrule
5ht6r   & 4.1{\scriptsize$\pm$0.0} & 1.4{\scriptsize$\pm$0.5} & 2.1{\scriptsize$\pm$0.1} & 5.3{\scriptsize$\pm$0.1} & 73.5{\scriptsize$\pm$0.7} & 6.1{\scriptsize$\pm$0.9} & 24.6{\scriptsize$\pm$0.4} \\
\bottomrule
\end{tabular}
\end{table}

\begin{table}[htbp]
\centering
\caption{\textbf{TopU-LBVS-Full GAT results - part 2.}
All values are reported as mean\,$\pm$\,std over three random seeds.
EF\,=\,Enrichment Factor. PR-AUC, ROC-AUC, BEDROC ($\alpha$\,=\,20), and LogAUC are reported on a 0--100 scale.
}
\label{tab:full_gat_2}
\setlength{\tabcolsep}{4pt}
\renewcommand{\arraystretch}{1.08}
\footnotesize
\begin{tabular}{l ccc||cccc}
\toprule
\textbf{Target} 
  & \textbf{EF@1\%} & \textbf{EF@5\%} & \textbf{EF@10\%}
  & \textbf{PR-AUC} & \textbf{ROC-AUC} & \textbf{BEDROC} & \textbf{LogAUC} \\
\midrule
\multicolumn{8}{c}{\textbf{Nuclear Receptor}} \\[1pt]
\midrule
andr    & 1.5{\scriptsize$\pm$0.0} & 2.0{\scriptsize$\pm$0.4} & 2.8{\scriptsize$\pm$0.3} & 6.2{\scriptsize$\pm$0.1} & 76.7{\scriptsize$\pm$0.2} & 8.8{\scriptsize$\pm$0.8} & 26.8{\scriptsize$\pm$0.2} \\
esr1    & 2.4{\scriptsize$\pm$1.5} & 2.7{\scriptsize$\pm$0.4} & 2.9{\scriptsize$\pm$0.2} & 5.9{\scriptsize$\pm$0.4} & 73.1{\scriptsize$\pm$1.2} & 11.5{\scriptsize$\pm$1.7} & 26.3{\scriptsize$\pm$1.1} \\
esr2    & 2.4{\scriptsize$\pm$0.8} & 2.3{\scriptsize$\pm$0.2} & 2.4{\scriptsize$\pm$0.1} & 6.1{\scriptsize$\pm$0.7} & 72.1{\scriptsize$\pm$0.1} & 8.0{\scriptsize$\pm$0.3} & 24.5{\scriptsize$\pm$0.1} \\
gcr     & 5.9{\scriptsize$\pm$2.2} & 5.0{\scriptsize$\pm$0.5} & 4.9{\scriptsize$\pm$0.1} & 11.7{\scriptsize$\pm$0.8} & 85.5{\scriptsize$\pm$0.5} & 26.1{\scriptsize$\pm$1.8} & 36.7{\scriptsize$\pm$0.9} \\
mcr     & 0.0{\scriptsize$\pm$0.0} & 2.0{\scriptsize$\pm$1.6} & 2.7{\scriptsize$\pm$0.9} & 5.9{\scriptsize$\pm$0.2} & 72.3{\scriptsize$\pm$1.9} & 7.7{\scriptsize$\pm$1.4} & 23.8{\scriptsize$\pm$0.7} \\
ppara   & 0.0{\scriptsize$\pm$0.0} & 1.8{\scriptsize$\pm$0.5} & 2.0{\scriptsize$\pm$0.4} & 4.9{\scriptsize$\pm$0.0} & 71.9{\scriptsize$\pm$0.4} & 6.5{\scriptsize$\pm$0.1} & 23.6{\scriptsize$\pm$0.1} \\
\bottomrule
\end{tabular}
\end{table}

\begin{table}[htbp]
\centering
\caption{\textbf{TopU-LBVS-Full GAT+FP hybrid model results - part 1.}
All values are reported as mean\,$\pm$\,std over three random seeds.
EF\,=\,Enrichment Factor. PR-AUC, ROC-AUC, BEDROC ($\alpha$\,=\,20), and LogAUC are reported on a 0--100 scale.
}
\label{tab:full_gatfp_1}
\setlength{\tabcolsep}{4pt}
\renewcommand{\arraystretch}{1.08}
\footnotesize
\begin{tabular}{l ccc||cccc}
\toprule
\textbf{Target} 
  & \textbf{EF@1\%} & \textbf{EF@5\%} & \textbf{EF@10\%}
  & \textbf{PR-AUC} & \textbf{ROC-AUC} & \textbf{BEDROC} & \textbf{LogAUC} \\
\midrule

\multicolumn{8}{c}{\textbf{Cytochrome P450}} \\[1pt]
\midrule
cp1a2   & 12.2{\scriptsize$\pm$1.2} & 7.1{\scriptsize$\pm$0.4} & 4.9{\scriptsize$\pm$0.2} & 16.0{\scriptsize$\pm$1.0} & 77.3{\scriptsize$\pm$1.1} & 33.1{\scriptsize$\pm$1.7} & 37.6{\scriptsize$\pm$1.3} \\
cp2c9   & 13.4{\scriptsize$\pm$1.4} & 7.5{\scriptsize$\pm$0.2} & 5.3{\scriptsize$\pm$0.1} & 17.2{\scriptsize$\pm$1.3} & 82.8{\scriptsize$\pm$0.3} & 36.3{\scriptsize$\pm$1.1} & 40.5{\scriptsize$\pm$0.6} \\
cp2d6   & 15.6{\scriptsize$\pm$0.9} & 8.7{\scriptsize$\pm$0.1} & 5.7{\scriptsize$\pm$0.2} & 21.8{\scriptsize$\pm$0.8} & 81.6{\scriptsize$\pm$1.0} & 41.1{\scriptsize$\pm$0.9} & 42.8{\scriptsize$\pm$0.5} \\
cp3a4   & 14.3{\scriptsize$\pm$2.1} & 8.4{\scriptsize$\pm$0.6} & 5.8{\scriptsize$\pm$0.1} & 21.7{\scriptsize$\pm$2.3} & 81.9{\scriptsize$\pm$0.5} & 40.8{\scriptsize$\pm$2.0} & 42.8{\scriptsize$\pm$1.2} \\
\midrule

\multicolumn{8}{c}{\textbf{GPCR}} \\[1pt]
\midrule
5ht6r   & 6.2{\scriptsize$\pm$0.0} & 3.8{\scriptsize$\pm$0.3} & 3.7{\scriptsize$\pm$0.3} & 9.1{\scriptsize$\pm$0.6} & 76.7{\scriptsize$\pm$1.5} & 17.8{\scriptsize$\pm$2.6} & 30.5{\scriptsize$\pm$1.1} \\
\bottomrule
\end{tabular}
\end{table}

\begin{table}[htbp]
\centering
\caption{\textbf{TopU-LBVS-Full GAT-FP hybrid model results - part 2.}
All values are reported as mean\,$\pm$\,std over three random seeds.
EF\,=\,Enrichment Factor. PR-AUC, ROC-AUC, BEDROC ($\alpha$\,=\,20), and LogAUC are reported on a 0--100 scale.
}
\label{tab:full_gatfp_2}
\setlength{\tabcolsep}{4pt}
\renewcommand{\arraystretch}{1.08}
\footnotesize
\begin{tabular}{l ccc||cccc}
\toprule
\textbf{Target} 
  & \textbf{EF@1\%} & \textbf{EF@5\%} & \textbf{EF@10\%}
  & \textbf{PR-AUC} & \textbf{ROC-AUC} & \textbf{BEDROC} & \textbf{LogAUC} \\
\midrule

\multicolumn{8}{c}{\textbf{Nuclear Receptor}} \\[1pt]
\midrule
andr    & 4.9{\scriptsize$\pm$1.8} & 5.3{\scriptsize$\pm$0.7} & 4.2{\scriptsize$\pm$0.3} & 10.5{\scriptsize$\pm$0.6} & 83.1{\scriptsize$\pm$0.8} & 23.8{\scriptsize$\pm$2.6} & 34.6{\scriptsize$\pm$1.1} \\
esr1    & 4.2{\scriptsize$\pm$1.1} & 4.4{\scriptsize$\pm$0.8} & 3.8{\scriptsize$\pm$0.1} & 7.3{\scriptsize$\pm$0.5} & 72.7{\scriptsize$\pm$0.3} & 18.5{\scriptsize$\pm$1.9} & 28.9{\scriptsize$\pm$0.8} \\
esr2    & 4.8{\scriptsize$\pm$1.7} & 4.0{\scriptsize$\pm$0.6} & 3.6{\scriptsize$\pm$0.5} & 8.9{\scriptsize$\pm$0.8} & 75.2{\scriptsize$\pm$1.4} & 18.2{\scriptsize$\pm$3.3} & 29.9{\scriptsize$\pm$1.7} \\
gcr     & 8.9{\scriptsize$\pm$0.0} & 6.3{\scriptsize$\pm$0.4} & 5.5{\scriptsize$\pm$0.4} & 14.8{\scriptsize$\pm$0.4} & 85.4{\scriptsize$\pm$0.6} & 32.1{\scriptsize$\pm$1.3} & 39.3{\scriptsize$\pm$0.6} \\
mcr     & 8.2{\scriptsize$\pm$0.0} & 3.9{\scriptsize$\pm$0.0} & 3.3{\scriptsize$\pm$0.5} & 11.9{\scriptsize$\pm$2.0} & 78.2{\scriptsize$\pm$1.5} & 19.7{\scriptsize$\pm$2.9} & 31.7{\scriptsize$\pm$1.2} \\
ppara   & 0.0{\scriptsize$\pm$0.0} & 2.0{\scriptsize$\pm$0.4} & 2.9{\scriptsize$\pm$0.1} & 6.4{\scriptsize$\pm$0.3} & 77.8{\scriptsize$\pm$1.3} & 9.8{\scriptsize$\pm$1.2} & 27.4{\scriptsize$\pm$0.9} \\
\bottomrule
\end{tabular}
\end{table}

\begin{table}[htbp]
\centering
\caption{\textbf{TopU-LBVS-Full GIN results - part 1.}
All values are reported as mean\,$\pm$\,std over three random seeds.
EF\,=\,Enrichment Factor. PR-AUC, ROC-AUC, BEDROC ($\alpha$\,=\,20), and LogAUC are reported on a 0--100 scale.
}
\label{tab:full_gin_1}
\setlength{\tabcolsep}{4pt}
\renewcommand{\arraystretch}{1.08}
\footnotesize
\begin{tabular}{l ccc||cccc}
\toprule
\textbf{Target} 
  & \textbf{EF@1\%} & \textbf{EF@5\%} & \textbf{EF@10\%}
  & \textbf{PR-AUC} & \textbf{ROC-AUC} & \textbf{BEDROC} & \textbf{LogAUC} \\
\midrule

\multicolumn{8}{c}{\textbf{Cytochrome P450}} \\[1pt]
\midrule
cp1a2   & 4.9{\scriptsize$\pm$0.7} & 4.7{\scriptsize$\pm$0.4} & 3.7{\scriptsize$\pm$0.2} & 8.0{\scriptsize$\pm$0.1} & 75.8{\scriptsize$\pm$1.1} & 20.2{\scriptsize$\pm$0.6} & 30.5{\scriptsize$\pm$0.4} \\
cp2c9   & 5.0{\scriptsize$\pm$0.0} & 2.9{\scriptsize$\pm$0.3} & 2.4{\scriptsize$\pm$0.2} & 6.3{\scriptsize$\pm$0.3} & 75.5{\scriptsize$\pm$0.7} & 12.3{\scriptsize$\pm$1.1} & 27.3{\scriptsize$\pm$0.5} \\
cp2d6   & 7.7{\scriptsize$\pm$1.8} & 6.5{\scriptsize$\pm$0.6} & 4.9{\scriptsize$\pm$0.2} & 12.5{\scriptsize$\pm$1.2} & 81.9{\scriptsize$\pm$0.9} & 28.4{\scriptsize$\pm$2.7} & 36.4{\scriptsize$\pm$1.5} \\
cp3a4   & 10.2{\scriptsize$\pm$0.9} & 6.2{\scriptsize$\pm$1.0} & 4.7{\scriptsize$\pm$0.4} & 13.5{\scriptsize$\pm$1.7} & 80.6{\scriptsize$\pm$0.8} & 29.3{\scriptsize$\pm$3.5} & 36.8{\scriptsize$\pm$1.7} \\
\midrule

\multicolumn{8}{c}{\textbf{GPCR}} \\[1pt]
\midrule
5ht6r   & 3.4{\scriptsize$\pm$1.0} & 2.1{\scriptsize$\pm$0.3} & 2.4{\scriptsize$\pm$0.3} & 6.3{\scriptsize$\pm$0.1} & 74.4{\scriptsize$\pm$0.8} & 8.7{\scriptsize$\pm$0.4} & 26.1{\scriptsize$\pm$0.4} \\
\bottomrule
\end{tabular}
\end{table}

\begin{table}[htbp]
\centering
\caption{\textbf{TopU-LBVS-Full GIN results - part 2.}
All values are reported as mean\,$\pm$\,std over three random seeds.
EF\,=\,Enrichment Factor. PR-AUC, ROC-AUC, BEDROC ($\alpha$\,=\,20), and LogAUC are reported on a 0--100 scale.
}
\label{tab:full_gin_2}
\setlength{\tabcolsep}{4pt}
\renewcommand{\arraystretch}{1.08}
\footnotesize
\begin{tabular}{l ccc||cccc}
\toprule
\textbf{Target} 
  & \textbf{EF@1\%} & \textbf{EF@5\%} & \textbf{EF@10\%}
  & \textbf{PR-AUC} & \textbf{ROC-AUC} & \textbf{BEDROC} & \textbf{LogAUC} \\
\midrule

\multicolumn{8}{c}{\textbf{Nuclear Receptor}} \\[1pt]
\midrule
andr    & 0.5{\scriptsize$\pm$0.7} & 2.5{\scriptsize$\pm$0.1} & 2.6{\scriptsize$\pm$0.2} & 6.1{\scriptsize$\pm$0.3} & 79.1{\scriptsize$\pm$0.5} & 9.2{\scriptsize$\pm$1.3} & 27.3{\scriptsize$\pm$0.7} \\
esr1    & 2.1{\scriptsize$\pm$0.8} & 2.3{\scriptsize$\pm$0.5} & 2.8{\scriptsize$\pm$0.3} & 5.7{\scriptsize$\pm$0.6} & 72.9{\scriptsize$\pm$1.2} & 9.8{\scriptsize$\pm$1.4} & 25.6{\scriptsize$\pm$0.9} \\
esr2    & 2.4{\scriptsize$\pm$0.8} & 2.1{\scriptsize$\pm$0.0} & 2.1{\scriptsize$\pm$0.1} & 5.8{\scriptsize$\pm$0.6} & 73.0{\scriptsize$\pm$1.0} & 6.9{\scriptsize$\pm$0.7} & 24.6{\scriptsize$\pm$0.9} \\
gcr     & 5.3{\scriptsize$\pm$1.5} & 5.6{\scriptsize$\pm$0.3} & 5.2{\scriptsize$\pm$0.1} & 11.9{\scriptsize$\pm$0.7} & 85.1{\scriptsize$\pm$0.5} & 27.9{\scriptsize$\pm$1.6} & 37.0{\scriptsize$\pm$0.8} \\
mcr     & 0.0{\scriptsize$\pm$0.0} & 2.0{\scriptsize$\pm$1.6} & 2.3{\scriptsize$\pm$0.5} & 6.2{\scriptsize$\pm$0.7} & 72.5{\scriptsize$\pm$0.6} & 7.8{\scriptsize$\pm$3.6} & 24.1{\scriptsize$\pm$1.4} \\
ppara   & 3.2{\scriptsize$\pm$1.1} & 2.8{\scriptsize$\pm$0.5} & 2.7{\scriptsize$\pm$0.2} & 6.0{\scriptsize$\pm$0.4} & 74.2{\scriptsize$\pm$1.1} & 12.0{\scriptsize$\pm$2.0} & 26.2{\scriptsize$\pm$0.8} \\
\bottomrule
\end{tabular}
\end{table}

\begin{table}[htbp]
\centering
\caption{\textbf{TopU-LBVS-Full GIN+FP hybrid model results - part 1.}
All values are reported as mean\,$\pm$\,std over three random seeds.
EF\,=\,Enrichment Factor. PR-AUC, ROC-AUC, BEDROC ($\alpha$\,=\,20), and LogAUC are reported on a 0--100 scale.
}
\label{tab:full_ginfp_1}
\setlength{\tabcolsep}{4pt}
\renewcommand{\arraystretch}{1.08}
\footnotesize
\begin{tabular}{l ccc||cccc}
\toprule
\textbf{Target}
  & \textbf{EF@1\%} & \textbf{EF@5\%} & \textbf{EF@10\%}
  & \textbf{PR-AUC} & \textbf{ROC-AUC} & \textbf{BEDROC} & \textbf{LogAUC} \\
\midrule

\multicolumn{8}{c}{\textbf{Cytochrome P450}} \\[1pt]
\midrule
cp1a2   & 13.2{\scriptsize$\pm$0.6} & 7.4{\scriptsize$\pm$0.4} & 4.9{\scriptsize$\pm$0.4} & 15.7{\scriptsize$\pm$0.2} & 77.6{\scriptsize$\pm$1.0} & 33.8{\scriptsize$\pm$0.9} & 38.0{\scriptsize$\pm$0.5} \\
cp2c9   & 14.2{\scriptsize$\pm$1.4} & 7.6{\scriptsize$\pm$0.3} & 5.3{\scriptsize$\pm$0.1} & 17.1{\scriptsize$\pm$0.6} & 83.1{\scriptsize$\pm$0.4} & 35.8{\scriptsize$\pm$0.5} & 40.6{\scriptsize$\pm$0.5} \\
cp2d6   & 15.8{\scriptsize$\pm$0.0} & 9.2{\scriptsize$\pm$0.5} & 5.8{\scriptsize$\pm$0.3} & 22.2{\scriptsize$\pm$1.9} & 82.4{\scriptsize$\pm$0.4} & 41.8{\scriptsize$\pm$1.6} & 43.5{\scriptsize$\pm$0.8} \\
cp3a4   & 15.6{\scriptsize$\pm$0.8} & 9.0{\scriptsize$\pm$0.3} & 6.2{\scriptsize$\pm$0.2} & 22.8{\scriptsize$\pm$1.6} & 82.6{\scriptsize$\pm$0.8} & 43.2{\scriptsize$\pm$1.8} & 44.0{\scriptsize$\pm$0.9} \\
\midrule

\multicolumn{8}{c}{\textbf{GPCR}} \\[1pt]
\midrule
5ht6r   & 3.4{\scriptsize$\pm$1.0} & 3.9{\scriptsize$\pm$0.2} & 3.5{\scriptsize$\pm$0.4} & 8.0{\scriptsize$\pm$0.3} & 77.6{\scriptsize$\pm$0.1} & 18.2{\scriptsize$\pm$0.9} & 30.4{\scriptsize$\pm$0.5} \\
\bottomrule
\end{tabular}
\end{table}

\begin{table}[htbp]
\centering
\caption{\textbf{TopU-LBVS-Full GIN+FP hybrid model results - part 2.}
All values are reported as mean\,$\pm$\,std over three random seeds.
EF\,=\,Enrichment Factor. PR-AUC, ROC-AUC, BEDROC ($\alpha$\,=\,20), and LogAUC are reported on a 0--100 scale.
}
\label{tab:full_ginfp_2}
\setlength{\tabcolsep}{4pt}
\renewcommand{\arraystretch}{1.08}
\footnotesize
\begin{tabular}{l ccc||cccc}
\toprule
\textbf{Target}
  & \textbf{EF@1\%} & \textbf{EF@5\%} & \textbf{EF@10\%}
  & \textbf{PR-AUC} & \textbf{ROC-AUC} & \textbf{BEDROC} & \textbf{LogAUC} \\
\midrule

\multicolumn{8}{c}{\textbf{Nuclear Receptor}} \\[1pt]
\midrule
andr    & 4.4{\scriptsize$\pm$0.0} & 5.3{\scriptsize$\pm$0.6} & 4.2{\scriptsize$\pm$0.1} & 9.8{\scriptsize$\pm$0.7} & 82.7{\scriptsize$\pm$1.4} & 22.9{\scriptsize$\pm$1.8} & 34.2{\scriptsize$\pm$1.2} \\
esr1    & 3.3{\scriptsize$\pm$0.8} & 3.9{\scriptsize$\pm$0.0} & 3.6{\scriptsize$\pm$0.2} & 7.1{\scriptsize$\pm$0.3} & 72.2{\scriptsize$\pm$0.7} & 17.5{\scriptsize$\pm$1.0} & 28.4{\scriptsize$\pm$0.5} \\
esr2    & 3.6{\scriptsize$\pm$1.5} & 3.7{\scriptsize$\pm$0.7} & 3.1{\scriptsize$\pm$0.4} & 7.6{\scriptsize$\pm$0.9} & 75.1{\scriptsize$\pm$0.8} & 16.5{\scriptsize$\pm$2.3} & 29.3{\scriptsize$\pm$1.2} \\
gcr     & 10.7{\scriptsize$\pm$0.0} & 6.1{\scriptsize$\pm$0.0} & 5.5{\scriptsize$\pm$0.3} & 14.0{\scriptsize$\pm$0.5} & 85.5{\scriptsize$\pm$0.1} & 31.2{\scriptsize$\pm$0.8} & 39.0{\scriptsize$\pm$0.4} \\
mcr     & 0.0{\scriptsize$\pm$0.0} & 2.6{\scriptsize$\pm$0.9} & 2.7{\scriptsize$\pm$0.5} & 7.3{\scriptsize$\pm$0.6} & 76.9{\scriptsize$\pm$1.4} & 11.8{\scriptsize$\pm$1.8} & 27.1{\scriptsize$\pm$1.2} \\
ppara   & 0.0{\scriptsize$\pm$0.0} & 1.5{\scriptsize$\pm$0.0} & 2.5{\scriptsize$\pm$0.2} & 5.8{\scriptsize$\pm$0.2} & 76.7{\scriptsize$\pm$0.3} & 7.3{\scriptsize$\pm$0.1} & 26.1{\scriptsize$\pm$0.3} \\
\bottomrule
\end{tabular}
\end{table}

\begin{table}
\centering
\caption{\textbf{TopU-LBVS-Full GPS results - part 1.}
All values are reported as mean\,$\pm$\,std over three random seeds.
EF\,=\,Enrichment Factor. PR-AUC, ROC-AUC, BEDROC ($\alpha$\,=\,20), and LogAUC are reported on a 0--100 scale.
}
\label{tab:full_gps_1}
\setlength{\tabcolsep}{4pt}
\renewcommand{\arraystretch}{1.08}
\footnotesize
\begin{tabular}{l ccc||cccc}
\toprule
\textbf{Target} 
  & \textbf{EF@1\%} & \textbf{EF@5\%} & \textbf{EF@10\%}
  & \textbf{PR-AUC} & \textbf{ROC-AUC} & \textbf{BEDROC} & \textbf{LogAUC} \\
\midrule

\multicolumn{8}{c}{\textbf{Cytochrome P450}} \\[1pt]
\midrule
cp1a2   & 3.2{\scriptsize$\pm$1.4} & 2.9{\scriptsize$\pm$0.9} & 2.4{\scriptsize$\pm$0.6} & 4.9{\scriptsize$\pm$0.9} & 66.1{\scriptsize$\pm$3.2} & 10.5{\scriptsize$\pm$4.3} & 23.3{\scriptsize$\pm$2.4} \\
cp2c9   & 3.1{\scriptsize$\pm$0.4} & 2.2{\scriptsize$\pm$0.1} & 1.9{\scriptsize$\pm$0.1} & 4.7{\scriptsize$\pm$0.8} & 66.7{\scriptsize$\pm$2.6} & 7.7{\scriptsize$\pm$0.9} & 22.6{\scriptsize$\pm$1.4} \\
cp2d6   & 8.9{\scriptsize$\pm$0.3} & 6.0{\scriptsize$\pm$0.1} & 4.1{\scriptsize$\pm$0.1} & 11.1{\scriptsize$\pm$0.3} & 78.0{\scriptsize$\pm$0.6} & 26.6{\scriptsize$\pm$0.3} & 34.4{\scriptsize$\pm$0.4} \\
cp3a4   & 5.8{\scriptsize$\pm$1.9} & 4.5{\scriptsize$\pm$0.1} & 3.7{\scriptsize$\pm$0.2} & 8.9{\scriptsize$\pm$0.7} & 77.8{\scriptsize$\pm$1.0} & 19.9{\scriptsize$\pm$1.4} & 31.5{\scriptsize$\pm$1.1} \\
\midrule

\multicolumn{8}{c}{\textbf{GPCR}} \\[1pt]
\midrule
5ht6r   & 0.7{\scriptsize$\pm$1.0} & 0.8{\scriptsize$\pm$0.3} & 1.5{\scriptsize$\pm$0.3} & 4.6{\scriptsize$\pm$0.1} & 72.7{\scriptsize$\pm$0.7} & 2.6{\scriptsize$\pm$0.2} & 22.9{\scriptsize$\pm$0.3} \\
\bottomrule
\end{tabular}
\end{table}

\begin{table}[htbp]
\centering
\caption{\textbf{TopU-LBVS-Full GPS results - part 2.}
All values are reported as mean\,$\pm$\,std over three random seeds.
EF\,=\,Enrichment Factor. PR-AUC, ROC-AUC, BEDROC ($\alpha$\,=\,20), and LogAUC are reported on a 0--100 scale.
}
\label{tab:full_gps_2}
\setlength{\tabcolsep}{4pt}
\renewcommand{\arraystretch}{1.08}
\footnotesize
\begin{tabular}{l ccc||cccc}
\toprule
\textbf{Target} 
  & \textbf{EF@1\%} & \textbf{EF@5\%} & \textbf{EF@10\%}
  & \textbf{PR-AUC} & \textbf{ROC-AUC} & \textbf{BEDROC} & \textbf{LogAUC} \\
\midrule

\multicolumn{8}{c}{\textbf{Nuclear Receptor}} \\[1pt]
\midrule
andr    & 2.4{\scriptsize$\pm$2.5} & 2.2{\scriptsize$\pm$0.8} & 2.5{\scriptsize$\pm$0.5} & 6.0{\scriptsize$\pm$0.9} & 74.8{\scriptsize$\pm$1.7} & 9.5{\scriptsize$\pm$4.1} & 26.3{\scriptsize$\pm$2.1} \\
esr1    & 2.4{\scriptsize$\pm$0.8} & 2.0{\scriptsize$\pm$0.1} & 2.1{\scriptsize$\pm$0.1} & 5.3{\scriptsize$\pm$0.2} & 70.1{\scriptsize$\pm$1.3} & 7.1{\scriptsize$\pm$0.8} & 23.7{\scriptsize$\pm$0.7} \\
esr2    & 2.4{\scriptsize$\pm$0.8} & 1.7{\scriptsize$\pm$0.6} & 1.7{\scriptsize$\pm$0.1} & 4.8{\scriptsize$\pm$0.5} & 69.1{\scriptsize$\pm$1.6} & 5.1{\scriptsize$\pm$0.9} & 22.3{\scriptsize$\pm$0.7} \\
gcr     & 3.0{\scriptsize$\pm$0.8} & 4.8{\scriptsize$\pm$0.4} & 4.5{\scriptsize$\pm$0.4} & 10.6{\scriptsize$\pm$0.1} & 84.7{\scriptsize$\pm$1.3} & 22.5{\scriptsize$\pm$2.2} & 34.8{\scriptsize$\pm$1.0} \\
mcr     & 2.7{\scriptsize$\pm$3.9} & 2.6{\scriptsize$\pm$2.4} & 3.0{\scriptsize$\pm$0.0} & 8.3{\scriptsize$\pm$3.2} & 73.9{\scriptsize$\pm$2.4} & 12.4{\scriptsize$\pm$7.0} & 26.7{\scriptsize$\pm$4.0} \\
ppara   & 0.8{\scriptsize$\pm$1.1} & 2.3{\scriptsize$\pm$0.6} & 1.9{\scriptsize$\pm$0.2} & 4.0{\scriptsize$\pm$0.3} & 63.1{\scriptsize$\pm$2.7} & 6.1{\scriptsize$\pm$1.2} & 20.2{\scriptsize$\pm$1.2} \\
\bottomrule
\end{tabular}
\end{table}

\begin{table}[htbp]
\centering
\caption{textbf{TopU-LBVS-Full GPS+FP hybrid model results - part 1.}
All values are reported as mean\,$\pm$\,std over three random seeds.
EF\,=\,Enrichment Factor. PR-AUC, ROC-AUC, BEDROC ($\alpha$\,=\,20), and LogAUC are reported on a 0--100 scale.
}
\label{tab:full_gpsfp_1}
\setlength{\tabcolsep}{4pt}
\renewcommand{\arraystretch}{1.08}
\footnotesize
\begin{tabular}{l ccc||cccc}
\toprule
\textbf{Target} 
  & \textbf{EF@1\%} & \textbf{EF@5\%} & \textbf{EF@10\%}
  & \textbf{PR-AUC} & \textbf{ROC-AUC} & \textbf{BEDROC} & \textbf{LogAUC} \\
\midrule

\multicolumn{8}{c}{\textbf{Cytochrome P450}} \\[1pt]
\midrule
cp1a2   & 13.2{\scriptsize$\pm$1.8} & 7.1{\scriptsize$\pm$0.6} & 5.0{\scriptsize$\pm$0.3} & 15.3{\scriptsize$\pm$1.2} & 77.8{\scriptsize$\pm$0.4} & 33.4{\scriptsize$\pm$1.4} & 37.7{\scriptsize$\pm$0.6} \\
cp2c9   & 15.1{\scriptsize$\pm$1.2} & 7.8{\scriptsize$\pm$0.5} & 5.3{\scriptsize$\pm$0.1} & 17.5{\scriptsize$\pm$0.8} & 83.2{\scriptsize$\pm$0.1} & 36.9{\scriptsize$\pm$0.9} & 40.9{\scriptsize$\pm$0.4} \\
cp2d6   & 15.1{\scriptsize$\pm$0.6} & 8.4{\scriptsize$\pm$0.6} & 5.4{\scriptsize$\pm$0.4} & 19.2{\scriptsize$\pm$1.6} & 81.8{\scriptsize$\pm$0.9} & 38.5{\scriptsize$\pm$2.5} & 41.4{\scriptsize$\pm$1.4} \\
cp3a4   & 16.4{\scriptsize$\pm$0.3} & 8.9{\scriptsize$\pm$0.3} & 6.2{\scriptsize$\pm$0.3} & 24.0{\scriptsize$\pm$0.4} & 82.5{\scriptsize$\pm$0.6} & 43.2{\scriptsize$\pm$0.3} & 44.3{\scriptsize$\pm$0.3} \\
\midrule

\multicolumn{8}{c}{\textbf{GPCR}} \\[1pt]
\midrule
5ht6r   & 5.5{\scriptsize$\pm$1.0} & 4.6{\scriptsize$\pm$0.9} & 3.8{\scriptsize$\pm$0.2} & 8.4{\scriptsize$\pm$0.6} & 77.4{\scriptsize$\pm$0.2} & 19.9{\scriptsize$\pm$2.3} & 30.8{\scriptsize$\pm$0.8} \\
\bottomrule
\end{tabular}
\end{table}

\begin{table}[htbp]
\centering
\caption{textbf{TopU-LBVS-Full GPS+FP hybrid model results - part 2.}
All values are reported as mean\,$\pm$\,std over three random seeds.
EF\,=\,Enrichment Factor. PR-AUC, ROC-AUC, BEDROC ($\alpha$\,=\,20), and LogAUC are reported on a 0--100 scale.
}
\label{tab:full_gpsfp_2}
\setlength{\tabcolsep}{4pt}
\renewcommand{\arraystretch}{1.08}
\footnotesize
\begin{tabular}{l ccc||cccc}
\toprule
\textbf{Target} 
  & \textbf{EF@1\%} & \textbf{EF@5\%} & \textbf{EF@10\%}
  & \textbf{PR-AUC} & \textbf{ROC-AUC} & \textbf{BEDROC} & \textbf{LogAUC} \\
\midrule

\multicolumn{8}{c}{\textbf{Nuclear Receptor}} \\[1pt]
\midrule
andr    & 3.9{\scriptsize$\pm$0.7} & 6.1{\scriptsize$\pm$0.3} & 4.5{\scriptsize$\pm$0.2} & 10.7{\scriptsize$\pm$0.4} & 82.7{\scriptsize$\pm$0.7} & 25.4{\scriptsize$\pm$0.9} & 35.0{\scriptsize$\pm$0.5} \\
esr1    & 3.3{\scriptsize$\pm$1.5} & 4.1{\scriptsize$\pm$0.4} & 3.7{\scriptsize$\pm$0.2} & 7.0{\scriptsize$\pm$0.5} & 71.0{\scriptsize$\pm$0.9} & 17.7{\scriptsize$\pm$1.2} & 28.2{\scriptsize$\pm$0.9} \\
esr2    & 5.9{\scriptsize$\pm$2.2} & 4.0{\scriptsize$\pm$0.2} & 3.5{\scriptsize$\pm$0.7} & 10.3{\scriptsize$\pm$0.9} & 75.7{\scriptsize$\pm$0.7} & 19.2{\scriptsize$\pm$0.8} & 30.7{\scriptsize$\pm$0.2} \\
gcr     & 5.9{\scriptsize$\pm$0.8} & 7.0{\scriptsize$\pm$0.7} & 5.5{\scriptsize$\pm$0.2} & 14.5{\scriptsize$\pm$1.3} & 85.9{\scriptsize$\pm$0.5} & 32.5{\scriptsize$\pm$3.0} & 39.3{\scriptsize$\pm$1.4} \\
mcr     & 2.7{\scriptsize$\pm$3.9} & 3.9{\scriptsize$\pm$0.0} & 3.7{\scriptsize$\pm$0.5} & 7.8{\scriptsize$\pm$0.5} & 77.7{\scriptsize$\pm$0.8} & 13.6{\scriptsize$\pm$1.4} & 28.0{\scriptsize$\pm$0.7} \\
ppara   & 1.6{\scriptsize$\pm$1.1} & 2.0{\scriptsize$\pm$0.4} & 2.9{\scriptsize$\pm$0.2} & 6.1{\scriptsize$\pm$0.1} & 77.0{\scriptsize$\pm$0.7} & 9.2{\scriptsize$\pm$0.3} & 26.9{\scriptsize$\pm$0.3} \\
\bottomrule
\end{tabular}
\end{table}


\begin{table}[htbp]
\centering
\caption{\textbf{TopU-LBVS-Full DMPNN results - part 1.}
  All values are reported as mean\,$\pm$\,std over three random seeds.
  EF\,=\,Enrichment Factor. PR-AUC, ROC-AUC, BEDROC ($\alpha$\,=\,20), and LogAUC are reported on a 0--100 scale.
}
\label{tab:full_dmpnn_1}
\setlength{\tabcolsep}{4pt}
\renewcommand{\arraystretch}{1.08}
\footnotesize
\begin{tabular}{l ccc||cccc}
\toprule
\textbf{Target}
  & \textbf{EF@1\%} & \textbf{EF@5\%} & \textbf{EF@10\%}
  & \textbf{PR-AUC} & \textbf{ROC-AUC} & \textbf{BEDROC} & \textbf{LogAUC} \\
\midrule

\multicolumn{8}{c}{\textbf{Cytochrome P450}} \\[1pt]
\midrule
cp1a2  & 11.7{\scriptsize$\pm$1.8} & 7.5{\scriptsize$\pm$0.7} & 5.1{\scriptsize$\pm$0.4} & 16.7{\scriptsize$\pm$1.5} & 80.6{\scriptsize$\pm$1.0} & 34.3{\scriptsize$\pm$2.6} & 39.0{\scriptsize$\pm$1.6} \\
cp2c9  & 11.4{\scriptsize$\pm$2.4} & 7.7{\scriptsize$\pm$0.7} & 5.5{\scriptsize$\pm$0.3} & 16.7{\scriptsize$\pm$3.1} & 83.3{\scriptsize$\pm$1.2} & 35.6{\scriptsize$\pm$2.8} & 40.2{\scriptsize$\pm$2.2} \\
cp2d6  & 16.1{\scriptsize$\pm$1.8} & 8.6{\scriptsize$\pm$0.5} & 5.8{\scriptsize$\pm$0.3} & 20.5{\scriptsize$\pm$2.4} & 84.4{\scriptsize$\pm$0.3} & 41.0{\scriptsize$\pm$3.0} & 43.3{\scriptsize$\pm$1.6} \\
cp3a4  & 14.9{\scriptsize$\pm$1.3} & 9.5{\scriptsize$\pm$0.5} & 6.0{\scriptsize$\pm$0.1} & 22.4{\scriptsize$\pm$1.8} & 85.6{\scriptsize$\pm$0.5} & 43.7{\scriptsize$\pm$1.6} & 44.9{\scriptsize$\pm$1.0} \\
\midrule

\multicolumn{8}{c}{\textbf{GPCR}} \\[1pt]
\midrule
5ht6r  & 4.8{\scriptsize$\pm$1.0}  & 3.7{\scriptsize$\pm$0.5} & 3.6{\scriptsize$\pm$0.3} & 7.6{\scriptsize$\pm$0.4}  & 75.5{\scriptsize$\pm$1.0} & 17.2{\scriptsize$\pm$0.6} & 29.1{\scriptsize$\pm$0.9} \\
\bottomrule
\end{tabular}
\end{table}

\begin{table}[htbp]
\centering
\caption{\textbf{TopU-LBVS-Full DMPNN results - part 2.}
  All values are reported as mean\,$\pm$\,std over three random seeds.
  EF\,=\,Enrichment Factor. PR-AUC, ROC-AUC, BEDROC ($\alpha$\,=\,20), and LogAUC are reported on a 0--100 scale.
}
\label{tab:full_dmpnn_2}
\setlength{\tabcolsep}{4pt}
\renewcommand{\arraystretch}{1.08}
\footnotesize
\begin{tabular}{l ccc||cccc}
\toprule
\textbf{Target}
  & \textbf{EF@1\%} & \textbf{EF@5\%} & \textbf{EF@10\%}
  & \textbf{PR-AUC} & \textbf{ROC-AUC} & \textbf{BEDROC} & \textbf{LogAUC} \\
\midrule

\multicolumn{8}{c}{\textbf{Nuclear Receptor}} \\[1pt]
\midrule
andr   & 4.4{\scriptsize$\pm$1.2}  & 5.2{\scriptsize$\pm$0.6} & 4.1{\scriptsize$\pm$0.0} & 9.6{\scriptsize$\pm$0.3}  & 82.2{\scriptsize$\pm$0.6} & 22.8{\scriptsize$\pm$0.9} & 33.6{\scriptsize$\pm$0.4} \\
esr1   & 5.9{\scriptsize$\pm$1.5}  & 4.5{\scriptsize$\pm$0.6} & 3.3{\scriptsize$\pm$0.2} & 8.1{\scriptsize$\pm$0.8}  & 76.3{\scriptsize$\pm$0.2} & 19.3{\scriptsize$\pm$1.7} & 29.9{\scriptsize$\pm$1.3} \\
esr2   & 4.2{\scriptsize$\pm$1.7}  & 3.2{\scriptsize$\pm$0.8} & 2.7{\scriptsize$\pm$0.4} & 7.4{\scriptsize$\pm$1.4}  & 75.9{\scriptsize$\pm$0.6} & 13.9{\scriptsize$\pm$3.0} & 28.2{\scriptsize$\pm$1.3} \\
gcr    & 8.3{\scriptsize$\pm$2.2}  & 7.2{\scriptsize$\pm$0.9} & 5.8{\scriptsize$\pm$0.2} & 15.3{\scriptsize$\pm$2.0} & 87.6{\scriptsize$\pm$0.3} & 34.1{\scriptsize$\pm$3.4} & 40.5{\scriptsize$\pm$1.7} \\
mcr    & 2.7{\scriptsize$\pm$3.9}  & 3.2{\scriptsize$\pm$0.9} & 2.3{\scriptsize$\pm$0.5} & 6.4{\scriptsize$\pm$1.7}  & 71.2{\scriptsize$\pm$2.7} & 8.9{\scriptsize$\pm$3.7}  & 23.9{\scriptsize$\pm$2.5} \\
ppara  & 0.8{\scriptsize$\pm$1.1}  & 2.0{\scriptsize$\pm$0.8} & 2.9{\scriptsize$\pm$0.5} & 6.3{\scriptsize$\pm$0.5}  & 77.3{\scriptsize$\pm$1.0} & 9.9{\scriptsize$\pm$2.5}  & 27.0{\scriptsize$\pm$1.1} \\
\bottomrule
\end{tabular}
\end{table}


\begin{table}[htbp]
\centering
\caption{\textbf{TopU-LBVS-Full Molformer results - part 1.}
  All values are reported as mean\,$\pm$\,std over three random seeds.
  EF\,=\,Enrichment Factor. PR-AUC, ROC-AUC, BEDROC ($\alpha$\,=\,20), and LogAUC are reported on a 0--100 scale.
}
\label{tab:full_molformer_1}
\setlength{\tabcolsep}{4pt}
\renewcommand{\arraystretch}{1.08}
\footnotesize
\begin{tabular}{l ccc||cccc}
\toprule
\textbf{Target}
  & \textbf{EF@1\%} & \textbf{EF@5\%} & \textbf{EF@10\%}
  & \textbf{PR-AUC} & \textbf{ROC-AUC} & \textbf{BEDROC} & \textbf{LogAUC} \\
\midrule

\multicolumn{8}{c}{\textbf{Cytochrome P450}} \\[1pt]
\midrule
cp1a2  & 7.8{\scriptsize$\pm$3.1}  & 6.1{\scriptsize$\pm$0.4} & 4.6{\scriptsize$\pm$0.2} & 11.4{\scriptsize$\pm$1.9} & 79.1{\scriptsize$\pm$0.4} & 26.7{\scriptsize$\pm$3.2} & 34.7{\scriptsize$\pm$1.8} \\
cp2c9  & 11.2{\scriptsize$\pm$1.4} & 7.8{\scriptsize$\pm$0.9} & 5.6{\scriptsize$\pm$0.4} & 16.7{\scriptsize$\pm$1.4} & 84.8{\scriptsize$\pm$0.1} & 36.5{\scriptsize$\pm$3.8} & 41.0{\scriptsize$\pm$1.3} \\
cp2d6  & 11.3{\scriptsize$\pm$0.9} & 7.5{\scriptsize$\pm$0.5} & 5.4{\scriptsize$\pm$0.2} & 16.0{\scriptsize$\pm$1.4} & 84.4{\scriptsize$\pm$0.6} & 34.8{\scriptsize$\pm$2.2} & 40.2{\scriptsize$\pm$1.2} \\
cp3a4  & 14.5{\scriptsize$\pm$1.8} & 9.0{\scriptsize$\pm$0.3} & 6.1{\scriptsize$\pm$0.1} & 21.0{\scriptsize$\pm$0.7} & 84.3{\scriptsize$\pm$0.1} & 42.5{\scriptsize$\pm$0.9} & 43.7{\scriptsize$\pm$0.4} \\
\midrule

\multicolumn{8}{c}{\textbf{GPCR}} \\[1pt]
\midrule
5ht6r  & 4.1{\scriptsize$\pm$1.7}  & 3.0{\scriptsize$\pm$0.4} & 2.8{\scriptsize$\pm$0.2} & 7.8{\scriptsize$\pm$1.0}  & 75.6{\scriptsize$\pm$0.3} & 13.0{\scriptsize$\pm$1.3} & 27.9{\scriptsize$\pm$0.5} \\
\bottomrule
\end{tabular}
\end{table}

\begin{table}[htbp]
\centering
\caption{\textbf{TopU-LBVS-Full Molformer results - part 2.}
  All values are reported as mean\,$\pm$\,std over three random seeds.
  EF\,=\,Enrichment Factor. PR-AUC, ROC-AUC, BEDROC ($\alpha$\,=\,20), and LogAUC are reported on a 0--100 scale.
}
\label{tab:full_molformer_2}
\setlength{\tabcolsep}{4pt}
\renewcommand{\arraystretch}{1.08}
\footnotesize
\begin{tabular}{l ccc||cccc}
\toprule
\textbf{Target}
  & \textbf{EF@1\%} & \textbf{EF@5\%} & \textbf{EF@10\%}
  & \textbf{PR-AUC} & \textbf{ROC-AUC} & \textbf{BEDROC} & \textbf{LogAUC} \\
\midrule

\multicolumn{8}{c}{\textbf{Nuclear Receptor}} \\[1pt]
\midrule
andr   & 4.9{\scriptsize$\pm$1.8}  & 3.2{\scriptsize$\pm$0.6} & 3.6{\scriptsize$\pm$0.6} & 8.3{\scriptsize$\pm$1.1}  & 81.1{\scriptsize$\pm$1.4} & 15.9{\scriptsize$\pm$2.8} & 31.2{\scriptsize$\pm$1.7} \\
esr1   & 2.7{\scriptsize$\pm$1.5}  & 3.6{\scriptsize$\pm$0.8} & 3.3{\scriptsize$\pm$0.2} & 6.6{\scriptsize$\pm$0.8}  & 74.9{\scriptsize$\pm$0.9} & 14.9{\scriptsize$\pm$3.2} & 28.0{\scriptsize$\pm$1.6} \\
esr2   & 2.4{\scriptsize$\pm$1.7}  & 3.0{\scriptsize$\pm$0.3} & 3.0{\scriptsize$\pm$0.3} & 6.3{\scriptsize$\pm$0.3}  & 74.8{\scriptsize$\pm$0.5} & 12.3{\scriptsize$\pm$1.0} & 27.2{\scriptsize$\pm$0.5} \\
gcr    & 8.3{\scriptsize$\pm$2.2}  & 6.5{\scriptsize$\pm$1.0} & 5.3{\scriptsize$\pm$0.2} & 15.7{\scriptsize$\pm$2.1} & 86.5{\scriptsize$\pm$0.3} & 32.4{\scriptsize$\pm$2.8} & 40.4{\scriptsize$\pm$1.7} \\
mcr    & 8.2{\scriptsize$\pm$0.0}  & 3.9{\scriptsize$\pm$1.6} & 3.3{\scriptsize$\pm$0.9} & 12.1{\scriptsize$\pm$2.1} & 73.5{\scriptsize$\pm$1.5} & 18.3{\scriptsize$\pm$4.3} & 29.0{\scriptsize$\pm$2.0} \\
ppara  & 4.8{\scriptsize$\pm$2.0}  & 4.2{\scriptsize$\pm$0.2} & 2.9{\scriptsize$\pm$0.2} & 7.2{\scriptsize$\pm$0.7}  & 75.7{\scriptsize$\pm$0.5} & 15.4{\scriptsize$\pm$2.1} & 28.5{\scriptsize$\pm$0.9} \\
\bottomrule
\end{tabular}
\end{table}

\subsection{Part III: Full Per-Model Metrics, TopU-LBVS-few}
\label{sec:partIII}

Each pair of tables below reports all seven evaluation metrics for a
single baseline model across all 93 targets under the TopU$\to$TopU
few-shot protocol. Targets are grouped by class; the Tier column
indicates the split strategy (Tier~1: 6:2:2, fewer than 50 actives;
Tier~2: 7:1:2, at least 50 actives).

\begin{table}[h!]
\centering
\caption{%
\textbf{TopU-LBVS-few Morgan-RF results - part 1.} All values are reported as mean\,$\pm$\,std over three random seeds.
EF\,=\,Enrichment Factor. PR-AUC, ROC-AUC, BEDROC, and LogAUC are reported on a 0--100 scale.
}
\label{tab:few_morganrf_1}
\setlength{\tabcolsep}{3pt}
\renewcommand{\arraystretch}{1.08}
\footnotesize
\begin{tabular}{l c ccc||cccc}
\toprule
\textbf{Target} & \textbf{Tier} & \textbf{EF@1\%} & \textbf{EF@5\%} & \textbf{EF@10\%} & \textbf{PR-AUC} & \textbf{ROC-AUC} & \textbf{BEDROC} & \textbf{LogAUC} \\
\midrule
\multicolumn{9}{c}{\textbf{Cytochrome P450}} \\[1pt]
\midrule
cp1a2 & 2 & 2.3{\scriptsize$\pm$1.6} & 1.6{\scriptsize$\pm$0.6} & 1.5{\scriptsize$\pm$0.3} & 3.7{\scriptsize$\pm$0.2} & 62.2{\scriptsize$\pm$3.6} & 3.6{\scriptsize$\pm$1.5} & 25.7{\scriptsize$\pm$1.7} \\
cp2c9 & 2 & 0.0{\scriptsize$\pm$0.0} & 0.3{\scriptsize$\pm$0.2} & 0.7{\scriptsize$\pm$0.4} & 2.5{\scriptsize$\pm$0.3} & 48.6{\scriptsize$\pm$5.1} & -4.3{\scriptsize$\pm$1.4} & 20.2{\scriptsize$\pm$4.5} \\
cp2d6 & 2 & 5.2{\scriptsize$\pm$2.6} & 1.9{\scriptsize$\pm$0.9} & 1.5{\scriptsize$\pm$0.3} & 3.7{\scriptsize$\pm$1.0} & 53.9{\scriptsize$\pm$2.5} & 5.6{\scriptsize$\pm$4.4} & 23.8{\scriptsize$\pm$3.5} \\
cp3a4 & 2 & 1.1{\scriptsize$\pm$1.5} & 0.9{\scriptsize$\pm$0.4} & 1.2{\scriptsize$\pm$0.2} & 3.6{\scriptsize$\pm$1.6} & 49.8{\scriptsize$\pm$3.0} & -0.4{\scriptsize$\pm$2.5} & 19.9{\scriptsize$\pm$3.0} \\
\midrule
\multicolumn{9}{c}{\textbf{GPCR}} \\[1pt]
\midrule
5ht6r & 1 & 0.0{\scriptsize$\pm$0.0} & 0.1{\scriptsize$\pm$0.2} & 0.3{\scriptsize$\pm$0.3} & 2.4{\scriptsize$\pm$0.1} & 45.4{\scriptsize$\pm$2.0} & -5.7{\scriptsize$\pm$1.1} & 23.0{\scriptsize$\pm$2.7} \\
\bottomrule
\end{tabular}%
\end{table}
\begin{table}[htbp]
\centering
\caption{\textbf{TopU-LBVS-few Morgan-RF results - part 2.}
All values are reported as mean\,$\pm$\,std over three random seeds.
EF\,=\,Enrichment Factor. PR-AUC, ROC-AUC, BEDROC, and LogAUC are reported on a 0--100 scale.
}
\label{tab:few_morganrf_2}
\setlength{\tabcolsep}{3pt}
\renewcommand{\arraystretch}{1.08}
\footnotesize
\begin{tabular}{l c ccc||cccc}
\toprule
\textbf{Target} & \textbf{Tier} & \textbf{EF@1\%} & \textbf{EF@5\%} & \textbf{EF@10\%} & \textbf{PR-AUC} & \textbf{ROC-AUC} & \textbf{BEDROC} & \textbf{LogAUC} \\
\midrule
\multicolumn{9}{c}{\textbf{Nuclear Receptor}} \\[1pt]
\midrule
andr & 2 & 0.0{\scriptsize$\pm$0.0} & 0.0{\scriptsize$\pm$0.0} & 0.7{\scriptsize$\pm$0.6} & 2.5{\scriptsize$\pm$0.4} & 46.0{\scriptsize$\pm$9.0} & -3.8{\scriptsize$\pm$2.0} & 17.8{\scriptsize$\pm$5.3} \\
esr1 & 2 & 0.0{\scriptsize$\pm$0.0} & 0.6{\scriptsize$\pm$0.2} & 0.8{\scriptsize$\pm$0.2} & 2.6{\scriptsize$\pm$0.1} & 51.1{\scriptsize$\pm$2.2} & -2.7{\scriptsize$\pm$0.9} & 21.6{\scriptsize$\pm$2.0} \\
esr2 & 2 & 0.0{\scriptsize$\pm$0.0} & 0.3{\scriptsize$\pm$0.3} & 0.7{\scriptsize$\pm$0.7} & 2.6{\scriptsize$\pm$0.5} & 48.1{\scriptsize$\pm$7.7} & -3.9{\scriptsize$\pm$2.1} & 23.1{\scriptsize$\pm$7.4} \\
gcr & 2 & 0.0{\scriptsize$\pm$0.0} & 0.0{\scriptsize$\pm$0.0} & 0.1{\scriptsize$\pm$0.1} & 2.4{\scriptsize$\pm$0.1} & 45.4{\scriptsize$\pm$2.6} & -5.7{\scriptsize$\pm$0.8} & 21.0{\scriptsize$\pm$0.8} \\
mcr & 1 & 0.0{\scriptsize$\pm$0.0} & 0.0{\scriptsize$\pm$0.0} & 0.0{\scriptsize$\pm$0.0} & 2.4{\scriptsize$\pm$0.0} & 39.6{\scriptsize$\pm$2.6} & -6.7{\scriptsize$\pm$0.0} & 28.9{\scriptsize$\pm$5.0} \\
ppara & 1 & 0.0{\scriptsize$\pm$0.0} & 0.0{\scriptsize$\pm$0.0} & 0.2{\scriptsize$\pm$0.2} & 2.4{\scriptsize$\pm$0.0} & 43.6{\scriptsize$\pm$1.9} & -6.1{\scriptsize$\pm$0.9} & 23.1{\scriptsize$\pm$1.1} \\
\bottomrule
\end{tabular}%
\end{table}

\begin{table}[htbp]
\centering
\caption{%
\textbf{TopU-LBVS-few Tanimoto-NN results - part 1.}
All values are reported as mean\,$\pm$\,std over three random seeds.
EF\,=\,Enrichment Factor. PR-AUC, ROC-AUC, BEDROC, and LogAUC are reported on a 0--100 scale.
}
\label{tab:few_tanimoto_1}
\setlength{\tabcolsep}{3.5pt}
\renewcommand{\arraystretch}{1.08}
\footnotesize
\begin{tabular}{l c ccc||cccc}
\toprule
\textbf{Target} & \textbf{Tier}
  & \textbf{EF@1\%} & \textbf{EF@5\%} & \textbf{EF@10\%}
  & \textbf{PR-AUC} & \textbf{ROC-AUC} & \textbf{BEDROC} & \textbf{LogAUC} \\
\midrule
\multicolumn{9}{c}{\textbf{Cytochrome P450}} \\[1pt]
\midrule
cp1a2 & 2 & 1.1{\scriptsize$\pm$1.6} & 1.2{\scriptsize$\pm$0.3} & 1.2{\scriptsize$\pm$0.5} & 3.0{\scriptsize$\pm$0.2} & 50.4{\scriptsize$\pm$2.2} & 2.1{\scriptsize$\pm$0.4} & 19.3{\scriptsize$\pm$1.0} \\
cp2c9 & 2 & 0.0{\scriptsize$\pm$0.0} & 0.3{\scriptsize$\pm$0.4} & 0.6{\scriptsize$\pm$0.2} & 2.6{\scriptsize$\pm$0.1} & 51.9{\scriptsize$\pm$1.9} & -3.5{\scriptsize$\pm$0.2} & 17.7{\scriptsize$\pm$0.5} \\
cp2d6 & 2 & 4.6{\scriptsize$\pm$3.2} & 3.3{\scriptsize$\pm$0.7} & 1.7{\scriptsize$\pm$0.3} & 4.6{\scriptsize$\pm$0.5} & 54.1{\scriptsize$\pm$3.1} & 8.6{\scriptsize$\pm$2.5} & 23.6{\scriptsize$\pm$2.0} \\
cp3a4 & 2 & 1.1{\scriptsize$\pm$1.5} & 0.9{\scriptsize$\pm$0.3} & 1.1{\scriptsize$\pm$0.2} & 2.5{\scriptsize$\pm$0.3} & 46.6{\scriptsize$\pm$4.1} & -0.6{\scriptsize$\pm$1.3} & 16.3{\scriptsize$\pm$2.2} \\
\midrule
\multicolumn{9}{c}{\textbf{GPCR}} \\[1pt]
\midrule
5ht6r & 1 & 0.0{\scriptsize$\pm$0.0} & 0.7{\scriptsize$\pm$0.9} & 1.3{\scriptsize$\pm$0.5} & 3.2{\scriptsize$\pm$0.7} & 54.3{\scriptsize$\pm$8.0} & -0.9{\scriptsize$\pm$1.9} & 23.3{\scriptsize$\pm$5.0} \\
\bottomrule
\end{tabular}
\end{table}

\begin{table}[htbp]
\centering
\caption{%
\textbf{TopU-LBVS-few Tanimoto-NN results - part 2.}
All values are reported as mean\,$\pm$\,std over three random seeds.
EF\,=\,Enrichment Factor. PR-AUC, ROC-AUC, BEDROC, and LogAUC are reported on a 0--100 scale.
}
\label{tab:few_tanimoto_2}
\setlength{\tabcolsep}{3.5pt}
\renewcommand{\arraystretch}{1.08}
\footnotesize
\begin{tabular}{l c ccc||cccc}
\toprule
\textbf{Target} & \textbf{Tier}
  & \textbf{EF@1\%} & \textbf{EF@5\%} & \textbf{EF@10\%}
  & \textbf{PR-AUC} & \textbf{ROC-AUC} & \textbf{BEDROC} & \textbf{LogAUC} \\
\midrule
\multicolumn{9}{c}{\textbf{Nuclear Receptor}} \\[1pt]
\midrule
andr & 2 & 0.0{\scriptsize$\pm$0.0} & 0.0{\scriptsize$\pm$0.0} & 0.0{\scriptsize$\pm$0.0} & 2.2{\scriptsize$\pm$0.2} & 44.3{\scriptsize$\pm$4.1} & -6.2{\scriptsize$\pm$0.2} & 15.1{\scriptsize$\pm$1.8} \\
esr1 & 2 & 0.0{\scriptsize$\pm$0.0} & 0.0{\scriptsize$\pm$0.0} & 0.1{\scriptsize$\pm$0.2} & 2.4{\scriptsize$\pm$0.2} & 50.4{\scriptsize$\pm$3.8} & -5.0{\scriptsize$\pm$0.8} & 16.5{\scriptsize$\pm$2.3} \\
esr2 & 2 & 0.0{\scriptsize$\pm$0.0} & 0.0{\scriptsize$\pm$0.0} & 0.3{\scriptsize$\pm$0.4} & 2.4{\scriptsize$\pm$0.2} & 47.2{\scriptsize$\pm$4.2} & -5.7{\scriptsize$\pm$1.1} & 16.6{\scriptsize$\pm$2.7} \\
gcr & 2 & 0.0{\scriptsize$\pm$0.0} & 0.0{\scriptsize$\pm$0.0} & 0.3{\scriptsize$\pm$0.4} & 3.3{\scriptsize$\pm$0.3} & 60.0{\scriptsize$\pm$3.3} & -4.5{\scriptsize$\pm$1.1} & 24.4{\scriptsize$\pm$1.5} \\
mcr & 1 & 0.0{\scriptsize$\pm$0.0} & 0.0{\scriptsize$\pm$0.0} & 0.0{\scriptsize$\pm$0.0} & 3.3{\scriptsize$\pm$1.1} & 36.0{\scriptsize$\pm$21.2} & -6.6{\scriptsize$\pm$0.1} & 20.8{\scriptsize$\pm$13.2} \\
ppara & 1 & 0.0{\scriptsize$\pm$0.0} & 0.0{\scriptsize$\pm$0.0} & 0.0{\scriptsize$\pm$0.0} & 2.4{\scriptsize$\pm$0.1} & 44.3{\scriptsize$\pm$2.7} & -6.6{\scriptsize$\pm$0.0} & 16.0{\scriptsize$\pm$1.2} \\
\bottomrule
\end{tabular}
\end{table}

\begin{table}[htbp]
\centering
\caption{\textbf{TopU-LBVS-few GAT results - part 1.}
All values are reported as mean\,$\pm$\,std over three random seeds.
EF\,=\,Enrichment Factor. PR-AUC, ROC-AUC, BEDROC, and LogAUC are reported on a 0--100 scale.
}
\label{tab:few_gat_1}
\setlength{\tabcolsep}{3.5pt}
\renewcommand{\arraystretch}{1.08}
\footnotesize
\begin{tabular}{l c ccc||cccc}
\toprule
\textbf{Target} & \textbf{Tier}
  & \textbf{EF@1\%} & \textbf{EF@5\%} & \textbf{EF@10\%}
  & \textbf{PR-AUC} & \textbf{ROC-AUC} & \textbf{BEDROC} & \textbf{LogAUC} \\
\midrule
\multicolumn{9}{c}{\textbf{Cytochrome P450}} \\[1pt]
\midrule
cp1a2 & 2 & 0.5{\scriptsize$\pm$0.7} & 1.0{\scriptsize$\pm$0.9} & 0.7{\scriptsize$\pm$0.5} & 2.5{\scriptsize$\pm$0.2} & 48.3{\scriptsize$\pm$1.4} & -2.1{\scriptsize$\pm$2.8} & 16.4{\scriptsize$\pm$1.7} \\
cp2c9 & 2 & 2.7{\scriptsize$\pm$1.9} & 1.4{\scriptsize$\pm$0.4} & 1.7{\scriptsize$\pm$0.6} & 4.1{\scriptsize$\pm$0.9} & 60.1{\scriptsize$\pm$2.7} & 3.5{\scriptsize$\pm$2.9} & 24.3{\scriptsize$\pm$3.0} \\
cp2d6 & 2 & 0.0{\scriptsize$\pm$0.0} & 1.2{\scriptsize$\pm$0.3} & 1.2{\scriptsize$\pm$0.7} & 3.0{\scriptsize$\pm$0.4} & 52.5{\scriptsize$\pm$2.2} & 1.4{\scriptsize$\pm$3.1} & 19.9{\scriptsize$\pm$1.8} \\
cp3a4 & 2 & 0.0{\scriptsize$\pm$0.0} & 0.9{\scriptsize$\pm$0.8} & 1.0{\scriptsize$\pm$0.7} & 3.0{\scriptsize$\pm$0.5} & 54.6{\scriptsize$\pm$2.7} & -0.4{\scriptsize$\pm$4.5} & 19.4{\scriptsize$\pm$2.6} \\
\midrule
\multicolumn{9}{c}{\textbf{GPCR}} \\[1pt]
\midrule
5ht6r & 1 & 2.7{\scriptsize$\pm$3.9} & 2.0{\scriptsize$\pm$0.0} & 1.3{\scriptsize$\pm$0.5} & 3.8{\scriptsize$\pm$0.5} & 45.4{\scriptsize$\pm$3.6} & 3.8{\scriptsize$\pm$1.4} & 23.5{\scriptsize$\pm$1.3} \\
\bottomrule
\end{tabular}
\end{table}

\begin{table}[htbp]
\centering
\caption{%
\textbf{TopU-LBVS-few GAT results - part 1.}
All values are reported as mean\,$\pm$\,std over three random seeds.
EF\,=\,Enrichment Factor. PR-AUC, ROC-AUC, BEDROC, and LogAUC are reported on a 0--100 scale.
}
\label{tab:few_gat_2}
\setlength{\tabcolsep}{3.5pt}
\renewcommand{\arraystretch}{1.08}
\footnotesize
\begin{tabular}{l c ccc||cccc}
\toprule
\textbf{Target} & \textbf{Tier}
  & \textbf{EF@1\%} & \textbf{EF@5\%} & \textbf{EF@10\%}
  & \textbf{PR-AUC} & \textbf{ROC-AUC} & \textbf{BEDROC} & \textbf{LogAUC} \\
\midrule
\multicolumn{9}{c}{\textbf{Nuclear Receptor}} \\[1pt]
\midrule
andr & 2 & 0.0{\scriptsize$\pm$0.0} & 0.5{\scriptsize$\pm$0.7} & 0.8{\scriptsize$\pm$0.6} & 2.7{\scriptsize$\pm$0.2} & 49.8{\scriptsize$\pm$0.8} & -3.3{\scriptsize$\pm$3.3} & 18.8{\scriptsize$\pm$1.7} \\
esr1 & 2 & 0.0{\scriptsize$\pm$0.0} & 0.6{\scriptsize$\pm$0.4} & 1.0{\scriptsize$\pm$0.7} & 3.0{\scriptsize$\pm$0.3} & 57.4{\scriptsize$\pm$2.3} & -2.0{\scriptsize$\pm$3.1} & 20.4{\scriptsize$\pm$2.3} \\
esr2 & 2 & 0.0{\scriptsize$\pm$0.0} & 0.6{\scriptsize$\pm$0.8} & 1.5{\scriptsize$\pm$0.8} & 3.6{\scriptsize$\pm$1.8} & 49.5{\scriptsize$\pm$12.7} & 1.0{\scriptsize$\pm$5.8} & 23.4{\scriptsize$\pm$9.6} \\
gcr & 2 & 2.7{\scriptsize$\pm$3.9} & 1.2{\scriptsize$\pm$0.8} & 0.9{\scriptsize$\pm$0.7} & 4.0{\scriptsize$\pm$2.5} & 44.9{\scriptsize$\pm$8.3} & 0.2{\scriptsize$\pm$6.4} & 19.4{\scriptsize$\pm$6.3} \\
mcr & 1 & 0.0{\scriptsize$\pm$0.0} & 2.7{\scriptsize$\pm$3.9} & 1.5{\scriptsize$\pm$2.1} & 6.4{\scriptsize$\pm$4.1} & 38.8{\scriptsize$\pm$25.6} & 2.1{\scriptsize$\pm$9.8} & 32.2{\scriptsize$\pm$23.2} \\
ppara & 1 & 6.8{\scriptsize$\pm$4.8} & 2.4{\scriptsize$\pm$0.0} & 2.1{\scriptsize$\pm$0.6} & 7.2{\scriptsize$\pm$1.4} & 56.5{\scriptsize$\pm$8.1} & 9.4{\scriptsize$\pm$2.9} & 31.4{\scriptsize$\pm$3.5} \\
\bottomrule
\end{tabular}
\end{table}

\begin{table}[htbp]
\centering
\caption{%
\textbf{TopU-LBVS-few GAT results - part 1.}
All values are reported as mean\,$\pm$\,std over three random seeds.
EF\,=\,Enrichment Factor. PR-AUC, ROC-AUC, BEDROC, LogAUC are reported on a 0--100 scale.
}
\label{tab:few_gatfp_1}
\setlength{\tabcolsep}{3.0pt}
\renewcommand{\arraystretch}{1.08}
\footnotesize
\begin{tabular}{l c ccc||cccc}
\toprule
\textbf{Target} & \textbf{Tier} & \textbf{EF@1\%} & \textbf{EF@5\%} & \textbf{EF@10\%} & \textbf{PR-AUC} & \textbf{ROC-AUC} & \textbf{BEDROC} & \textbf{LogAUC} \\
\midrule
\multicolumn{9}{c}{\textbf{Cytochrome P450}} \\[1pt]
\midrule
cp1a2 & 2 & 1.1{\scriptsize$\pm$1.6} & 1.5{\scriptsize$\pm$0.6} & 1.1{\scriptsize$\pm$0.3} & 2.6{\scriptsize$\pm$0.4} & 46.7{\scriptsize$\pm$3.1} & 0.0{\scriptsize$\pm$2.4} & 17.0{\scriptsize$\pm$2.4} \\
cp2c9 & 2 & 0.0{\scriptsize$\pm$0.0} & 0.3{\scriptsize$\pm$0.4} & 0.7{\scriptsize$\pm$0.2} & 2.3{\scriptsize$\pm$0.1} & 45.7{\scriptsize$\pm$2.2} & -3.1{\scriptsize$\pm$1.3} & 16.0{\scriptsize$\pm$0.9} \\
cp2d6 & 2 & 1.1{\scriptsize$\pm$1.6} & 1.2{\scriptsize$\pm$0.3} & 1.1{\scriptsize$\pm$0.3} & 2.6{\scriptsize$\pm$0.4} & 42.1{\scriptsize$\pm$1.7} & 0.2{\scriptsize$\pm$1.9} & 15.6{\scriptsize$\pm$1.4} \\
cp3a4 & 2 & 2.1{\scriptsize$\pm$1.5} & 1.1{\scriptsize$\pm$0.8} & 1.0{\scriptsize$\pm$0.0} & 3.0{\scriptsize$\pm$0.3} & 47.2{\scriptsize$\pm$6.6} & 0.2{\scriptsize$\pm$2.2} & 17.4{\scriptsize$\pm$1.7} \\
\midrule
\multicolumn{9}{c}{\textbf{GPCR}} \\[1pt]
\midrule
5ht6r & 1 & 0.0{\scriptsize$\pm$0.0} & 0.7{\scriptsize$\pm$0.9} & 0.7{\scriptsize$\pm$0.9} & 2.7{\scriptsize$\pm$0.6} & 45.5{\scriptsize$\pm$5.8} & -2.3{\scriptsize$\pm$4.1} & 19.5{\scriptsize$\pm$3.8} \\
\bottomrule
\end{tabular}
\end{table}

\begin{table}[htbp]
\centering
\caption{%
\textbf{TopU-LBVS-few GAT results - part 2.}
All values are reported as mean\,$\pm$\,std over three random seeds.
EF\,=\,Enrichment Factor. PR-AUC, ROC-AUC, BEDROC, LogAUC are reported on a 0--100 scale.
}
\label{tab:few_gatfp_2}
\setlength{\tabcolsep}{3.0pt}
\renewcommand{\arraystretch}{1.08}
\footnotesize
\begin{tabular}{l c ccc||cccc}
\toprule
\textbf{Target} & \textbf{Tier} & \textbf{EF@1\%} & \textbf{EF@5\%} & \textbf{EF@10\%} & \textbf{PR-AUC} & \textbf{ROC-AUC} & \textbf{BEDROC} & \textbf{LogAUC} \\
\midrule
\multicolumn{9}{c}{\textbf{Nuclear Receptor}} \\[1pt]
\midrule
andr & 2 & 0.0{\scriptsize$\pm$0.0} & 1.5{\scriptsize$\pm$1.2} & 0.8{\scriptsize$\pm$0.6} & 2.8{\scriptsize$\pm$0.2} & 45.8{\scriptsize$\pm$4.9} & -0.6{\scriptsize$\pm$3.4} & 19.2{\scriptsize$\pm$1.7} \\
esr1 & 2 & 0.0{\scriptsize$\pm$0.0} & 0.3{\scriptsize$\pm$0.4} & 0.3{\scriptsize$\pm$0.4} & 2.2{\scriptsize$\pm$0.4} & 41.8{\scriptsize$\pm$5.5} & -3.8{\scriptsize$\pm$2.8} & 14.2{\scriptsize$\pm$3.2} \\
esr2 & 2 & 0.0{\scriptsize$\pm$0.0} & 0.0{\scriptsize$\pm$0.0} & 0.0{\scriptsize$\pm$0.0} & 2.0{\scriptsize$\pm$0.1} & 35.3{\scriptsize$\pm$3.7} & -5.6{\scriptsize$\pm$0.2} & 12.7{\scriptsize$\pm$1.7} \\
gcr & 2 & 0.0{\scriptsize$\pm$0.0} & 0.6{\scriptsize$\pm$0.8} & 0.9{\scriptsize$\pm$0.0} & 2.9{\scriptsize$\pm$0.3} & 50.9{\scriptsize$\pm$5.1} & -2.3{\scriptsize$\pm$1.3} & 21.2{\scriptsize$\pm$2.1} \\
mcr & 1 & 0.0{\scriptsize$\pm$0.0} & 0.0{\scriptsize$\pm$0.0} & 0.0{\scriptsize$\pm$0.0} & 2.3{\scriptsize$\pm$0.4} & 21.0{\scriptsize$\pm$13.5} & -6.7{\scriptsize$\pm$0.0} & 9.3{\scriptsize$\pm$6.5} \\
ppara & 1 & 0.0{\scriptsize$\pm$0.0} & 0.0{\scriptsize$\pm$0.0} & 0.4{\scriptsize$\pm$0.6} & 2.7{\scriptsize$\pm$0.6} & 45.4{\scriptsize$\pm$7.0} & -4.9{\scriptsize$\pm$2.0} & 18.8{\scriptsize$\pm$5.4} \\
\bottomrule
\end{tabular}
\end{table}

\begin{table}[htbp]
\centering
\caption{%
\textbf{TopU-LBVS-few GIN results - part 1.}
All values are reported as mean\,$\pm$\,std over three random seeds.
EF\,=\,Enrichment Factor. PR-AUC, ROC-AUC, BEDROC, and LogAUC are reported on a 0--100 scale.
}
\label{tab:few_gin_1}
\setlength{\tabcolsep}{3.0pt}
\renewcommand{\arraystretch}{1.08}
\footnotesize
\begin{tabular}{l c ccc||cccc}
\toprule
\textbf{Target} & \textbf{Tier}
  & \textbf{EF@1\%} & \textbf{EF@5\%} & \textbf{EF@10\%}
  & \textbf{PR-AUC} & \textbf{ROC-AUC} & \textbf{BEDROC} & \textbf{LogAUC} \\
\midrule
\multicolumn{9}{c}{\textbf{Cytochrome P450}} \\[1pt]
\midrule
cp1a2 & 2 & 0.0{\scriptsize$\pm$0.0} & 1.2{\scriptsize$\pm$0.7} & 0.9{\scriptsize$\pm$0.2} & 2.9{\scriptsize$\pm$0.4} & 53.9{\scriptsize$\pm$6.4} & -0.6{\scriptsize$\pm$2.3} & 19.4{\scriptsize$\pm$2.5} \\
cp2c9 & 2 & 1.4{\scriptsize$\pm$1.9} & 1.9{\scriptsize$\pm$1.0} & 1.5{\scriptsize$\pm$0.5} & 3.6{\scriptsize$\pm$0.4} & 55.9{\scriptsize$\pm$2.7} & 3.3{\scriptsize$\pm$3.4} & 23.1{\scriptsize$\pm$1.7} \\
cp2d6 & 2 & 0.0{\scriptsize$\pm$0.0} & 0.5{\scriptsize$\pm$0.3} & 1.0{\scriptsize$\pm$0.4} & 3.0{\scriptsize$\pm$0.2} & 55.2{\scriptsize$\pm$1.4} & -0.9{\scriptsize$\pm$0.4} & 20.5{\scriptsize$\pm$0.9} \\
cp3a4 & 2 & 0.0{\scriptsize$\pm$0.0} & 1.1{\scriptsize$\pm$0.3} & 1.3{\scriptsize$\pm$0.3} & 3.2{\scriptsize$\pm$0.6} & 56.7{\scriptsize$\pm$7.7} & 2.0{\scriptsize$\pm$1.6} & 21.4{\scriptsize$\pm$3.6} \\
\midrule
\multicolumn{9}{c}{\textbf{GPCR}} \\[1pt]
\midrule
5ht6r & 1 & 2.7{\scriptsize$\pm$3.9} & 2.0{\scriptsize$\pm$1.6} & 2.7{\scriptsize$\pm$0.5} & 5.8{\scriptsize$\pm$0.8} & 60.9{\scriptsize$\pm$11.6} & 8.8{\scriptsize$\pm$2.4} & 33.1{\scriptsize$\pm$5.3} \\
\bottomrule
\end{tabular}
\end{table}
\begin{table}[htbp]
\centering
\caption{%
\textbf{TopU-LBVS-few GIN results - part 2.}
All values are reported as mean\,$\pm$\,std over three random seeds.
EF\,=\,Enrichment Factor. PR-AUC, ROC-AUC, BEDROC, and LogAUC are reported on a 0--100 scale.
}
\label{tab:few_gin_2}
\setlength{\tabcolsep}{3.0pt}
\renewcommand{\arraystretch}{1.08}
\footnotesize
%
\end{table}

\begin{table}[t]
\centering
\caption{%
\textbf{TopU-LBVS-few GIN-FP results - part 1.}
All values are reported as mean\,$\pm$\,std over three random seeds.
EF\,=\,Enrichment Factor. PR-AUC, ROC-AUC, BEDROC,  and LogAUC are reported on a 0--100 scale.
}
\label{tab:few_ginfp_1}
\footnotesize
%
%

\end{table}
\begin{table}[t]
\centering
\caption{%
\textbf{TopU-LBVS-few GIN-FP results - part 2.}
All values are reported as mean\,$\pm$\,std over three random seeds.
EF\,=\,Enrichment Factor. PR-AUC, ROC-AUC, BEDROC, and LogAUC are reported on a 0--100 scale.
}
\label{tab:few_ginfp_2}
\footnotesize
%
%

\end{table}

\begin{table}[t]
\centering
\caption{%
\textbf{TopU-LBVS-few GPS results - part 1.}
All values are reported as mean\,$\pm$\,std over three random seeds.
EF\,=\,Enrichment Factor. PR-AUC, ROC-AUC, BEDROC,  and LogAUC are reported on a 0--100 scale.
}
\label{tab:few_gps_1}
\footnotesize
%
%

\end{table}

\begin{table}[t]
\centering
\caption{%
\textbf{TopU-LBVS-few GPS results - part 2.}
All values are reported as mean\,$\pm$\,std over three random seeds.
EF\,=\,Enrichment Factor. PR-AUC, ROC-AUC, BEDROC,  and LogAUC are reported on a 0--100 scale.
}
\label{tab:few_gps_2}
\footnotesize
%
%

\end{table}

\begin{table}[t]
\centering
\caption{%
\textbf{TopU-LBVS-few GPS-FP results - part 1.}
All values are reported as mean\,$\pm$\,std over three random seeds.
EF\,=\,Enrichment Factor. PR-AUC, ROC-AUC, BEDROC, and LogAUC are reported on a 0--100 scale.
}
\label{tab:few_gpsfp_1}
\resizebox{\textwidth}{!}{%
%
%
}
\end{table}
\begin{table}[t]
\centering
\caption{%
\textbf{TopU-LBVS-few GPS-FP results - part 2.}
All values are reported as mean\,$\pm$\,std over three random seeds.
EF\,=\,Enrichment Factor. PR-AUC, ROC-AUC, BEDROC, and LogAUC are reported on a 0--100 scale.
}
\label{tab:few_gpsfp_2}
\resizebox{\textwidth}{!}{%
%
%
}
\end{table}

\begin{table}[t]
\centering
\caption{%
\textbf{TopU-LBVS-few DMPNN results - part 1.}
All values are reported as mean\,$\pm$\,std over three random seeds.
EF\,=\,Enrichment Factor. PR-AUC, ROC-AUC, BEDROC,  and LogAUC are reported on a 0--100 scale.
}
\label{tab:few_dmpnn_1}
\setlength{\tabcolsep}{3pt}
\renewcommand{\arraystretch}{1.08}
\footnotesize
%
\end{table}
\begin{table}[htbp]
\centering
\caption{%
\textbf{TopU-LBVS-few DMPNN results - part 2.}
All values are reported as mean\,$\pm$\,std over three random seeds.
EF\,=\,Enrichment Factor. PR-AUC, ROC-AUC, BEDROC,  and LogAUC are reported on a 0--100 scale.
}
\label{tab:few_dmpnn_2}
\setlength{\tabcolsep}{3pt}
\renewcommand{\arraystretch}{1.08}
\footnotesize
%
\end{table}

\begin{table}[htbp]
\centering
\caption{%
\textbf{TopU-LBVS-few Molformer results - part 1.}
All values are reported as mean\,$\pm$\,std over three random seeds.
EF\,=\,Enrichment Factor. PR-AUC, ROC-AUC, BEDROC,  and LogAUC are reported on a 0--100 scale.
}
\label{tab:few_molformer_1}
\setlength{\tabcolsep}{3pt}
\renewcommand{\arraystretch}{1.08}
\scriptsize
%
\end{table}
\begin{table}[htbp]
\centering
\caption{%
\textbf{TopU-LBVS-few Molformer results - part 2.}
All values are reported as mean\,$\pm$\,std over three random seeds.
EF\,=\,Enrichment Factor. PR-AUC, ROC-AUC, BEDROC,  and LogAUC are reported on a 0--100 scale.
}
\label{tab:few_molformer_2}
\setlength{\tabcolsep}{3pt}
\renewcommand{\arraystretch}{1.08}
\scriptsize
%
\end{table}

\end{document}